\documentclass[journal, 10pt, twocolumn]{IEEEtran}

\usepackage{cite}

\ifCLASSINFOpdf
\else
   \usepackage[dvips]{graphicx}
   \DeclareGraphicsExtensions{.eps}
\fi
\usepackage[cmex10]{amsmath}
\usepackage{algorithmic}

\usepackage{amssymb}

\usepackage{array}

\usepackage{mdwmath}
\usepackage{mdwtab}

\usepackage{eqparbox}

\usepackage{multirow}

\usepackage{fixltx2e}

\usepackage{enumerate}

\usepackage[hyphens]{url}
\usepackage{hyperref}
\usepackage{breakurl}

\usepackage{tabularx}

\begin{document}
%
\title{Data-Driven Learning of a Union of Sparsifying Transforms Model for Blind Compressed Sensing}



%
%
%

\author{Saiprasad~Ravishankar,~\IEEEmembership{Member,~IEEE,}~Yoram~Bresler,~\IEEEmembership{Fellow,~IEEE}
\thanks{DOI: 10.1109/TCI.2016.2567299. $\copyright$ 2016 IEEE. Personal use of this material is permitted. Permission from IEEE must be obtained for all other uses, in any current or future media, including reprinting/republishing this material for advertising or promotional purposes, creating new collective works, for resale or redistribution to servers or lists, or reuse of any copyrighted component of this work in other works.}
\thanks{This work was supported in part by the National Science Foundation (NSF) under grant CCF-1320953. A short version of this work appears elsewhere \cite{saibrespsie}.}
\thanks{S. Ravishankar is with the Department of Electrical Engineering and Computer Science, University of Michigan, Ann Arbor, MI, 48109 USA email: ravisha@umich.edu. Y. Bresler is with the Department of Electrical and Computer Engineering and the  Coordinated Science Laboratory, University of Illinois, Urbana-Champaign, IL, 61801 USA e-mail: ybresler@illinois.edu.}}
\maketitle
 
\begin{abstract}
Compressed sensing is a powerful tool in applications such as magnetic resonance imaging (MRI). It enables accurate recovery of images from highly undersampled measurements by exploiting the sparsity of the images or image patches in a transform domain or dictionary. In this work, we focus on blind compressed sensing (BCS), where the underlying sparse signal model is a priori unknown, and propose a framework to simultaneously reconstruct the underlying image as well as the unknown model from highly undersampled measurements. Specifically, our model is that the patches of the underlying image(s) are approximately sparse in a transform domain. We also extend this model to a union of transforms model that better captures the diversity of features in natural images. The proposed block coordinate descent type algorithms for blind compressed sensing are highly efficient, and are guaranteed to converge to at least the partial global and partial local minimizers of the highly non-convex BCS problems. Our numerical experiments show that the proposed framework usually leads to better quality of image reconstructions in MRI compared to several recent image reconstruction methods. Importantly, the learning of a union of sparsifying transforms leads to better image reconstructions than a single adaptive transform.
\end{abstract}










%

\IEEEpeerreviewmaketitle

\section{Introduction}
\label{sec1} 

The sparsity of signals and images in transform domains or dictionaries is a key property that has been exploited in several applications including compression  \cite{jpg2}, denoising, and in inverse problems in imaging. Sparsity in either a fixed or data-adaptive dictionary or transform is fundamental to the success of popular techniques such as compressed sensing that aim to reconstruct images from a few sensor measurements. In this work, we focus on methods for blind compressed sensing, where not only the image but also the dictionary or transform is estimated from the measurements. In the following, we briefly review compressed sensing and blind compressed sensing, before summarizing our contributions in this work.

\subsection{Compressed Sensing}

Compressed sensing (CS) \cite{tao1, don, cand} (see also \cite{feng96a, BreFen-C96c, Fen-PT97, VenBre-C98b, BreGasVen-C99, GasBre-C00a, YeBreMou-J02, Bre-C2008a} for the earliest versions of CS for Fourier-sparse signals and for Fourier imaging) is a technique that enables accurate reconstructions of images from far fewer measurements than the number of unknowns. To do so, it assumes that the underlying image is sufficiently sparse in some transform domain or dictionary, and that the measurement acquisition procedure is incoherent, in an appropriate sense, with the transform.
The image reconstruction problem in CS is often formulated (using a convex relaxation of the $\ell_0$ counting ``norm" for sparsity) as follows
\cite{lustig}
\begin{equation}\label{eq1}
\min_{x}\:\left \| Ax-y \right \|_{2}^{2}+\lambda \left \| \Psi x \right \|_{1}
\end{equation}
Here, $x \in \mathbb{C}^{p}$ is a vectorized version of the image to be reconstructed, $\Psi \in \mathbb{C}^{t \times p}$ is a sparsifying transform for the image (often chosen as orthonormal), $y \in \mathbb{C}^{m}$ denotes the imaging measurements, and $A \in \mathbb{C}^{m \times p}$, with $m \ll p$ is the sensing or measurement matrix for the application.

Compressed sensing has become an increasingly attractive tool for imaging in recent years. CS has been applied to several imaging modalities such as magnetic resonance imaging (MRI) \cite{lustig, lustig2, Char, josh, Yoo, kal, Qu11}, computed tomography (CT) \cite{chen, choi11, CT11}, and Positron emission tomography (PET) imaging \cite{vali1, malc2}, demonstrating high quality reconstructions from few measurements. Such compressive measurements may help reduce the radiation dosage in CT, or reduce the scan times in MRI.

In this work, we will develop methods that apply to compressed sensing and other general inverse problems. We illustrate our methods in the particular application of MRI.
MRI is a non-invasive and non-ionizing imaging modality that offers a variety of contrast mechanisms, and enables excellent visualization of anatomical structures and physiological functions.  
However, the data in MRI, which are samples in k-space or the spatial Fourier transform of the object, are acquired sequentially in time. 
Hence, a drawback of MRI that affects both clinical throughput and image quality is that it is a relatively slow imaging technique.
Although there have been advances in scanner hardware \cite{pMRI-Survey} and pulse sequences, the rate at which MR data are acquired is limited by MR physics and  physiological constraints on RF energy deposition.
CS accelerates the data acquisition in MRI by collecting fewer k-space measurements than mandated by Nyquist sampling conditions. In particular, for MRI, the sensing matrix $A$ in \eqref{eq1} is $ F_{u}  \in \mathbb{C}^{m \times p} $, the undersampled Fourier encoding matrix. 

\subsection{Blind Compressed Sensing}

While compressed sensing techniques typically work with fixed sparsifying transforms such as Wavelets, finite differences (total variation) \cite{lustig, ma}, Contourlets \cite{xia}, etc. to reconstruct images, there has been a growing interest in data-driven models in recent years. Some recent works considered learning dictionaries \cite{Chen2010} or tight frames \cite{wangliuj} from reference images, but in these methods, the model is kept fixed during the CS image reconstruction process, and not adapted to better sparsify and reconstruct the features/dynamics of the underlying (unknown) images. In this work, we instead focus on the subject of blind compressed sensing (BCS)  \cite{bresai, symul, Glei12, lingal1, wangying, josecab1, huangbays, awate, josecab2, swang1, syber, luke2, sabressiims1}. In BCS, the sparse model for the underlying image(s) or image patches is assumed unknown a priori. The goal in BCS is then to reconstruct both the image(s) as well as the dictionary or transform from only the undersampled measurements. Thus, the BCS problem is harder than conventional compressed sensing. However, BCS allows the sparse model to be better adaptive to the current (unknown) image(s).

In an early work \cite{fowl1}, Fowler proposed a method for recovering the principal eigenvectors of data (principal components) from random projections. This work shares similarities with BCS in its attempt to estimate a model for data from compressive measurements. However, while the prior work \cite{fowl1} learns an under-complete principal components model, BCS can enable the learning of much richer data models by exploiting sparsity criteria.

The sparse model in BCS can take a variety of forms. For example, the well-known synthesis dictionary model suggests that a real-world signal $ z  \in \mathbb{C}^{n} $ can be approximately represented as a linear combination of a small number of (or, a sparse set of)  atoms or columns from a synthesis dictionary $ D  \in \mathbb{C}^{n \times m}$, i.e., $z = D \alpha + e$ with $\alpha \in \mathbb{C}^{m} $ sparse, or $\left \| \alpha \right \|_{0}\ll n$, and $e$ is the approximation or modeling error in the signal domain \cite{ambruck}. The alternative sparsifying transform model (which is a generalized analysis model \cite{sabres}) suggests that the signal $z $ is approximately sparsifiable using a transform $ W  \in \mathbb{C}^{m \times n} $, i.e., $Wz = \alpha + \eta$, where $\alpha \in \mathbb{C}^{m}$ is sparse in some sense, and $\eta$ is a small residual error in the \emph{transform domain} rather than in the signal domain. The advantage of the transform model over the synthesis dictionary model is that sparse coding (the process of finding $\alpha$ for a signal $z$, using a given $D$ or $W$) can be performed cheaply by thresholding \cite{sabres}, whereas it is NP-hard (Non-deterministic Polynomial-time hard) in the latter case \cite{npb, npa}. In recent years, the data-driven adaptation of such sparse signal models has received increasing attention and has been shown to be advantageous in several applications \cite{elad2, elad3, elad6, bresai, doubsp2l, sbclsTS2, saiwen}.

In prior work on BCS \cite{bresai}, we proposed synthesis dictionary-based blind compressed sensing for MRI. The overlapping patches of the underlying image were modeled as sparse in an unknown patch-based dictionary (of size much smaller than the image), and this dictionary was learnt jointly with the image from undersampled k-space measurements. BCS techniques can provide much better image reconstructions for MRI compared to conventional CS methods that use only a fixed sparsifying transform or dictionary \cite{bresai, symul, lingal1, wangying,  josecab2, swang1}. However, previous dictionary-based BCS methods, which typically solve non-convex or NP-hard problems by block coordinate descent type approaches, tend to be computationally expensive, and lack any convergence guarantees.

\subsection{Contributions}

In this work, we focus on the efficient sparsifying transform model \cite{sabres}, and study a particular transform-based blind compressed sensing framework that has not been explored in prior work \cite{sabressiims1, syber}. The proposed framework is to simultaneously reconstruct the underlying image(s) and learn the transform model from compressive measurements.
First, we model the patches of the underlying image(s) as approximately sparse in a \emph{single} (square) transform domain. 
We then further extend this model to a \emph{union of transforms model} (also known as OCTOBOS model \cite{saiwen}) that is better suited to capture the diversity of features in natural images.
The transforms in our formulations are constrained to be \emph{unitary}. This results in computationally cheap transform update and image update steps in the proposed block coordinate descent type BCS algorithms. We also work with an $\ell_0$ penalty  (instead of constraint)  for sparsity in our formulations, which enables a very efficient and exact sparse coding step involving thresholding in our block coordinate descent algorithms.
The $\ell_0$ penalty also plays a key role in enabling the generalization of the proposed formulation and algorithm for single transform BCS to the union of transforms case.
We present convergence results for our algorithms that solve the single transform or union of transforms BCS problems. In both cases, the algorithms are guaranteed to converge to at least the partial global and partial local minimizers of the highly non-convex  BCS problems. 
Our numerical experiments show that the proposed BCS framework usually leads to better quality of image reconstructions in MRI compared to several recent image reconstruction methods. Importantly, the learning of a union of sparsifying transforms leads to better image reconstructions than when learning a single transform. The data adaptive regularizers proposed in this work can be used in general inverse problem settings, and are not restricted to compressed sensing. 

\subsection{Relation to Recent Works}

In prior work, we proposed the idea of learning square sparsifying transforms from training signals \cite{sabres, sbclsTS2}. A method for learning a union of transforms model from training data has also been proposed \cite{saiwen}. However, these works did not consider the problem of jointly estimating images and image models from compressive measurements (i.e., blind compressed sensing). The latter idea was considered in recent papers \cite{sabressiims1, syber}\footnote{The method in \cite{syber} lacks any convergence analysis and also involves many parameters (e.g., error thresholds to determine patch-wise sparsity levels) that may be hard to tune in practice.}, where methods for simultaneously reconstructing images and learning square sparsifying transforms for image patches were considered. In this work, we instead investigate a novel and efficient framework for blind compressed sensing involving the richer union of transforms model. We use as a building block a specific square transform-based blind compressed sensing formulation involving an $\ell_0$ sparsity penalty and unitary transform constraint that is related to formulations ((P2) and (P3)) in prior work \cite{sabressiims1}, but was not explicitly considered therein. We show promise for the proposed methods for MR image reconstruction, where they achieve improved or faster reconstructions compared to our recent TLMRI method \cite{sabressiims1}, which uses a single adaptive transform.

The application of the methods proposed in this work for MRI was briefly considered in a very recent conference publication \cite{saibrespsie}. However, unlike the conference work, here, we also provide detailed theoretical convergence results for the proposed union of transforms-based blind compressed sensing method. An empirical study of the convergence and (blind) learning behavior of the proposed methods is also presented here, along with expanded experimental results and comparisons.  Importantly, the theoretical and empirical convergence results presented in this work are for union of transforms-based blind compressed sensing rather than for (the simpler) transform learning (from training signals) \cite{saiwen, sbclsTS2}. The theoretical results here generalize results from our prior work  \cite{sabressiims1} to related as well as more complex scenarios.


\subsection{Organization}

The rest of this paper is organized as follows. Section \ref{sec2} describes our transform learning-based blind compressed sensing formulations and their properties. 
Section \ref{sec3} derives efficient block coordinate descent algorithms for the proposed problems, and discusses the algorithms' computational costs. Section \ref{sec40} discusses the theoretical convergence properties of the proposed algorithms.
Section \ref{sec4} presents experimental results demonstrating the practical convergence behavior and performance of the proposed schemes for the MRI application. Section \ref{sec5} presents our conclusions and proposals for future work.

\section{Blind Compressed Sensing Problem Formulations}
\label{sec2} 


The image reconstruction  Problem \eqref{eq1} for compressed sensing is a particular instance of the following constrained regularized inverse problem, with $\mathcal{S} = \mathbb{C}^{p}$
\begin{equation}\label{reginveq1}
\min_{x \in \mathcal{S}}\:\left \| Ax-y \right \|_{2}^{2}+ \zeta (x)
\end{equation}
The regularizer $\zeta (x) = \lambda \left \| \Psi x \right \|_{1}$ encourages sparsity of the image in a fixed sparsifying transform $\Psi$. To overcome the limitations of such a non-adaptive CS formulation, or the limitations of the recent dictionary-based BCS methods, we explore sparsifying transform-based BCS formulations in this work. These are discussed in the following subsections.

\subsection{Unitary BCS}
 Sparsifying transform learning has been demonstrated to be effective and efficient in several applications, while also enjoying good convergence properties \cite{sbclsTS2, saonli1, saonli2, saiwen}. Here, we propose to use the following transform learning regularizer \cite{sbclsTS2}
\begin{align*}
\zeta(x) = & \frac{1}{\nu} \min_{W, B}\:  \sum_{j=1}^{N} \begin{Bmatrix}
\left \| W P_{j}x- b_{j} \right \|_{2}^{2} + \eta^{2}  \left \| b_{j} \right \|_{0}
\end{Bmatrix} \\
&\;\;\;\;\;\; \mathrm{s.t.}\;\;  W^{H}W=I
\end{align*}
along with the constraint set $\mathcal{S} = \left \{ x \in \mathbb{C}^{p} : \left \| x \right \|_{2}\leq C \right \}$ within Problem \eqref{reginveq1} to arrive at the following transform BCS formulation
\begin{align}
\nonumber (\mathrm{P1})\:\: & \min_{x,W, B}\:  \nu \left \| Ax-y \right \|_{2}^{2}  +  \sum_{j=1}^{N} \begin{Bmatrix}
\left \| W P_{j}x- b_{j} \right \|_{2}^{2} + \eta^{2}  \left \| b_{j} \right \|_{0}
\end{Bmatrix}\\
\nonumber & \;\, \, \mathrm{s.t.}\;\;  W^{H}W=I, \;\; \left \| x \right \|_{2}\leq C.
\end{align}
Here, $ P_{j} \in \mathbb{C}^{n \times p} $ represents the operator that extracts a patch\footnote{For 2D imaging, this would be a $d \times d$ patch, with $n = d^{2}$ pixels. For 3D or 4D imaging, the corresponding 3D or 4D patches would have sizes $d \times d \times d$ or $d \times d \times d \times d$, with $n = d^{3}$ or $n = d^{4}$, respectively.} as a vector $P_{j}x  \in \mathbb{C}^{n}$ from the image $x$, and $W \in \mathbb{C}^{n \times n}$ is a square sparsifying transform for the patches of the image.
A total of $N$ overlapping image patches are assumed, and $\nu >0$, $\eta>0$ are weights in (P1).
The term $\left \| W P_{j}x- b_{j} \right \|_{2}^{2}$ in the cost denotes the sparsification error or transform domain residual \cite{sabres} for the $j$th image patch, with $b_{j}$ denoting the transform \emph{sparse code} (i.e., the sparse approximation to the transformed patch).
The penalty $\left \| b_{j} \right \|_{0}$ counts the number of non-zeros in $b_{j}$. We use $B \in \mathbb{C}^{n \times N}$ to denote the matrix that has the sparse codes $b_{j}$ as its columns. The constraint $W^{H}W=I$, with $I$ denoting the $n \times n$ identity matrix, restricts the set of feasible transforms to unitary matrices. The constraint $ \left \| x \right \|_{2}\leq C$ with $C>0$ in (P1) enforces any prior knowledge on the signal energy (or, range). 

In the absence of the $\left \| x \right \|_{2}\leq C$ condition, the objective in (P1) is non-coercive. In particular, consider $W = I$ (a unitary matrix) and $x_{\alpha} = x_{0} + \alpha z$, where $x_{0}$ is a solution to $y=Ax$, $\alpha \in \mathbb{R}$, and $z \in \mathcal{N}(A)$ with $\mathcal{N}(A)$ denoting the null space of $A$. Then, as $\alpha \to \infty$ with $b_{j}$ set to $W P_{j} x_{\alpha}$, the objective in (P1) remains always finite (non-coercive). The constraint $\left \| x \right \|_{2}\leq C$ alleviates possible problems (e.g., unbounded iterates in algorithms) due to such a non-coercive objective. It can also be alternatively replaced with constraints such as box constraints depending on the application and underlying image properties.


While a single weight $\eta^{2}$ is used for the sparsity penalties $ \left \| b_{j} \right \|_{0}$ $\forall$ $j$ in (P1), one could also use different weights $\eta_{j}^{2}$ for the penalties corresponding to different patches, if such weights are known, or estimated.
When measurements from multiple images (or frames, or slices) are available, then by considering the summation of the corresponding objective functions for each image, Problem (P1) can be easily extended to enable joint reconstruction of the images using a single adaptive (spatial) transform.
For applications such as dynamic MRI, one can also work with adaptive spatiotemporal sparsifying transforms of 3D patches in (P1).

We have studied some transform BCS methods in very recent works \cite{sabressiims1, syber}. However, the formulation (P1) investigated here was not explored in the prior work. Exploiting both a unitary transform constraint (as opposed to a penalty that enables well-conditioning \cite{sabressiims1})  and a sparsity penalty (as opposed to a sparsity constraint \cite{sabressiims1}) in the transform BCS formulation leads to a very efficient block coordinate descent algorithm in this work. Moreover, Problem (P1) and the algorithm proposed to solve it can be readily extended to accomodate richer models as shown in the following discussions.

\subsection{Union of Transforms BCS} \label{ubcsform}

Here, we extend the single transform model in Problem (P1) to a union of transforms model (similar to \cite{saiwen}).
In this model, we consider a collection (union) of square transforms $ \{W_{k}\}_{k=1}^{K}$  with $W_{k} \in \mathbb{C}^{n \times n} \, \forall \, k $, and each image patch is assumed to have a corresponding ``best matching transform" (i.e., a transform that best sparsifies the particular patch) in this collection.
A motivation for the proposed model is that natural images or image patches need not be sufficiently sparsifiable by a single transform. For example, image patches from different regions of an image usually contain different types of features, or textures. Thus, having a union of transforms would allow groups of patches with common features (or, textures) to be better sparsified by their own specific transform.


Such a union of square transforms can be interpreted as an overcomplete transform, also called OverComplete TransfOrm model with BlOck coSparsity constraint, or OCTOBOS. The equivalent overcomplete transform is obtained by stacking the square `sub-transforms' as $W=\begin{bmatrix}W_{1}^{T}\mid W_{2}^{T}\mid &...& \mid W_{K}^{T}\end{bmatrix}^{T}$. The matrix $ W \in \mathbb{R}^{m \times n}$, with $m = K n $, and thus, $m>n$ (overcomplete transform) for $K>1$.
Proposition 1 of  \cite{saiwen} proves the equivalence between the following two (sparse coding) problems, where the first one involves the union of transforms, and the second one is based directly on an overcomplete (OCTOBOS) one.
\begin{align} 
  & \min_{1 \leq k \leq K}\: \min_{\alpha^{k}}\: \left \| W_{k}z-\alpha^{k} \right \|_{2}^{2} \;\: \text{s.t.}\; \:  \left \| \alpha^{k} \right \|_{0}\leq s\; \forall \, k \label{rege2}\\
  & \;\;\, \min_{\alpha}\: \left \| Wz - \alpha \right \|_{2}^{2} \;\; \text{s.t.}\; \:  \left \| \alpha \right \|_{0,s}\geq 1\;  \label{rege}
\end{align}
Here, $z \in \mathbb{C}^{n}$ is a given signal, and $\alpha \in \mathbb{C}^{m}$ in \eqref{rege} is obtained by stacking $K$ blocks $\alpha^{k} \in \mathbb{C}^{n}$, $1 \leq k \leq K$. The operation $\left \| \alpha \right \|_{0,s} \triangleq \sum_{k=1}^{K}I(\left \| \alpha^{k} \right \|_{0} \leq s) $ with $I(\cdot)$ denoting the indicator function, counts the number of blocks of $\alpha$ with at least $n-s$ zeros (co-sparse blocks), where $s$ is a parameter.
Proposition 1 of \cite{saiwen} showed that the minimum sparsification errors (objectives) in \eqref{rege2} and \eqref{rege} are identical and that the sparse minimizer(s) in \eqref{rege2} (i.e., best/minimizing sparse code(s) over $1 \leq k \leq K$) are simply the block(s) with at least $ n-s$ zeros of the minimizer(s) in \eqref{rege}.



We have investigated the learning of a union of transforms, or OCTOBOS learning, from training data in a recent work \cite{saiwen}. Here, we propose to use the following union of transforms learning regularizer
\begin{align*}
\zeta(x) =  \frac{1}{\nu}  & \min_{\left \{W_{k}, b_{j}, C_{k} \right \}} \:  \sum_{k=1}^{K} \sum_{j \in C_{k}} \begin{Bmatrix}
\left \| W_{k} P_{j}x- b_{j} \right \|_{2}^{2} + \eta^{2}  \left \| b_{j} \right \|_{0}
\end{Bmatrix} \\
& \;\; \mathrm{s.t.}\;\;  W_{k}^{H}W_{k}=I \;\, \forall \, k, \;\, \left\{C_{k}\right\} \in G
\end{align*}
along with the constraint set $\mathcal{S} = \left \{ x \in \mathbb{C}^{p} : \left \| x \right \|_{2}\leq C \right \}$ within Problem \eqref{reginveq1} to arrive at the following union of transforms BCS formulation:
\begin{align}
\nonumber (\mathrm{P2}) & \min_{x, B, \left \{W_{k}, C_{k} \right \}}    \sum_{k=1}^{K} \sum_{j \in C_{k}} \begin{Bmatrix}
\left \| W_{k} P_{j}x- b_{j} \right \|_{2}^{2} + \eta^{2}  \left \| b_{j} \right \|_{0}
\end{Bmatrix}\\
\nonumber & \;\;\;\;\;\;\;\;\; \;\;\;\;\;\;\;\;\;\; \;\;\;\;+ \nu \left \| Ax-y \right \|_{2}^{2}  \\
\nonumber & \;\,\;\;\;\;\;\; \, \mathrm{s.t.}\;\;  W_{k}^{H}W_{k}=I \;\, \forall \, k, \;\, \left\{C_{k}\right\} \in G, \;\; \left \| x \right \|_{2}\leq C.
\end{align}
Here and in the remainder of this work, when certain indexed variables are enclosed within braces, it means that we are considering the set of variables over the range of the indices.
The set $\left\{C_{k}\right\}_{k=1}^{K}$ in (P2) indicates a clustering of the image patches $\left \{ P_{j}x \right \}_{j=1}^{N}$ into $K$ disjoint sets.
The cluster $C_k$ contains the indices $j$ corresponding to the patches $P_{j} x$ in the $k$th cluster. The patches in the $k$th cluster are considered (best) matched to the transform $W_k$. The set $G$ in (P2) is the set of all possible partitions of the set of integers $[1:N]  \triangleq \left \{1, 2, ..., N  \right \}$ into $K$ disjoint subsets, i.e., 
\[ G =  \left \{  \left \{ C_{k}  \right \} :  \bigcup_{k=1 }^{K}C_{k} = [1:N], \, C_{j} \bigcap C_{k} = \emptyset, \, \forall \, j \neq k      \right \} \]
 
The term $\sum_{k=1}^{K} \sum_{j \in C_{k}} \left \| W_{k} P_{j}x- b_{j} \right \|_{2}^{2}$ in (P2) is the sparsification error of the patches of $x$ in the (richer) union of transforms model.
Problem (P2) is to jointly reconstruct the image $x$ and learn the (unknown) union of transforms for the image patches, as well as cluster the patches, using only the compressive imaging measurements. The optimal objective function value (i.e., the minimum value) in Problem (P2) can only be lower than the corresponding optimal value in (P1). This is obvious because the single transform model in (P1) is a subset of the richer (or more general) union of transforms model in (P2).

The recent PANO method \cite{Qu2014843} for MR image reconstruction also involves a patch grouping methodology, but differs from the method proposed here in several important aspects: (i) the patch grouping criterion; (ii) the type and dimension of sparsifying transform; and (iii) the use of a reference reconstruction vs. joint clustering and reconstruction. In particular, in the PANO method, the patches of a reference reconstruction are grouped together according to their similarity measured in terms of the Euclidean $\ell_{2}$ distance. A penalty based on the sparsity of such groups of similar (2D) patches in a \emph{fixed} 3D transform (Haar wavelet) domain is used as a regularizer in the CS image reconstruction problem.
Unlike the PANO method, Problem (P2) clusters together patches that are best sparsified by a common \emph{adaptive} transform, i.e., the clustering measure is based on the sparsification error.  Thus the clustered patches need not be similar in Euclidean distance and the adapted clusterings in (P2) can be quite general.
Furthermore, in (P2), because the transform is adapted to the patches in the cluster, the transform depends on the clustering, and the clustering depends on the transform. Another difference is that, unlike the 3D (fixed) transform in PANO, for 2D patches, the adapted transform here is a 2D transform sparsifying each patch individually. Finally, unlike PANO, (P2) jointly clusters patches and reconstructs $x$, and is not based on reference reconstructions.






\section{Algorithms and Properties}
\label{sec3}  

\subsection{Algorithms}

Problems (P1) and (P2) involve highly nonconvex and non-differentiable (in fact, discontinuous) objectives, as well as nonconvex constraints. Because of the lack of analytical solutions, iterative approaches are commonly adopted for problems of this kind. Here, we adopt iterative block coordinate descent algorithms for (P1) and (P2) that lead to highly efficient solutions for the corresponding subproblems. Another advantage of block coordinate descent is that it does not require the choice of additional parameters such as step sizes. We first describe our algorithm for (P2). The algorithm for (P1) is just a special case (with $K=1$) of the one for (P2).




In one step of our proposed block coordinate descent algorithm for (P2) called the \emph{sparse coding and clustering step}, we solve for $\left\{C_{k}\right\}$ and $B$ in (P2) with the other variables fixed. In another step called the \emph{transform update step}, we solve for the transforms $\left\{W_{k}\right\}$  in (P2), while keeping all other variables fixed. In the third step called the \emph{image update step}, we update only the image $x$, with the other variables fixed. We now describe these steps in detail.

\subsubsection{Sparse Coding and Clustering Step}
In this step, we solve the following optimization problem:
\begin{align}
 \nonumber (\mathrm{P3}) \;\; & \min_{\left \{ C_{k} \right \}, \left \{ b_{j} \right \}}\: \sum _{k=1}^K  \sum_{j \in C_{k}} \begin{Bmatrix}
\left \| W_{k} P_{j}x- b_{j} \right \|_{2}^{2} + \eta^{2}  \left \| b_{j} \right \|_{0}
\end{Bmatrix} \\
\nonumber &   \,\,\,\,\,\,\,\, \mathrm{s.t.}\; \: \left\{C_{k}\right\} \in G.
\end{align} 
By first performing the (inner) optimization with respect to the $b_{j}$'s in (P3), it is easy to observe that Problem (P3) can be rewritten in the following equivalent form:
\begin{align} 
  & \hspace{-0.03in} \sum _{j=1}^{N}  \min_{1 \leq k \leq K}  \begin{Bmatrix}
\left \| W_{k} P_{j}x-   H_{\eta} (W_{k} P_{j}x) \right \|_{2}^{2} + \eta^{2}  \left \|  H_{\eta} (W_{k} P_{j}x)  \right \|_{0}
\end{Bmatrix} \label{bi1}
\end{align} 
where the minimization over $k$ for each patch $P_{j}x$ ($1 \leq j \leq N$) determines the cluster $C_{k}$ in (P3) to which that patch belongs. The hard-thresholding operator $H_{\eta} (\cdot)$ appears in \eqref{bi1} because of the aforementioned (inner) optimization with respect to the $b_{j}$'s \cite{sbclsTS2} in (P3), and $H_{\eta} (\cdot)$  is defined as follows, where $\alpha \in \mathbb{C}^{n}$ is any vector, and the subscript $i$ indexes vector entries.
\begin{equation} \label{bcs5}
 \left ( H_{\eta} (\alpha) \right )_{i}=\left\{\begin{matrix}
 0&, \;\;\left | \alpha_{i} \right | < \eta \\
\alpha_{i}  & ,\;\;\left | \alpha_{i} \right | \geq \eta 
\end{matrix}\right.
\end{equation}
For each patch $P_{j} x$, the optimal cluster index $\hat{k}_{j}$ in \eqref{bi1} is then 
\begin{equation}
\hat{k}_{j} = \underset{k}{\arg \min}\, \left \| W_{k} P_{j}x-   H_{\eta} (W_{k} P_{j}x) \right \|_{2}^{2} + \eta^{2}  \left \|  H_{\eta} (W_{k} P_{j}x)  \right \|_{0}
\label{spie1}
\end{equation}
The optimal sparse code $\hat{b}_{j}$ in (P3) is then $H_{\eta} (W_{\hat{k}_{j}} P_{j}x) $ \cite{sbclsTS2}.
There is no coupling between the sparse coding/clustering problems in \eqref{bi1} for the different image patches $\left \{ P_{j}x \right \}_{j=1}^{N}$. Thus, they are clustered and sparse coded in parallel. 

The optimal cluster membership or the optimal sparse code for any particular patch $P_{j}x$ in (P3) need not be unique. When there are multiple optimal cluster indices in \eqref{spie1}, we pick the lowest such index.
The optimal sparse code for the patch $P_{j}x$ is not unique when the condition $\left | \begin{pmatrix}
W_{\hat{k}_{j}} P_{j}x
\end{pmatrix}_{i} \right | = \eta$ is satisfied for some $i$ (cf. \cite{sbclsTS2} for a similar scenario and an explanation).
The definition in \eqref{bcs5} chooses \emph{one} of the multiple optimal solutions (corresponding to the transform $W_{\hat{k}_{j}}$) in this case.

Note that if instead of employing a sparsity penalty (i.e., penalizing $\sum_{j=1}^{N}  \left \| b_{j} \right \|_{0}$), we were to constrain the term $\sum_{j=1}^{N}  \left \| b_{j} \right \|_{0}$ (i.e., force it to have an upper bound of $s$ \cite{sabressiims1}) in (P2), then the sparse coding and clustering step of such a modified BCS problem shown below suffers from the drawback that the sparsity constraint creates inter-patch coupling, which in turn leads to exponential scaling of the computation with the number of patches.
\begin{align}
\nonumber & \min_{\left \{ C_{k} \right \}, B}\: \sum _{k=1}^K  \sum_{j \in C_{k}} \left \| W_{k} P_{j}x- b_{j} \right \|_{2}^{2}   \\
 &   \,\,\,\,\, \mathrm{s.t.}\; \: \sum_{j=1}^{N} \left \| b_{j} \right \|_{0} \leq s,\, \left\{C_{k}\right\} \in G. \label{alterbcs3}
\end{align} 
For a fixed clustering $\left \{ C_{k} \right \}$, the optimal $B$ above is readily obtained by zeroing out all but the $s$ largest magnitude elements of the matrix $\begin{bmatrix}
 W_{k_1} P_{1}x \mid W_{k_2} P_{2}x \mid ... \mid W_{k_N} P_{N}x
\end{bmatrix}
$, where $k_j$ denotes the cluster index of patch $P_{j}x$. However, this requires examination of all the sparsified patches \emph{jointly}. Now, consider the objective value attained in  \eqref{alterbcs3}  for the clustering $\left \{ C_{k} \right \}$ and its corresponding optimal $B$, to which we refer as the sparsification error for that clustering. The exact solution to Problem \eqref{alterbcs3} requires computing this sparsification error for each possible clustering, and then picking the clustering that achieves the minimum error. Because there are $K^{N}$ possible clusterings, the cost of computing the solution scales exponentially with the number of patches as $O(N n^{2} K^{N})$. Thus,  Problem \eqref{alterbcs3} is computationally intractable. \footnote{One way to modify Problem \eqref{alterbcs3} is to set a constraint of the form $ \left \| b_{j} \right \|_{0} \leq s$ for each patch. In this case, the sparse coding and clustering problem has a cheap solution \cite{saiwen}. However, different regions of natural images typically carry different amounts of information, and therefore, a fixed sparsity level for each patch often does not work well in practice. In contrast, both Problems \eqref{alterbcs3} and (P3) encourage variable sparsity levels for individual patches.}
This is one of the reasons for pursuing formulations with sparsity penalties (rather than constraint) in this work. Furthermore, employing a sparsity penalty leads to a simpler sparse coding solution (with a given clustering) involving hard-thresholding, whereas using a sparsity constraint (as in \eqref{alterbcs3}) for sparse coding necessitates projections onto the  $s$-$\ell_0$ ball \cite{sabressiims1} using a computationally more expensive sorting procedure.


\subsubsection{Transform Update Step}

In this step, we solve (P2) with respect to the cluster transforms $\left\{W_{k}\right\}$, with all other variables fixed. This results in the following optimization problem:
\begin{align}
\nonumber  & \min_{\left \{W_{k} \right \}}    \sum_{k=1}^{K} \sum_{j \in C_{k}} \begin{Bmatrix}
\left \| W_{k} P_{j}x- b_{j} \right \|_{2}^{2} + \eta^{2}  \left \| b_{j} \right \|_{0}
\end{Bmatrix}\\
 & \; \, \mathrm{s.t.}\;\;  W_{k}^{H}W_{k}=I \;\, \forall \, k. \label{tuovprob1}
\end{align}
The above problem is in fact separable (since the objective is in summation form) into $K$ independent constrained optimization problems, each involving a particular square transform $W_{k}$. The $k$th ($1 \leq k \leq K$) such optimization problem is as follows:
\begin{align}
\nonumber (\mathrm{P4})\:\:\: & \min_{W_{k}}  \sum_{j \in C_{k}} \left \| W_{k} P_{j}x- b_{j} \right \|_{2}^{2} \;\; \mathrm{s.t.}\;\;  W_{k}^{H}W_{k}=I. 
\end{align}
Denoting by $X_{C_{k}}$, the matrix that has the patches $P_{j}x$ for $j \in C_{k}$, as its columns, and denoting by $B_{C_{k}}$, the matrix whose columns are the corresponding sparse codes, Problem (P4) can be written in compact form as
\begin{equation} \label{bcs8}
 \min_{W_{k}}\:   \left \| W_{k} X_{C_{k}} - B_{C_{k}} \right \|_{F}^{2}  \; \,\, \mathrm{s.t.}\;\; W_{k}^{H}W_{k}=I.
\end{equation}
 Now, let $X_{C_{k}} B_{C_{k}}^{H}$ have a full singular value decomposition (SVD) of $U \Sigma V^{H}$. Then, a global minimizer \cite{sabres3, sbclsTS2} in \eqref{bcs8} is $\hat{W}_{k}= VU^{H}$. This solution is unique if and only if  $X_{C_{k}} B_{C_{k}}^{H}$  is non-singular.
To solve Problem \eqref{tuovprob1}, Problem (P4) is solved for each $k$, which can be done in parallel.

\subsubsection{Image Update Step}

In this step, we solve (P2) with respect to the unknown image $x$, keeping the other variables fixed. The corresponding optimization problem is as follows.
\begin{align}
\nonumber (\mathrm{P5})\:\:\: & \min_{x}  \:  \nu \left \| Ax-y \right \|_{2}^{2}  + \sum_{k=1}^{K} \sum_{j \in C_{k}} \left \| W_{k} P_{j}x- b_{j} \right \|_{2}^{2} \\
\nonumber & \;\; \, \mathrm{s.t.}\;\; \left \| x \right \|_{2}\leq C.
\end{align}
Problem (P5) is a least squares problem with an $\ell_{2}$ (or, alternatively squared $\ell_{2}$) constraint \cite{golkah1}.
It can be solved for example using the projected gradient method, or using the Lagrange multiplier method \cite{golkah1}.
In the latter case, the corresponding Lagrangian formulation is simply a least squares problem.
Therefore, the solution to (P5) satisfies the following Normal Equation
\begin{align} 
\nonumber & \left (  \sum_{j=1}^{N} P_{j}^{T} P_{j} \; +\;\nu\: A^{H}A + \hat{\mu} I \right )x=  \sum_{k=1}^{K} \sum_{j \in C_{k}} P_{j}^{T}W_{k}^{H} b_{j}\; \\
& \;\;\;\;\;\;\;\;\;\;\;\;\;\;\;\;\;\;\;\;\;\;\;\;\;\;\;\;\;\;\;\;\;\;\;\; +\:\nu \: A^{H}y \label{bcs10}
\end{align}
where $\hat{\mu} \geq 0$ is the optimally chosen Lagrange multiplier. The optimal $\hat{\mu}$ is the smallest non-negative real for which the solution\footnote{The solution in \eqref{bcs10} (for any $\hat{\mu} \geq 0$) is unique if the set of patches in our formulation covers all pixels in the image. This is because $\sum_{j=1}^{N} P_{j}^{T} P_{j}$ is a positive definite diagonal matrix in this case.} (i.e., the $x$) in \eqref{bcs10} satisfies the norm constraint in (P5). Problem (P5) can be solved by solving the Lagrangian least squares problem (or, corresponding normal equation) repeatedly (by CG) for various multiplier values (tuned in steps) until the $\left \| x \right \|_{2}\leq C$ condition is satisfied.


\begin{figure*}[!t]
\begin{tabular}{p{17.8cm}}
\hline
Union of Transforms-Based BCS Algorithm A2 for (P2) for MRI\\
\hline
\textbf{Inputs:} \: $ y $ - CS measurements, $\eta$ - weight, $\nu$ - weight, $C$ - bound on $\left \| x \right \|_{2}$, $J$ - number of iterations.\\
 \textbf{Outputs:} \: $ x $  - reconstructed image,  $\left \{W_{k} \right \}$ - adapted union of transforms,  $\left \{ C_{k} \right \}$ - learnt clustering of patches, $B$ - matrix with sparse codes of patches as columns. \\
\textbf{Initial Estimates:} \: $x^{0}, \left \{W_{k}^{0}, C_{k}^{0} \right \}, B^{0}$.\\
\textbf{For $ t = 1 : J$ Repeat}\\
\vspace{-0.1in}
\begin{enumerate} 
\item \textbf{Transform Update Step:} \textbf{For $k = 1: K$ do}
\begin{enumerate}
\item Form the matrices $Q_{k}$ and $R_{k}$ with $P_{j}x^{t-1}$ and $b_{j}^{t-1}$, for $j \in C_{k}^{t-1}$, as their columns, respectively. 
\item Set $U \Sigma V^{H}$ as the full SVD of $Q_{k} R_{k}^{H}$. $W_{k}^{t} = VU^{H}$.
\end{enumerate}
\item \textbf{Sparse Coding and Clustering Step:} \textbf{For $j = 1: N$ do}
\begin{enumerate}
\item If $j=1$, set $C_{k}^{t} = \emptyset$  $\forall$ $k$.
\item Compute $\gamma_{k} =  \left \| W_{k}^{t} P_{j}x^{t-1}-   H_{\eta} \begin{pmatrix}
W_{k}^{t} P_{j}x^{t-1}
\end{pmatrix} \right \|_{2}^{2} + \eta^{2}  \left \|  H_{\eta} \begin{pmatrix}
W_{k}^{t} P_{j}x^{t-1}
\end{pmatrix}  \right \|_{0}$, $1 \leq k \leq K$. \newline
Set $\hat{k} = \min \left \{ k : \gamma_{k} = \min_{k} \gamma_{k} \right \}$. Set $C_{\hat{k}}^{t}  \leftarrow  C_{\hat{k}}^{t} \cup \left \{ j \right \}$.
\item $b_{j}^{t} = H_{\eta} \begin{pmatrix}
W_{\hat{k}}^{t} P_{j}x^{t-1}
\end{pmatrix} $. 
\end{enumerate}
\item \textbf{Image Update Step:}
\begin{enumerate} 
\item Compute the image $c = \sum_{k=1}^{K} \sum_{j \in C_{k}^{t}} P_{j}^{T}\begin{pmatrix}
W_{k}^{t}
\end{pmatrix}^{H} b_{j}^{t}$. $S \leftarrow \mathrm{FFT}(c)$.
\item Compute $f(0)$ as per \eqref{bcs13b}. If $f(0) \leq C^{2}$, set $\hat{\mu}=0$. Else, use Newton's method to solve the equation $f(\hat{\mu}) = C^{2}$ for $\hat{\mu}$.
\item Update $S$ to be the right hand side of \eqref{bcs13}. $x^{t} = \mathrm{IFFT}(S)$.
\end{enumerate}
\end{enumerate}
\textbf{End} \\
\hline
\end{tabular}
\caption{Algorithm for (P2). The superscript $t$ denotes the iterates in the algorithm. The encoding matrix $F$ in (single-coil) MRI is assumed normalized and the abbreviations FFT and IFFT denote the fast implementations of the normalized 2D DFT and 2D IDFT, respectively. The algorithm for (P1) is identical to the one above except that there is no clustering involved in the sparse coding and clustering step.} \label{im5pbcs}
\vspace{-0.15in}
\end{figure*}

We now discuss the solution to (P5) for the specific case of single-coil MRI. (In the case of multi-coil or parallel MRI, when $A$ is for example a SENSE type sensing matrix \cite{swang1}, the aforementioned iterative strategies can be used to solve (P5).)
Recall that $A= F_{u}$ for (single-coil) MRI, and we assume that the k-space measurements are obtained by subsampling on a uniform Cartesian grid.
Assuming that periodically positioned, overlapping image patches (patch \emph{overlap stride} \cite{bresai} denoted by $r$) are used in our formulations, and that the patches that overlap the image boundaries `wrap around' on the opposite side of the image \cite{bresai}, we have that the matrix $\sum_{j=1}^{N} P_{j}^{T} P_{j} = \beta I$, with $\beta = \frac{n}{r^{2}}$. Then, equation \eqref{bcs10} simplifies for MRI as
\begin{align} 
\nonumber & \left (\beta I +\nu\: F F_{u}^{H}F_{u}F^{H} + \hat{\mu} I \right )Fx = F \sum_{k=1}^{K} \sum_{j \in C_{k}} P_{j}^{T}W_{k}^{H} b_{j} \\
& \;\;\;\;\;\;\;\;\;\;\;\;\;\;\;\;\;\;\;\;\;\;\;\;\;\;\;\;\;\;\;\;\;\;\;\; + \nu FF_{u}^{H}y \label{bcs12b}
\end{align}
where $ F  \in \mathbb{C}^{p \times p} $ denotes the full Fourier encoding matrix assumed normalized ($ F^{H}F = I$), and $ FF_{u}^{H}F_{u}F^{H} $ is a diagonal matrix of ones and zeros, with the ones at those entries that correspond to sampled locations in k-space.


Denote $S \triangleq  F \sum_{k=1}^{K} \sum_{j \in C_{k}} P_{j}^{T}W_{k}^{H} b_{j} $ and $ S_{0}\triangleq FF_{u}^{H}y $. $S_0$ represents the undersampled k-space measurements expanded to full (matrix) size, by inserting zeros at non-sampled locations. The solution to \eqref{bcs10} for single-coil MRI, in Fourier space, is then
\begin{equation}\label{bcs13}
Fx_{\hat{\mu}} \:(k_{x},k_{y})=\left\{\begin{matrix}
\frac{S(k_{x},k_{y})}{\beta + \hat{\mu}} &,\:(k_{x},k_{y})\notin \Omega  \\
\frac{S(k_{x},k_{y})+\nu\, S_{0}(k_{x},k_{y})}{\beta+\nu + \hat{\mu}}   & ,\:(k_{x},k_{y})\in \Omega
\end{matrix}\right.
\end{equation}
where $(k_{x},k_{y})$ indexes k-space locations, and $ \Omega $ is the subset of k-space that is sampled.
Note that the optimal Lagrange multiplier $\hat{\mu}$ is the smallest non-negative real such that
\begin{align}
\nonumber f(\hat{\mu}) \triangleq \left \| x_{\hat{\mu}} \right \|_{2}^{2} & = \sum_{(k_{x},k_{y})\notin \Omega} \frac{\left | S(k_{x},k_{y}) \right |^{2}}{\left ( \beta + \hat{\mu} \right )^{2}} \\
&+ \sum_{(k_{x},k_{y})\in \Omega} \frac{\left | S(k_{x},k_{y})+\nu\, S_{0}(k_{x},k_{y}) \right |^{2}}{\left ( \beta + \nu + \hat{\mu} \right )^{2}} \leq C^{2}  \label{bcs13b}
\end{align}
We check if the above condition is satisfied for $\hat{\mu}=0$  first. If not, then we apply Newton's method to find the optimal $\hat{\mu}$ in $f(\hat{\mu}) = C^{2}$.
The optimal $\hat{x}$ in (P5) for MRI is the 2D inverse FFT of the optimal $ Fx_{\hat{\mu}} $ in \eqref{bcs13}.

The unitary property of the transforms $W_{k}$ leads to efficient solutions in the image update step for MRI. In particular, if the $W_{k}$'s were not unitary, the matrix $ \sum_{j=1}^{N} P_{j}^{T} P_{j}$ in \eqref{bcs10} and later equations would be replaced with $\sum_{k=1}^{K} \sum_{j \in C_{k}} P_{j}^{T}W_{k}^{H}W_{k} P_{j}$. The latter matrix is neither diagonal nor readily diagonalizable. Hence, we cannot exploit the simple closed-form solution in \eqref{bcs13} in this case and would have to employ slower iterative solution techniques for MRI.


The overall algorithm corresponding to the BCS Problem (P2) is shown in Fig. \ref{im5pbcs}.
The algorithm begins with an initial estimate $x^{0}, \left \{W_{k}^{0}, C_{k}^{0} \right \}, B^{0}$  (e.g., $x^0 = A^{\dagger} y$ (assuming $\left \| A^{\dagger} y \right \|_{2} \leq C$), a random or k-means clustering initialization $\left \{C_{k}^{0} \right \}$,  $W_{k}^{0} = \mathrm{2D \, DCT}$ $\forall$ $k$, and $B^0$ set to be the minimizer of (P2) for these $x^{0}, \left \{W_{k}^{0}, C_{k}^{0} \right \}$).
Each outer iteration of the algorithm involves the sparse coding and clustering, transform update, and image update steps. (In general, one could alternate between some of these steps more frequently than between others.)
Our algorithm for solving Problem (P1) is similar to that for Problem (P2), except that we work with a single cluster ($K=1$) in the former case. In particular, the sparse coding and clustering step for (P2) is replaced by just a sparse coding step (in a single unitary transform) for (P1).

\subsection{Computational Costs} \label{compscost}

Here, we briefly analyze the computational costs of our algorithms for Problems (P1) and (P2), called Algorithm A1 and A2, respectively.


For a fixed number of clusters (constant $K$), the computational cost per iteration of Algorithm A2 for (P2) for MRI scales as $O(n^{2} N)$. Thus, the cost scales quadratically with the parameter $n$ (number of pixels in a patch) and linearly with $N$ (number of patches). The cost per iteration of Algorithm A1 for (P1) scales similarly with respect to these parameters. These costs are dominated by the computations for matrix-matrix or matrix-vector products in our algorithms.
In contrast, overcomplete dictionary-based BCS methods such as DLMRI \cite{bresai} that learn a dictionary $D \in \mathbb{C}^{n \times m}$  ($m \geq n$) from compressive measurements have a cost per outer iteration that scales as $ O(n m s N \hat{J}) $ \cite{bresai, sabressiims1}, where $s$ is the synthesis sparsity level per patch, and $\hat{J}$ is the number of inner dictionary learning (K-SVD \cite{elad}) iterations in DLMRI. The DLMRI cost is dominated by synthesis sparse coding (an NP-hard problem). Assuming $m \propto n$ and $s \propto n$, the cost per iteration of DLMRI scales as $O(n^{3} N \hat{J})$.
Thus, the per-iteration cost of Algorithm A1 or A2 scales much better with patch size than that for prior synthesis dictionary-based BCS methods. This would be particularly advantageous in the context of higher-dimensional imaging applications such as 3D or 4D imaging, where the corresponding 3D or 4D patches are much bigger than the patches in 2D imaging. As illustrated in Section \ref{sec4}, the proposed algorithms tend to converge quickly in practice. Therefore, the per-iteration computational advantages usually translate to a net computational advantage in practice.


Clearly the union of transforms based Algorithm A2 involves more computations/operations than the single transform based Algorithm A1. As the number of clusters $K$ varies, the computational cost per iteration of Algorithm A2 for MRI scales as $O(Kn^{2} N)$ (i.e., the cost scales linearly with the number of clusters). In particular, these computations (with respect to parameter $K$) are dominated by the clustering step, where the product between each $W_{k}$ ($1 \leq k \leq K$) and every patch needs to be computed (to determine the optimal matching transforms or clusters). The computations in Algorithm A2 can be reduced by performing the clustering step less often (than the sparse coding, transform update, and image update steps) in the block coordinate descent algorithm.


\section{Convergence Properties} \label{sec40}

Since (P1) and (P2) are highly non-convex, standard results on convegence of block coordinate descent methods \cite{tseng6} do not apply. In fact, in certain scenarios, one can easily construct non-convergent iterate sequences for Algorithm A1 or A2 (cf. Section 4 of \cite{sabressiims1} for examples of such scenarios for related algorithms). Here, we present convergence results for Algorithms A1 and A2 assuming that the various steps (such as SVD computations) are performed exactly.
Each outer iteration of our algorithms involves a transform update step, a sparse coding and clustering step (only sparse coding in the case of (P1)), and an image update step. 

\subsection{Notations}

Problem (P1) is a constrained and non-convex minimization problem. By replacing each constraint with an equivalent barrier penalty (a function that takes the value $+\infty$ when the constraint is violated, and is zero otherwise), Problem (P1) can be written in an unconstrained form involving the following objective function:
\begin{align}
g(W, B, x)  & =  \nu \left \| Ax-y \right \|_{2}^{2}  + \varphi(W)  +  \chi(x) \\
\nonumber &  \;\;\; +  \sum_{j=1}^{N} \begin{Bmatrix}
\left \| W P_{j}x- b_{j} \right \|_{2}^{2} + \eta^{2}  \left \| b_{j} \right \|_{0}
\end{Bmatrix}
\end{align}
where $\varphi(W)$ and $\chi(x)$ are the barrier penalties corresponding to the unitary transform constraint and energy constraint on $x$, respectively.  Problem (P2) can also be written in the following unconstrained form:
\begin{align}
 h(W, B,\Gamma, x) & =  \nu \left \| Ax-y \right \|_{2}^{2}  +  \chi(x) + \sum_{k=1}^{K} \varphi(W_{k}) \label{ubcsfn}\\
\nonumber &  \;\;\; + \sum_{k=1}^{K} \sum_{j \in C_{k}} \begin{Bmatrix}
\left \| W_{k} P_{j}x- b_{j} \right \|_{2}^{2} + \eta^{2}  \left \| b_{j} \right \|_{0}
\end{Bmatrix}
\end{align}
where $\varphi(W_{k})$ is the barrier penalty corresponding to the unitary constraint on $W_{k}$, $W \in \mathbb{C}^{Kn \times n}$ is obtained by stacking the $W_{k}$'s on top of one another (or, equivalent OCTOBOS), and the row vector $\Gamma \in \mathbb{R}^{1 \times N}$ is such that its $j$th element $\Gamma_{j} \in \left \{ 1, .., K \right \}$ denotes the cluster index (label) corresponding to the patch $P_{j}x$. As discussed previously, the clusters $\begin{Bmatrix}
C_{k}
\end{Bmatrix}$ partition $[1:N]$. Here, we refer to patch cluster memberships using the row vector variable $\Gamma$ rather than using the $C_{k}$'s.

We denote the iterates (outputs) in each iteration  $t$ of Algorithm A1  by the set  $\left ( W^{t}, B^{t}, x^{t} \right )$. For Algorithm A2, the iterates are denoted by the set $(W^{t}, B^{t}, \Gamma^{t}, x^{t})$, where $W^{t}$ in this case denotes the matrix obtained by stacking the cluster-specific transforms $W_{k}^{t}$ ($1 \leq k \leq K$), and $\Gamma^{t}$ is a row vector containing the patch cluster indices $\Gamma_{j}^{t}$ ($1 \leq j \leq N$) as its elements.


\subsection{Main Results}

For Algorithm A1 proposed for Problem (P1), the convergence results take the same form as those for similar schemes presented in a very recent work \cite{sabressiims1}. The result for (P1) is summarized in the following Theorem and corollaries, where for a matrix $H$, $\left \| H \right \|_{\infty} \triangleq \max_{i,j} \left | H_{ij} \right |$, and by `globally convergent', we mean convergence from any initialization. 

\newtheorem{theorem}{Theorem}
\newtheorem{corollary}{Corollary}
\newtheorem{lemma}{Lemma}
\newtheorem{proposition}{Proposition}

\begin{theorem}\label{theorem1bc}
For an initial $(W^{0}, B^{0}, x^{0})$,  the objective sequence $\left \{ g^{t} \right \}$ in Algorithm A1 with $g^{t} \triangleq g\left ( W^{t}, B^{t}, x^{t}  \right )$ is monotone decreasing, and converges to a finite value, say $g^{*} = g^{*}(W^{0}, B^{0}, x^{0})$.
Moreover, the bounded iterate sequence  $\left \{ W^{t}, B^{t}, x^{t} \right \}$ is such that all its accumulation points are equivalent and achieve the same value $g^{*}$ of the objective. The sequence $\left \{ a^{t} \right \}$ with $a^{t} \triangleq \left \| x^{t} - x^{t-1} \right \|_{2}$, converges to zero.
Every accumulation point $(W, B, x)$ of $\left \{ W^{t}, B^{t}, x^{t} \right \}$ is a critical point \cite{sabressiims1, vari1} of the objective $g$ satisfying the following partial global optimality conditions
\begin{align} 
x \in & \underset{\tilde{x}}{\arg\min} \; \,  g\left (W, B, \tilde{x} \right ) \label{cnbcs4}\\
W \in & \underset{\tilde{W}}{\arg\min} \; \,  g\left (\tilde{W}, B, x \right ) \label{cnbcs5}\\
B \in & \underset{\tilde{B}}{\arg\min} \; \,  g\left (W, \tilde{B}, x \right ) \label{cnbcs6}
\end{align}
Each $(W, B, x)$ also satisfies the following partial local optimality condition that holds for all $\Delta x \in \mathbb{C}^{p}$, and all $\Delta B \in \mathbb{C}^{n \times N}$ satisfying $\left \| \Delta B \right \|_{\infty} < \eta/2$:
\begin{equation} 
g(W, B + \Delta B, x + \Delta x) \geq  g(W, B, x) = g^{*} \label{cnbcs5b}
\end{equation}
\end{theorem}

\begin{corollary}\label{corollary1a}
For each $(W^{0}, B^{0}, x^{0})$, the iterate sequence in Algorithm A1 converges to an equivalence class of critical points that are also partial minimizers satisfying \eqref{cnbcs4}, \eqref{cnbcs5}, \eqref{cnbcs6}, and \eqref{cnbcs5b}.
\end{corollary}

\begin{corollary}\label{corollary1b}
Algorithm A1 is globally convergent to a subset of the set of critical points of the non-convex objective $g\left (W, B, x \right )$. The subset includes all critical points $(W, B, x)$, that are at least partial global minimizers of $g(W, B, x)$ with respect to each of $W$, $B$, and $x$, and partial local minimizers of $g(W, B, x)$ with respect to $(B, x)$. 
\end{corollary}

Theorem \ref{theorem1bc} establishes that for each initial $(W^{0}, B^{0}, x^{0})$, the iterate sequence in Algorithm A1 converges to an equivalence class of accumulation points (corresponding to the same objective value $g^{*} = g^{*}(W^{0}, B^{0}, x^{0})$ -- that could vary with initialization). The equivalent accumulation points are all critical points (generalized stationary points \cite{vari1}) and at least partial minimizers of the objective $g$. 


In the case of  the Algorithm A2 proposed for (P2), because the cluster memberships are discrete rather than continuous variables, we do not have a critical points \cite{sabressiims1, vari1} property as for Algorithm A1. Instead, we establish the following convergence results for Algorithm A2. 


\begin{theorem}\label{theorem2bc}
For an initial $(W^{0}, B^{0}, \Gamma^{0}, x^{0})$, the objective sequence $\left \{ h^{t} \right \}$ in Algorithm A2 with $h^{t} \triangleq h\left ( W^{t}, B^{t}, \Gamma^{t}, x^{t}  \right )$ is monotone decreasing, and converges to a finite value, say $h^{*} = h^{*}(W^{0}, B^{0}, \Gamma^{0}, x^{0})$.
Moreover, the iterate sequence  $\left \{ W^{t}, B^{t}, \Gamma^{t}, x^{t} \right \}$ is bounded, and all its accumulation points are equivalent and achieve the same value $h^{*}$ of the objective. The sequence $\left \{ a^{t} \right \}$ with $a^{t} \triangleq \left \| x^{t} - x^{t-1} \right \|_{2}$, converges to zero.
Every accumulation point $(W, B, \Gamma, x)$ of $\left \{ W^{t}, B^{t}, \Gamma^{t}, x^{t} \right \}$ satisfies the following partial global optimality conditions
\begin{align} 
x \in & \underset{\tilde{x}}{\arg\min} \; \,  h\left (W, B,  \Gamma, \tilde{x} \right ) \label{cnbcs4un}\\
W \in & \underset{\tilde{W}}{\arg\min} \; \, h\left (\tilde{W}, B,  \Gamma, x \right ) \label{cnbcs5un}\\
(B, \Gamma) \in & \underset{\tilde{B}, \tilde{\Gamma}}{\arg\min} \; \,  h\left (W, \tilde{B}, \tilde{\Gamma}, x \right ) \label{cnbcs6un}
\end{align}
Each $(W, B, \Gamma, x)$ also satisfies the following partial local optimality condition that holds for all $\Delta x \in \mathbb{C}^{p}$, and all $\Delta B \in \mathbb{C}^{n \times N}$ satisfying $\left \| \Delta B \right \|_{\infty} < \eta/2$:
\begin{align} 
h(W, B + \Delta B, \Gamma, x + \Delta x) \geq & h(W, B,\Gamma, x) = h^{*} \label{cnbcs5bun}
\end{align}
\end{theorem}

Theorem \ref{theorem2bc} establishes that for each initial $(W^{0}, B^{0}, \Gamma^{0}, x^{0})$, the iterate sequence in Algorithm A2 converges to an equivalence class of accumulation points. The equivalent accumulation points are at least partial minimizers of the objective $h$.
In light of Theorem \ref{theorem2bc}, results similar to Corollaries \ref{corollary1a} and \ref{corollary1b} apply for Algorithm A2, as follows.

\begin{corollary}\label{corollary2a}
For each $(W^{0}, B^{0}, \Gamma^{0}, x^{0})$, the iterate sequence in Algorithm A2 converges to an equivalence class of accumulation points that are also partial minimizers satisfying \eqref{cnbcs4un}, \eqref{cnbcs5un}, \eqref{cnbcs6un}, and \eqref{cnbcs5bun}.
\end{corollary}

\begin{corollary}\label{corollary2b}
The iterate sequence in Algorithm A2 is globally convergent to the set of partial minimizers of the non-convex objective $h\left ( W, B, \Gamma, x  \right )$.
The set includes all points $(W, B, \Gamma, x)$, that are at least partial global minimizers of $h(W, B, \Gamma, x)$ with respect to each of $W$, $(B, \Gamma)$, $x$, and partial local minimizers of $h(W, B, \Gamma, x)$ with respect to $(B, x)$. 
\end{corollary}

Notice that for both Algorithms A1 and A2, $\left \| x^{t} - x^{t-1} \right \|_{2} \to 0$. This is a necessary but not sufficient condition for convergence of the entire sequence $\left \{ x^{t} \right \}$.
In general, the set of points to which Algorithm A2 or A1 converges may be larger than the set of global minimizers in Problem (P2) or (P1). We leave the investigation of the conditions under which the proposed algorithms converge to the set of global minimizers of the proposed problems to future work.

A brief proof of Theorem \ref{theorem2bc} is included in the supplementary material available online \cite{suppl123}. The proof draws on a few results from recent works  \cite{sabressiims1, sbclsTS2}, but is for the more complex union of transforms-based blind compressed sensing scenario.
Since Algorithm A1 for (P1) is simply a special case (with $K=1$) of Algorithm A2 for (P2), we do not provide a separate proof for Theorem~\ref{theorem1bc}.


\section{NUMERICAL EXPERIMENTS}
\label{sec4}

\subsection{Framework} \label{sec4framwork}

\begin{figure}[!t]
\begin{center}
\begin{tabular}{cc}
\includegraphics[height=1.2in]{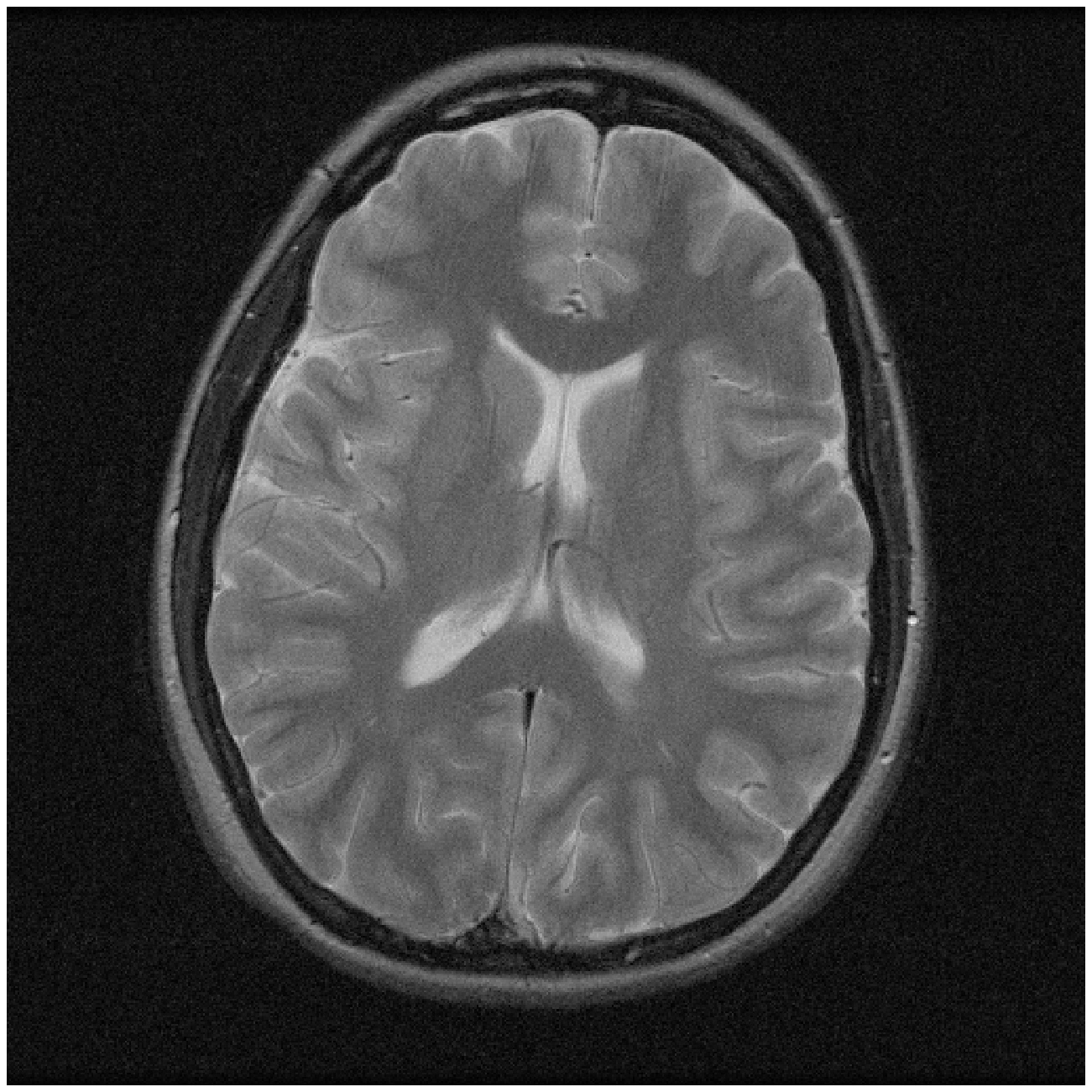}&
\includegraphics[height=1.2in]{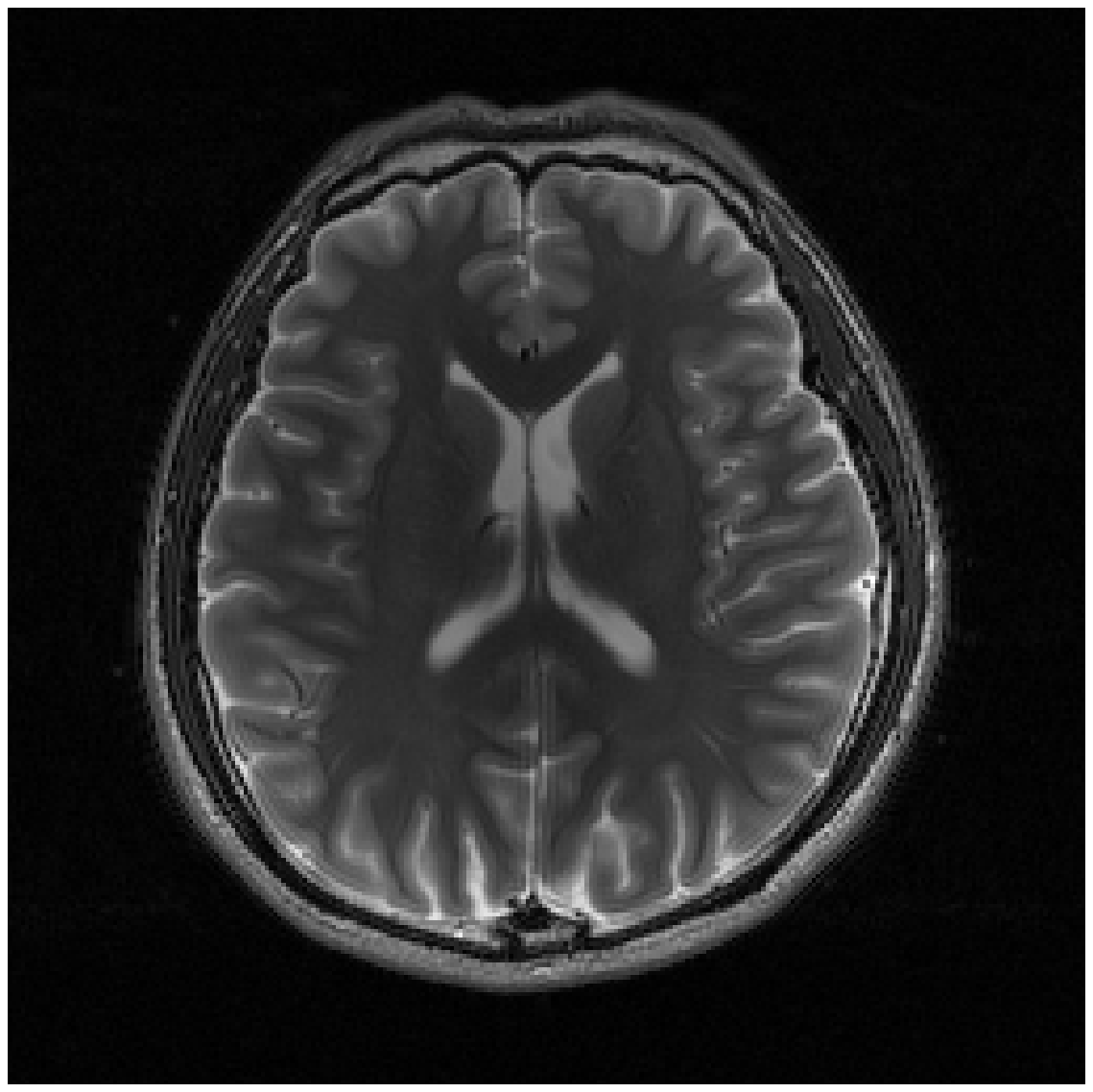} \\
(a) & (b) \\
\end{tabular}
\begin{tabular}{cc}
\includegraphics[height=1.2in]{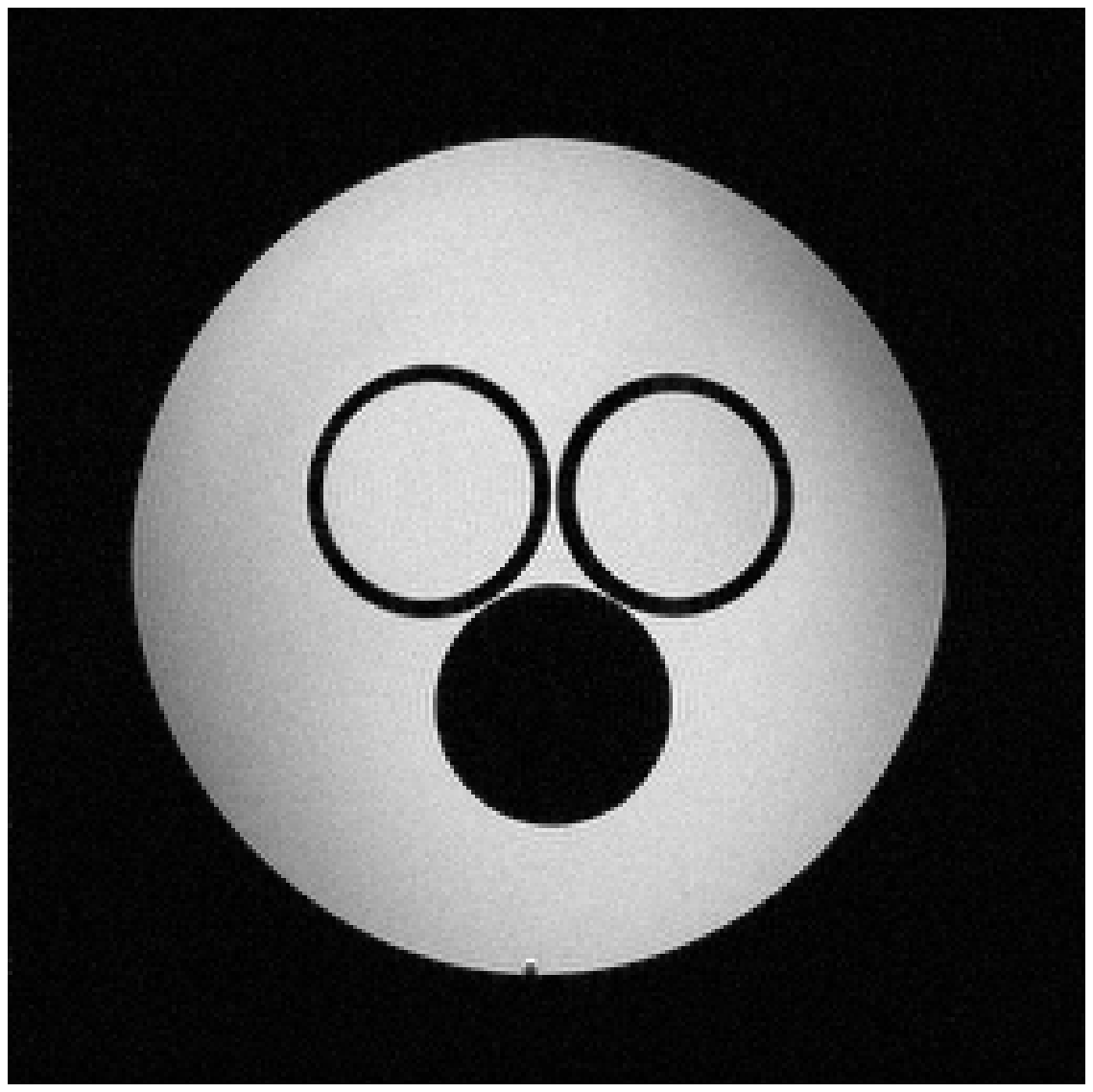}  &
\includegraphics[height=1.2in]{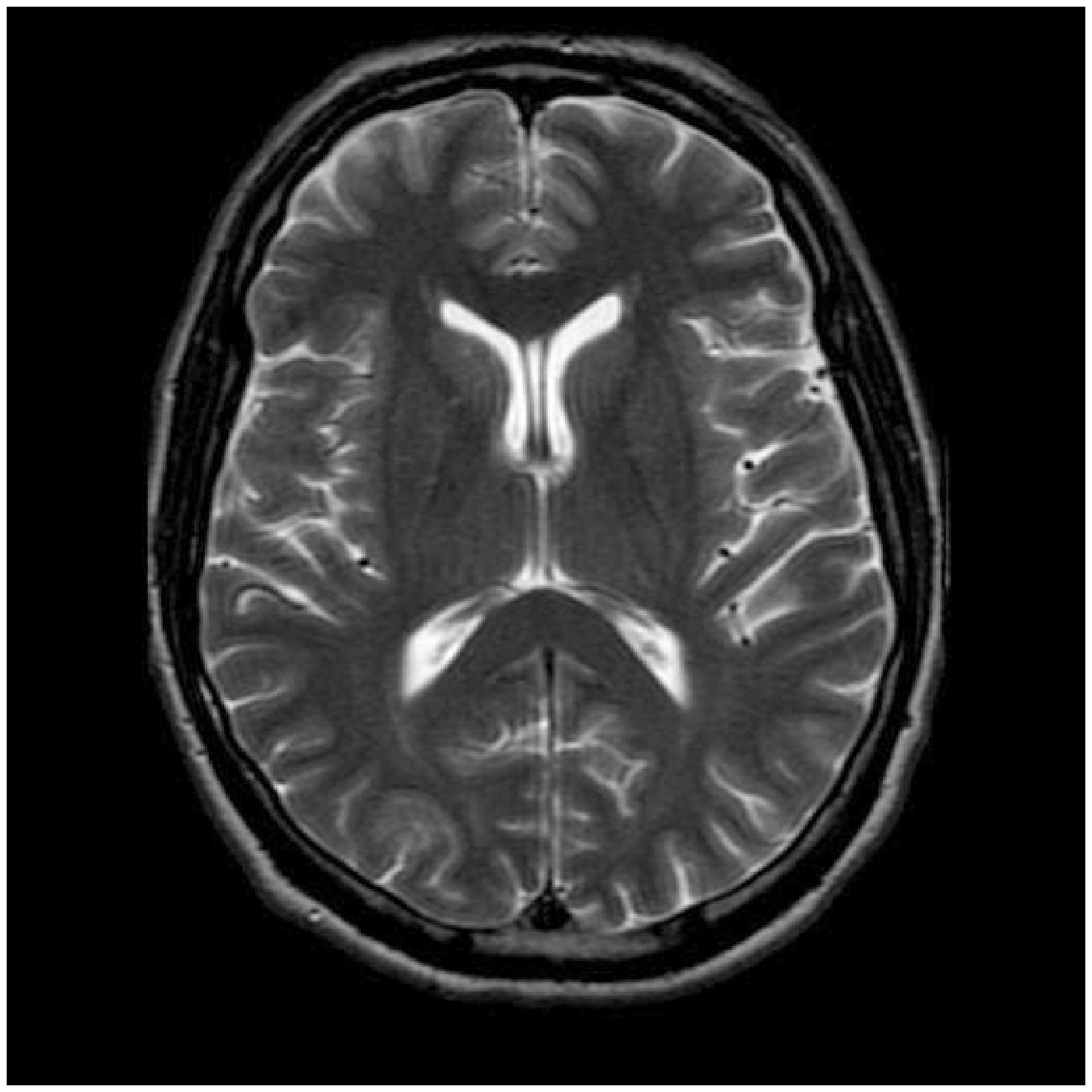}  \\
(c) &  (d)  \\
\includegraphics[height=1.2in]{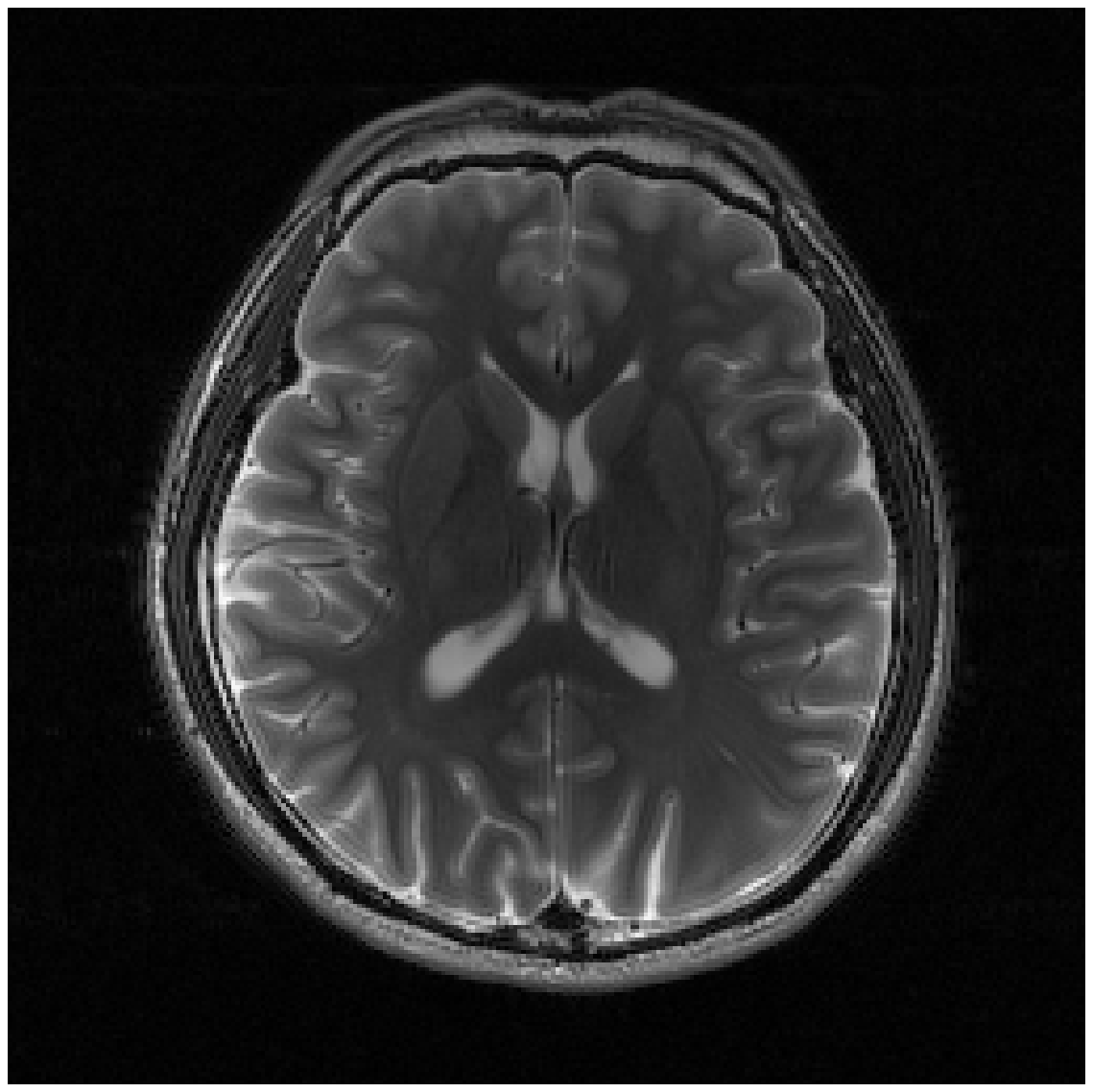} &
\includegraphics[height=1.2in]{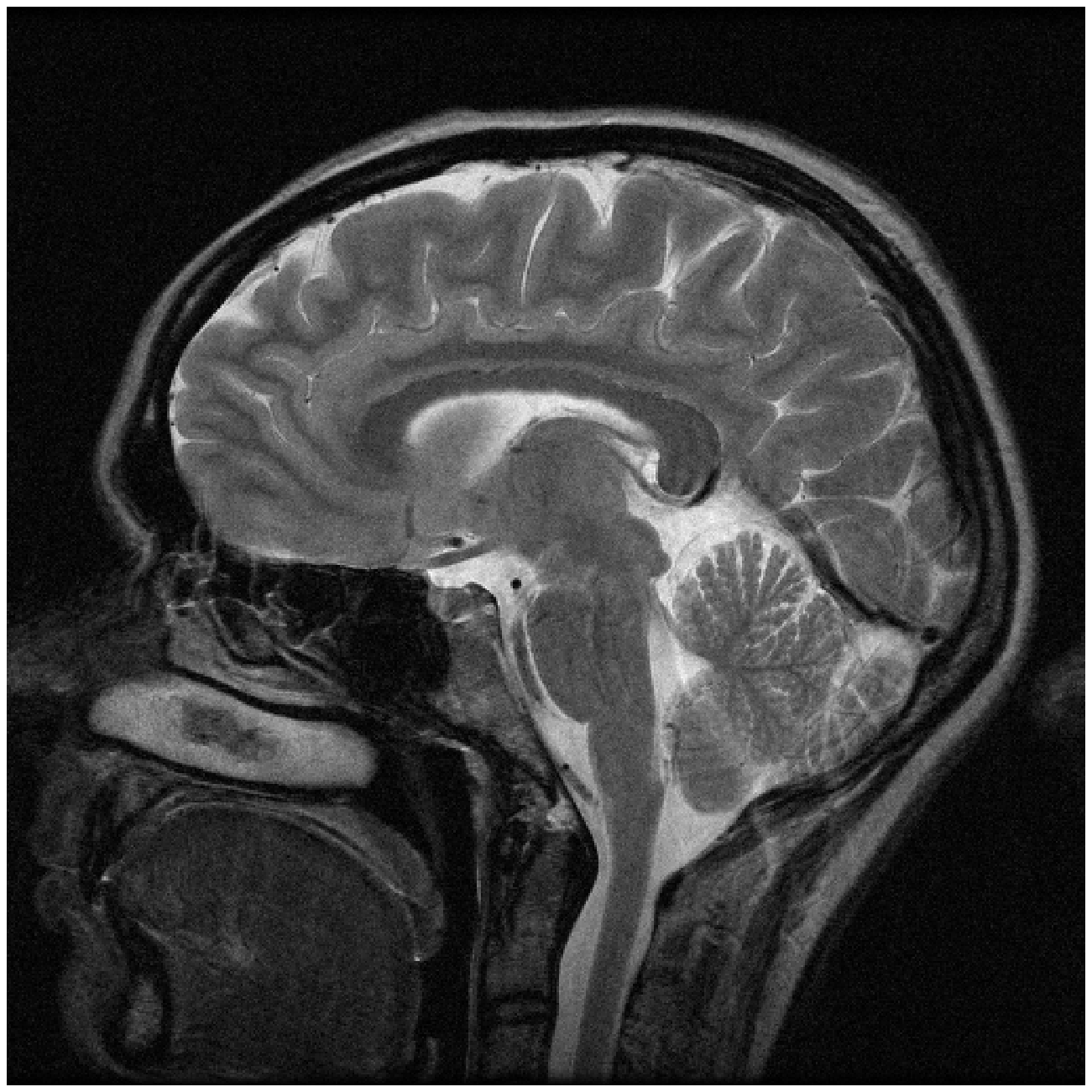} \\
 (e) &  (f)  \\
\end{tabular}
\caption{Test data (only magnitudes are displayed here): (a) A $512 \times 512$ complex-valued brain image that is available for download at \protect\url{http://web.stanford.edu/class/ee369c/data/brain.mat}; (b)  $256 \times 256$ complex-valued T2 weighted brain image that is publicly available \cite{PANOweb}, and was acquired from a healthy volunteer at a 3 T Siemens Trio Tim MRI scanner using the T2-weighted turbo spin echo sequence (TR/TE = 6100/99 ms, $220 \times 220$ mm$^{2}$ field of view, 3 mm slice thickness); (c) water phantom data (complex-valued and size $256 \times 256$) that is publicly available \cite{Quweb}, and was acquired at a 7 T Varian MRI system (Varian, Palo Alto, CA, USA) with the spin echo sequence (TR/TE =2000/100 ms, $80 \times 80$ mm$^{2}$ field of view, 2 mm slice thickness); (d) $512 \times 512$ real-valued (magnitude) MR image that was used in the simulations in a prior work \cite{bresai}; (e) $256 \times 256$ complex-valued T2 weighted brain image that is publicly available \cite{Quweb}, and was acquired from a healthy volunteer at a 3 T Siemens Trio Tim MRI scanner using the T2-weighted turbo spin echo sequence (TR/TE = 6100/99 ms, $220 \times 220$ mm$^{2}$ field of view, 3 mm slice thickness); (f) $512 \times 512$ complex-valued reference sagittal slice provided by Prof. Michael Lustig, UC Berkeley.
The image in (b) has been rotated clockwise by 90$^{\circ}$ here from the orientation in \cite{PANOweb} for display purposes. In the experiments, we use the same orientation as in \cite{PANOweb}.}
\label{im1bcs}
\end{center}
\vspace{-0.2in}
\end{figure}

We study the convergence behavior and effectiveness of the proposed BCS methods involving (P1) and (P2) for compressed sensing MRI (CS MRI). The MR data used in our experiments are shown\footnote{The images have pixel intensities (magnitudes) in the range [0, 1] (normalized). We use a gamma correction to (better) display some of the images and results in this work.} in Figure \ref{im1bcs}, and are labeled a-f. 
We simulate various undersampling patterns in k-space\footnote{We simulate the k-space of an image $x$ using the command fftshift(fft2(ifftshift(x))) in Matlab.}
including variable density 2D random sampling\footnote{Although 2D random sampling is not practically realizable for 2D imaging, the sampling scheme is feasible when data corresponding to multiple image slices are jointly acquired, and the frequency encode (readout) direction is chosen perpendicular to the image plane. In this case, one could apply an inverse Fourier transform for such 3D data along the (fully sampled) readout direction, and then perform decoupled 2D reconstructions (slice by slice). The BCS methods would learn 2D models in this case (a different model for each 2D slice) that sparsify spatial features. Our experiments in this work with 2D random sampling are meant to simulate such 2D reconstructions.} \cite{josh, bresai}, and Cartesian sampling with variable density random phase encodes (1D random). 
We then use the proposed algorithms for (P1) and (P2) to reconstruct the images from undersampled measurements.
Our algorithm for (P1) for MRI is called Unitary Transform learning MRI (UTMRI), and our method for (P2) for MRI is referred to as UNIon of Transforms lEarning MRI (UNITE-MRI).


We compare the reconstructions provided by our methods to those provided by the following schemes: 1) the Sparse MRI method \cite{lustig} that utlilizes wavelets and total variation as \emph{fixed} transforms; 2) the DLMRI method \cite{bresai} that learns adaptive overcomplete synthesis dictionaries; 3) the PANO method \cite{Qu2014843} that exploits the non-local similarities between image patches (similar to \cite{dbov}), and employs a 3D transform to sparsify groups of similar patches; and 4) the PBDWS method \cite{Qu12} that is a recent \emph{partially} adaptive sparsifying transform based reconstruction method that uses redundant wavelets and trained patch-based geometric directions. 
We also include in our comparisons the TLMRI method that was proposed and used in the experiments in a very recent work  \cite{sabressiims1}. The TLMRI method (Algorithm A1 in \cite{sabressiims1}) is for a variant of Problem (P1) involving a sparsity constraint (instead of penalties) and a transform regularizer $- \log \left | \det W \right | + 0.5 \left \| W \right \|_{F}^{2}$ that controls the condition number of $W$.

We simulated the Sparse MRI, PBDWS, PANO,  DLMRI, and TLMRI methods using the software implementations available from the respective authors' websites \cite{lus33, Quweb, PANOweb, dlmri1, tlmri1}.
 We used the built-in parameter settings in the first three implementations, which performed well in our experiments\footnote{Upon tuning the parameters of these methods (from their default settings) for a subset of the data used in our experiments, we did not observe any marked performance improvements with tuning.}.
Specifically, for the PBDWS method, the shift invariant discrete Wavelet transform (SIDWT) based reconstructed image is used as the \emph{guide} (initial) image \cite{Qu12, Quweb}.
We employed the zero-filling reconstruction (produced within the PANO demo code \cite{PANOweb}) as the initial guide image for the PANO method \cite{Qu2014843, PANOweb}.


The DLMRI implementation \cite{dlmri1} used image patches of size $6 \times 6$ \cite{bresai}, and learned a four fold overcomplete dictionary $D \in \mathbb{R}^{36\times 144}$ using 25 iterations of the algorithm.
The patch stride $r=1$, and $14400$ (found empirically) randomly selected patches are used during the dictionary learning step (executed for 20 iterations) of the DLMRI algorithm.
Mean-subtraction is not performed for the patches prior to the dictionary learning step of DLMRI. (We adopted this strategy for DLMRI as it led to better performance in our experiments.)
A maximum sparsity level (of $s= 7$ per patch) is employed together with an error threshold (for sparse coding) during the dictionary learning step.
The $\ell_{2}$ error threshold per patch varies linearly from $0.48$ to $0.04$ over the DLMRI iterations, except in the case of Figs. \ref{im1bcs}(a), \ref{im1bcs}(c), and \ref{im1bcs}(f) (noisier data), where it varies from $0.48$ to $0.15$ over the iterations. 
These parameter settings (all other settings are as per the indications in the DLMRI-Lab toolbox \cite{dlmri1}) worked quite well for DLMRI. 


For UTMRI and UNITE-MRI, image patches of size $6 \times 6$ were again used ($n= 36$ like for DLMRI), $r=1$ (with patch wrap around), $\nu = 10^{6}/p$ (where $p$ is the number of image pixels), $C=10^{5}$, and $K=16$.
The image, transforms, and sparse coefficients in the algorithms are initialized\footnote{While we use the naive zero-filling Fourier reconstruction to initialize $x$ in our experiments here for simplicity, one could also use other better initializations for $x$ such as the SIDWT based reconstructed image \cite{Qu12}, or the reconstructions produced by recent methods (e.g., PBDWS, PANO, etc.). We have observed empirically that better initializations may lead to faster convergence of our algorithms, and our methods typically tend to improve the image quality compared to the initializations (assuming properly chosen parameters).} as indicated in Section \ref{sec3}. The clusters in UNITE-MRI were initialized appropriately.
Both algorithms ran for 120 iterations in the experiments in Sections \ref{resu} and \ref{resu3}.
The parameter $\eta$ in our methods is set to $0.007$, except in the case of Figs. \ref{im1bcs}(a), \ref{im1bcs}(c), and \ref{im1bcs}(f) (noisier data), where it is set to 0.05, and in the experiment in Section \ref{conver} (where our algorithms' convergence behavior is illustrated through an example), where it is set to 0.07.
We use even larger values of $\eta$ during the initial several iterations of our algorithms in Sections \ref{resu} and \ref{resu3}, leading to faster convergence and aliasing removal. 

Finally, for the recent TLMRI algorithm \cite{tlmri1}, we employed similar parameter settings and initializations as for UTMRI, but initialized the sparse coefficients as indicated in Section 5.1 of our prior work \cite{sabressiims1}. Additionally, the parameters $\hat{M}=1$ and $\lambda_{0}=0.2$ for TLMRI, and the sparsity parameter $s = 0.28 \times n N$ except in the case of Figs. \ref{im1bcs}(a), (c), and (f), where $s = 0.1 \times n N$. Even lower sparsity levels $s$ are used during the initial several iterations of TLMRI in Sections \ref{resu} and \ref{resu3}.

All simulations used Matlab. The computing platform used for the experiments was an Intel Core i5 CPU at 2.5 GHz and 4 GB memory, employing a 64-bit Windows 7 operating system.



Similar to prior work \cite{bresai}, we employ the peak-signal-to-noise ratio (PSNR), and high frequency error norm (HFEN) metrics to measure the quality of MR image reconstructions. The PSNR (expressed in decibels (dB)) is computed as the ratio of the peak intensity value of a reference image to the root mean square reconstruction error (computed between image magnitudes) relative to the reference.
The HFEN metric quantifies the quality of reconstruction of edges or finer features. A rotationally symmetric Laplacian of Gaussian (LoG) filter is used, whose kernel is of size $15 \times 15  $ pixels, and with a standard deviation of 1.5 pixels \cite{bresai}. HFEN is computed as the $\ell_{2}$ norm of the difference between the LoG filtered reconstructed and reference magnitude images.

\subsection{Convergence Behavior} \label{conver}

\begin{figure*}[!t]
\begin{center}
\begin{tabular}{cccc}
\includegraphics[height=1.28in]{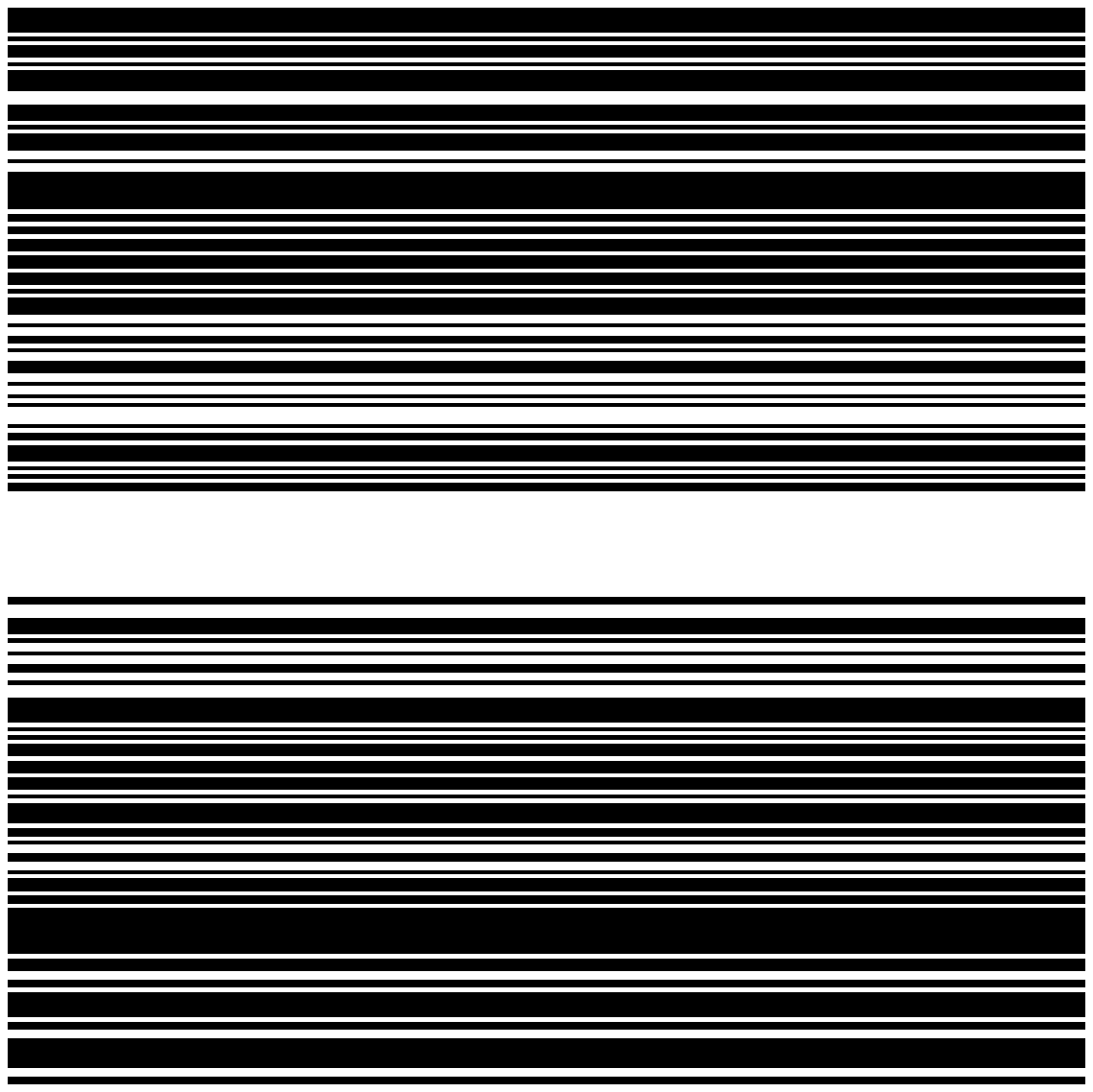}&
\includegraphics[height=1.28in]{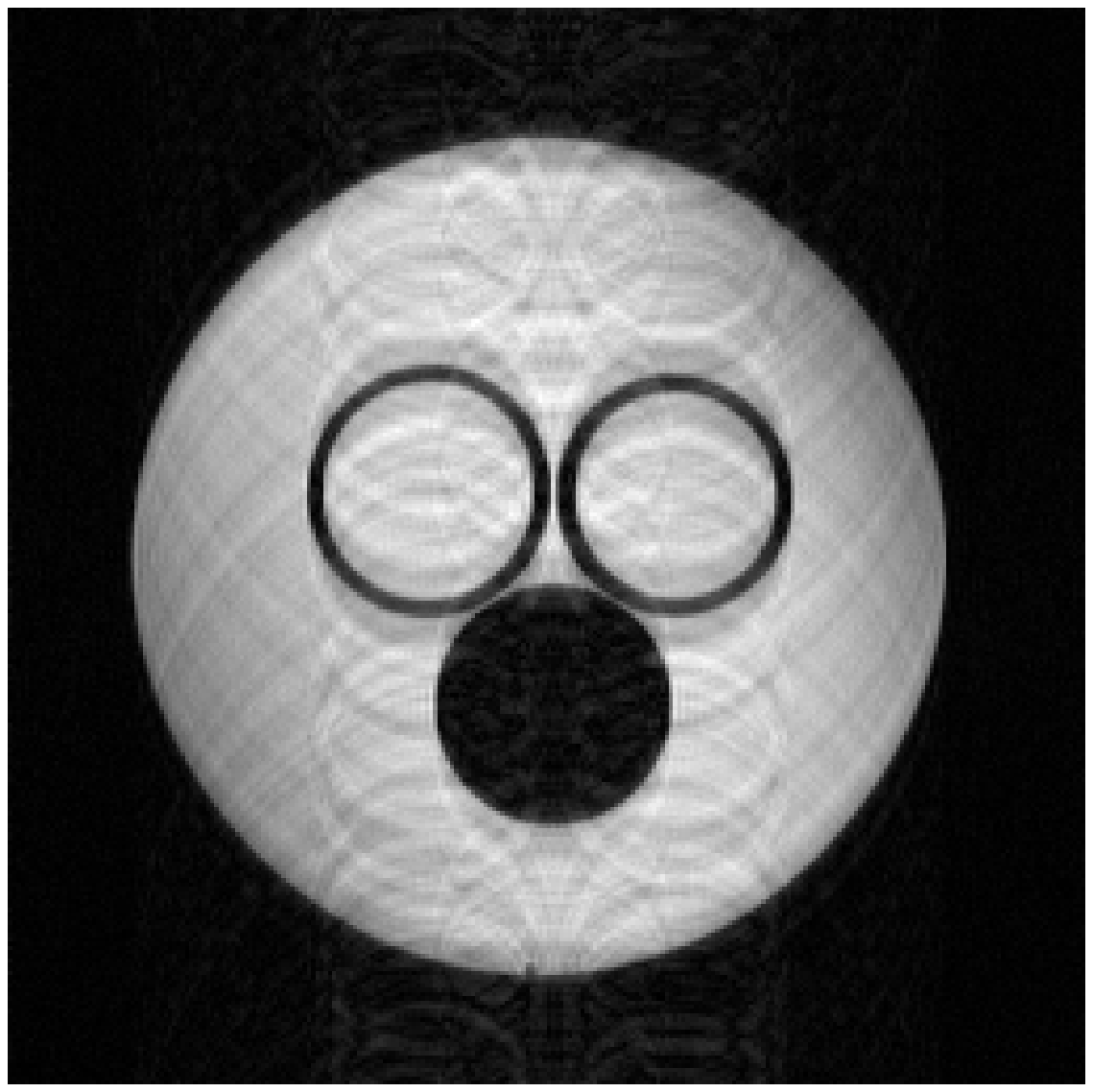}&
\includegraphics[height=1.28in]{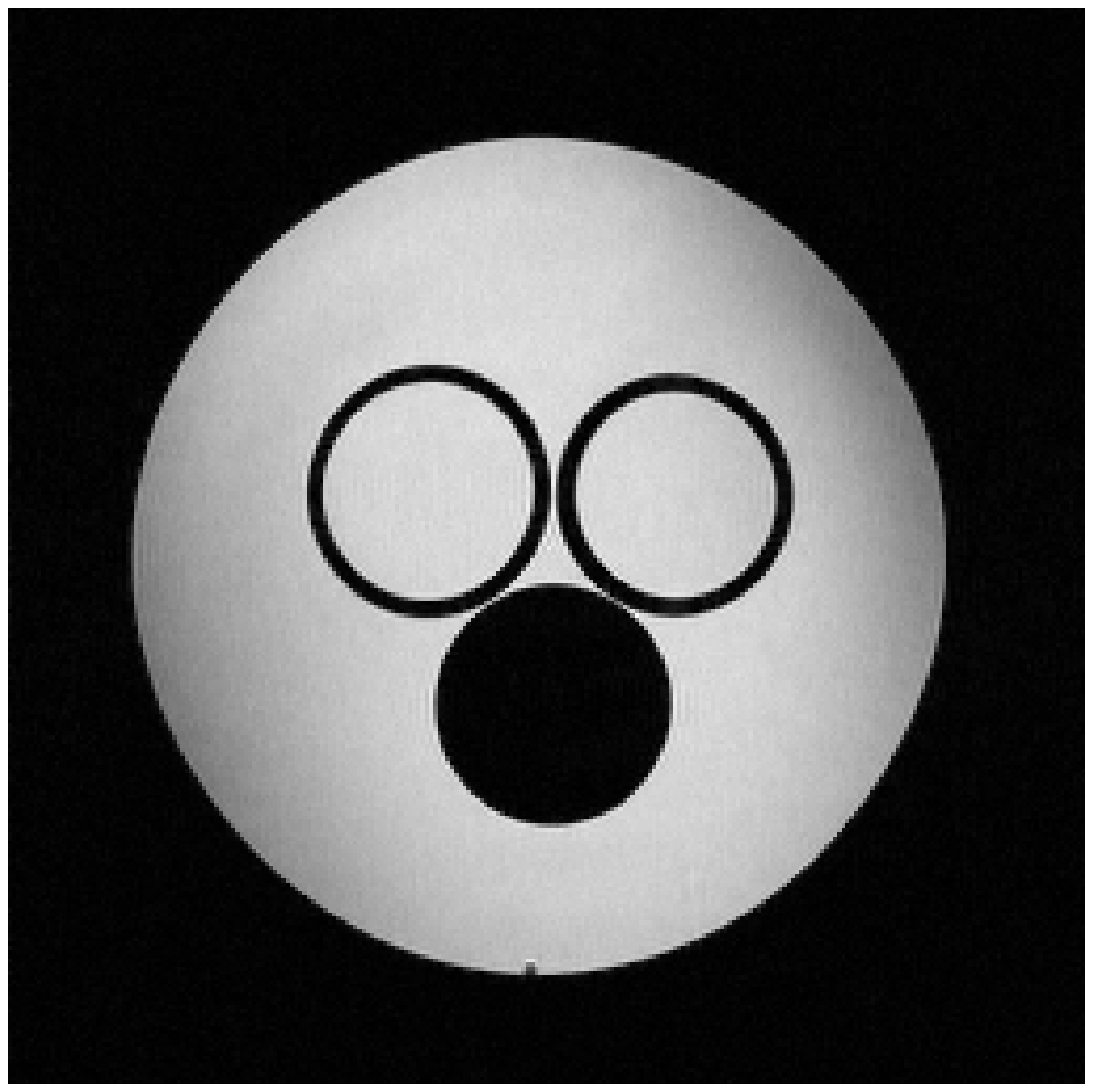}&
\includegraphics[height=1.2in]{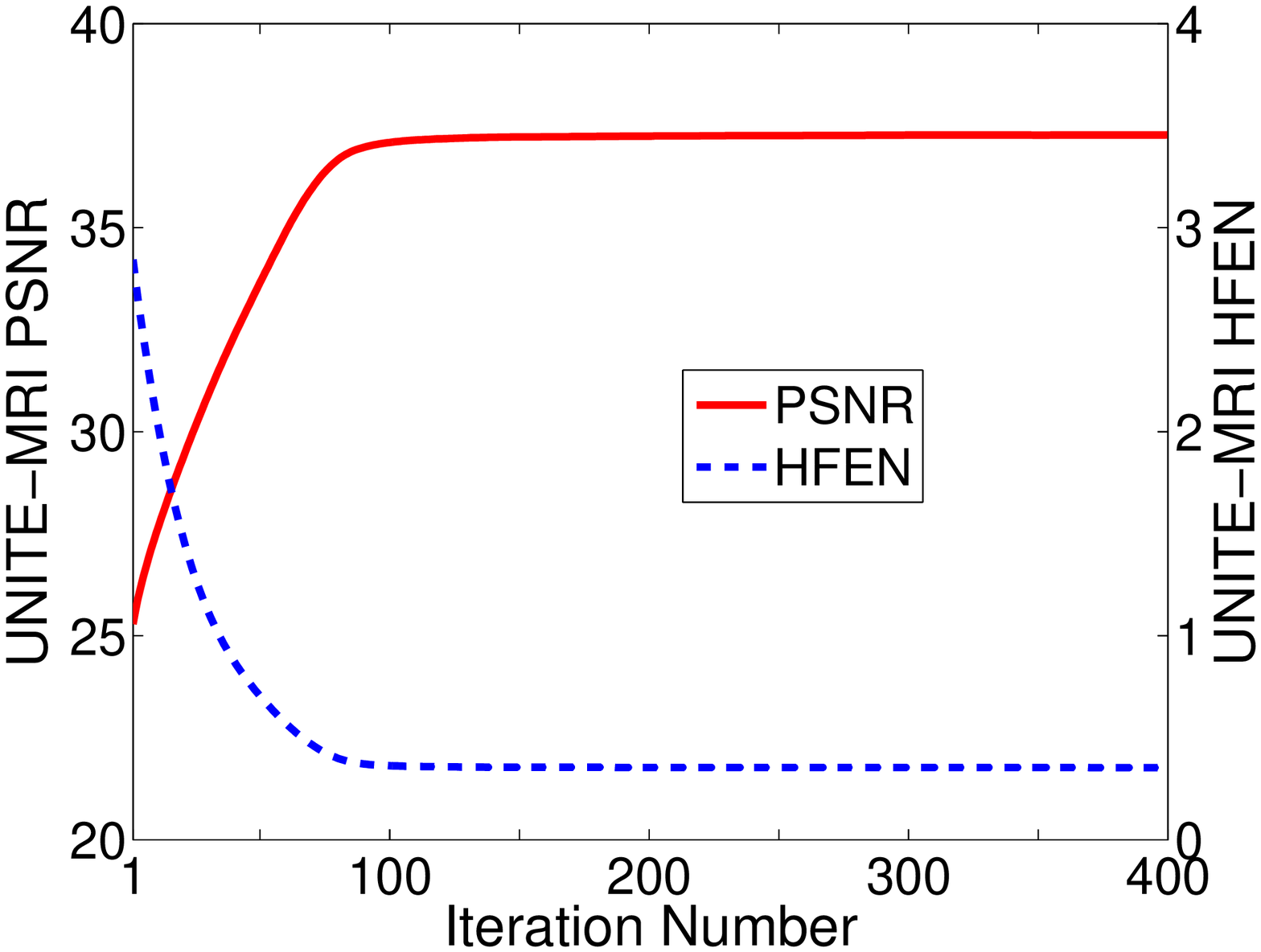}\\
(a) & (b) & (c) & (d)\\
\end{tabular}
\begin{tabular}{ccc}
\includegraphics[height=1.2in]{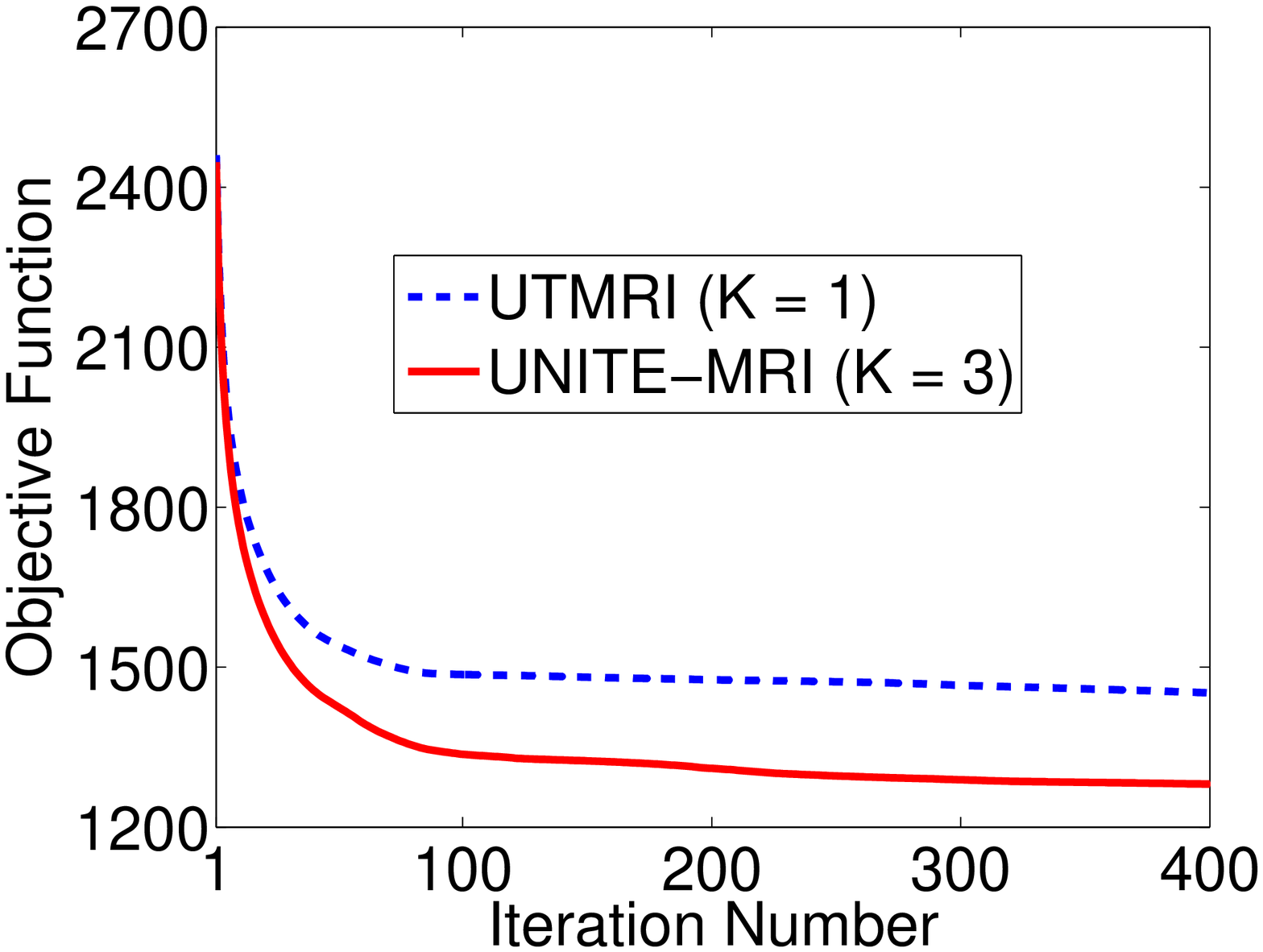}&
\includegraphics[height=1.2in]{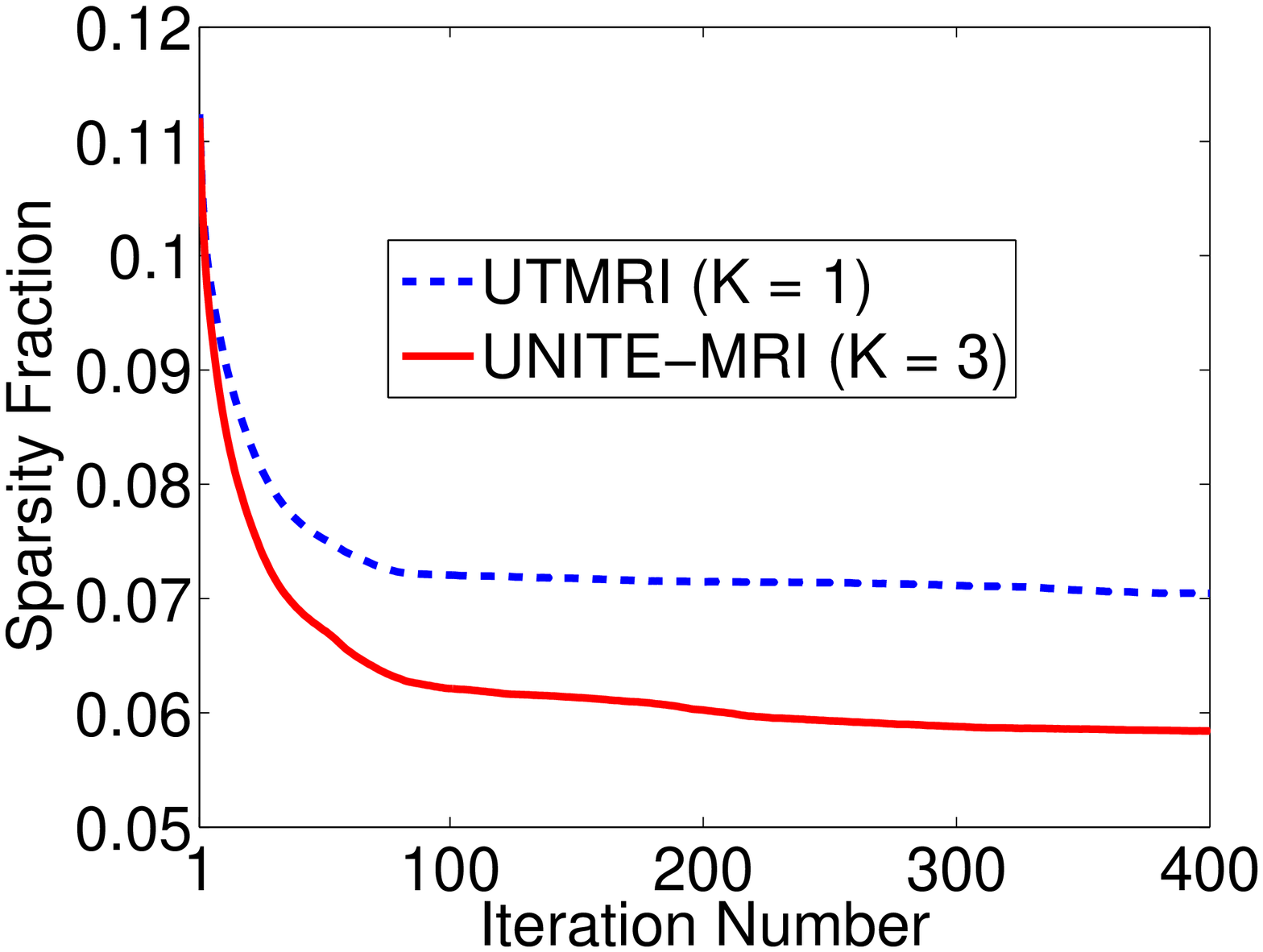}&
\includegraphics[height=1.2in]{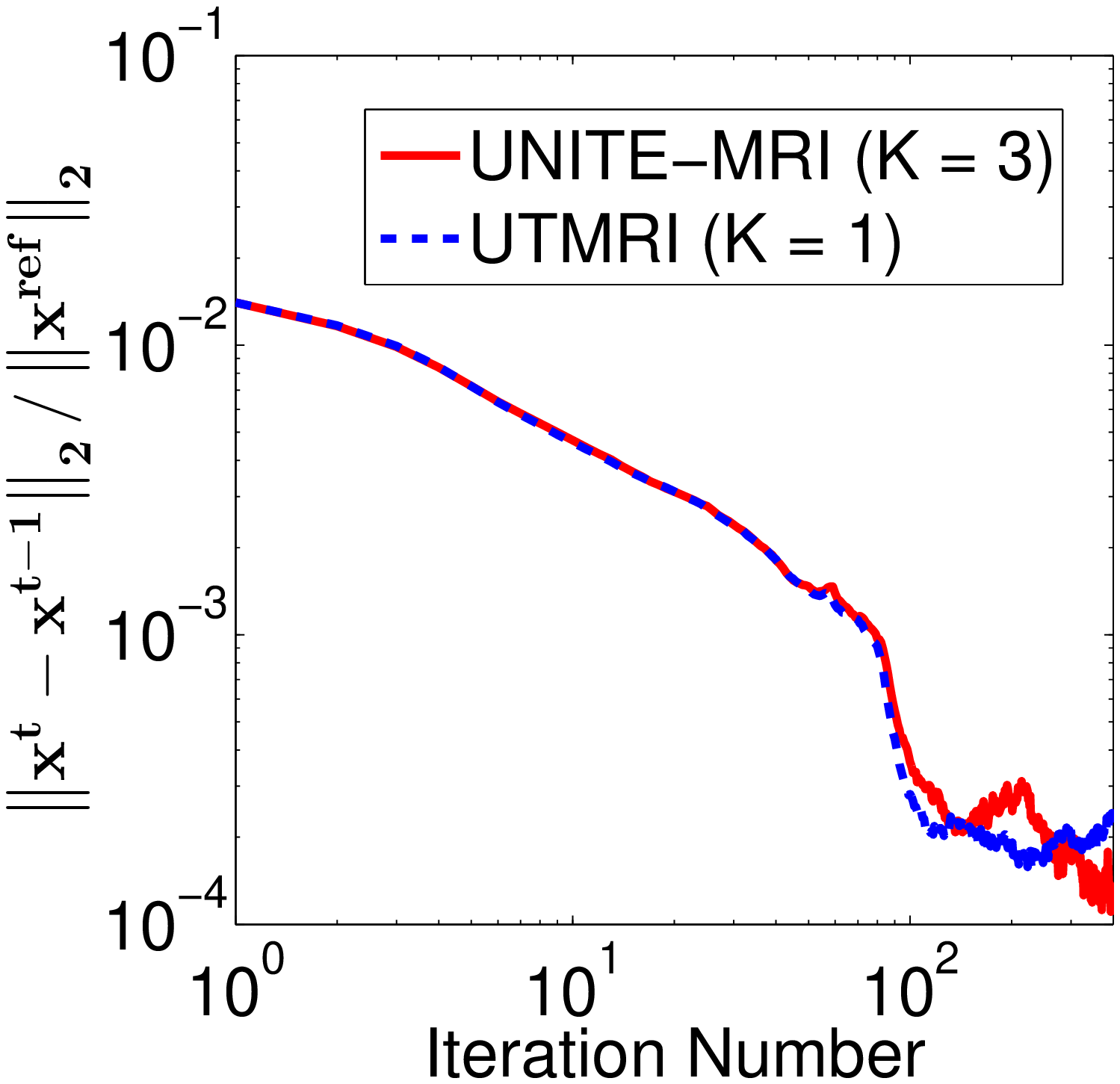}\\
(e) & (f) & (g)\\
\end{tabular}
\caption{Convergence behavior of UTMRI and UNITE-MRI ($K=3$) with Cartesian sampling and 2.5x undersampling: (a) sampling mask in k-space; (b) magnitude of initial zero-filling reconstruction ($24.9$ dB); (c) UNITE-MRI reconstruction magnitude ($37.3$ dB); (d) PSNR and HFEN for UNITE-MRI; (e) objective function values for UNITE-MRI and UTMRI; (f) Sparsity fractions (i.e., fraction of non-zeros in the sparse code matrix $B$) for UNITE-MRI and UTMRI; and (g) changes between successive iterates ($\left \| x^{t}-x^{t-1} \right \|_{2}$) normalized by the norm of the reference image ($\left \| x^{\mathrm{ref}} \right \|_{2} = 122.2$) for UNITE-MRI and UTMRI.}
\label{imcvbcs}
\end{center}
\vspace{-0.2in}
\end{figure*}

\begin{figure}[!t]
\begin{center}
\begin{tabular}{cc}
\includegraphics[height=1.2in]{fig4/UNITEMRIrecon}&
\includegraphics[height=1.2in]{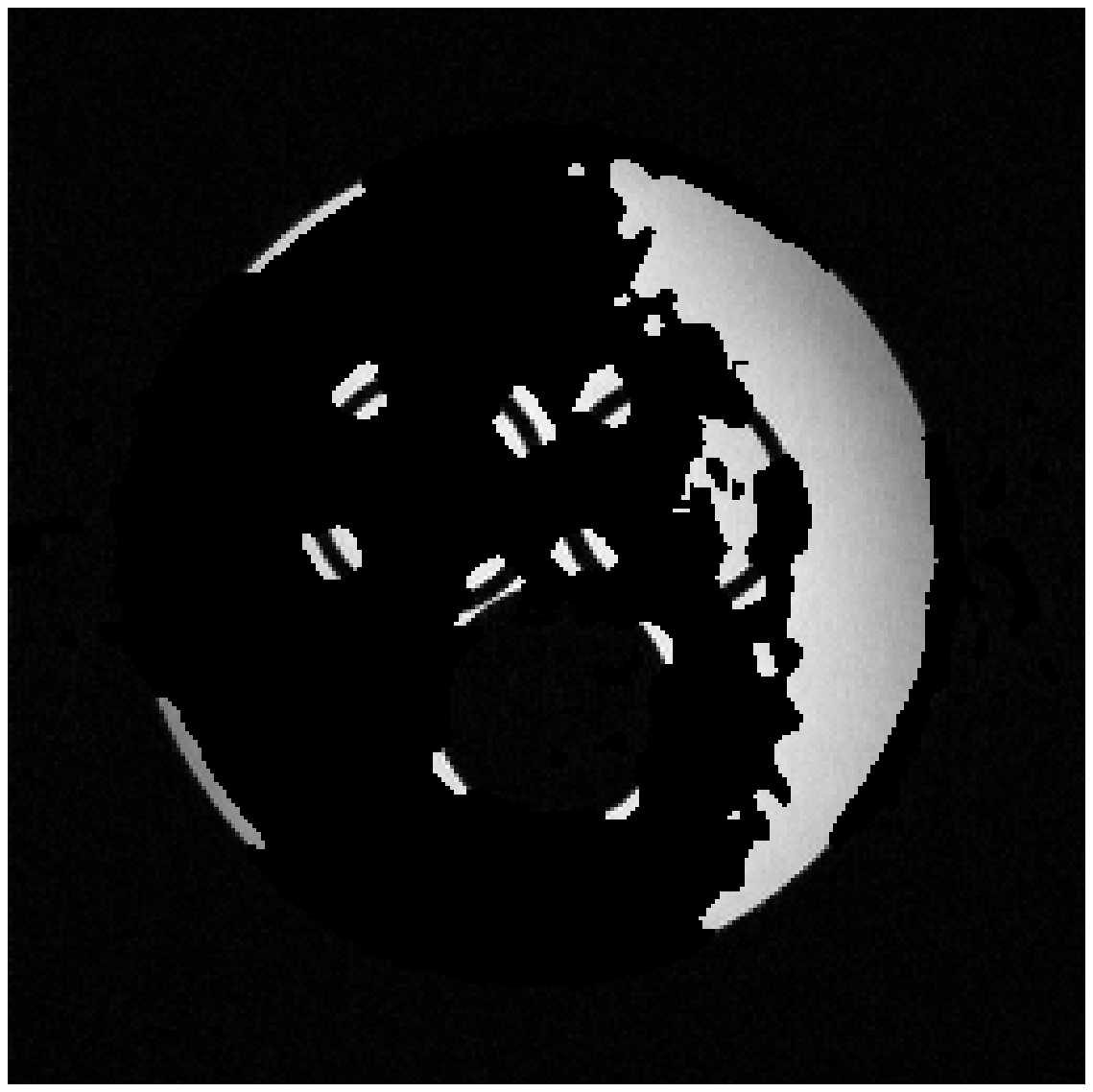} \\
(a) & (b) \\
\end{tabular}
\begin{tabular}{cc}
\includegraphics[height=1.2in]{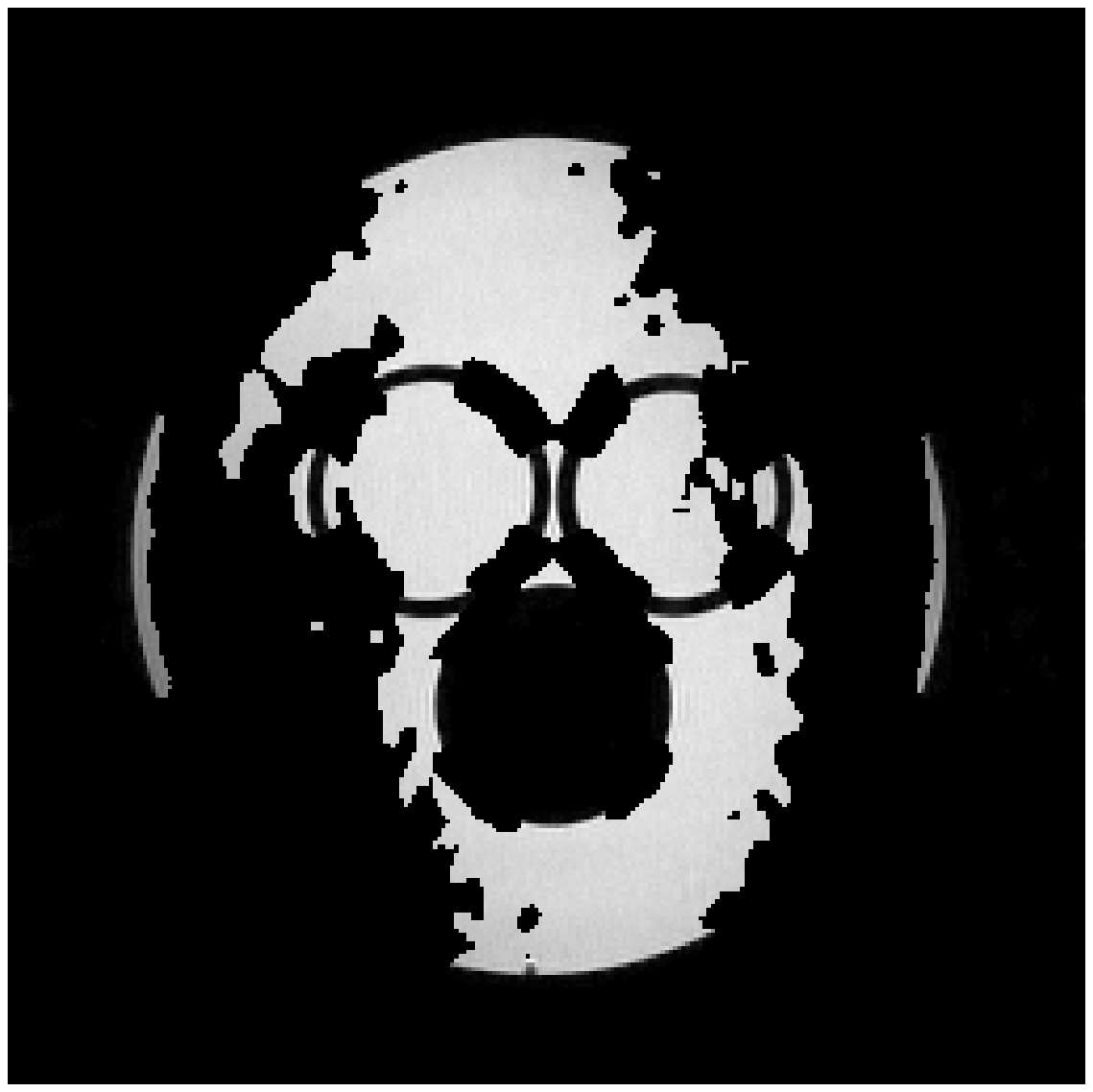}  &
\includegraphics[height=1.2in]{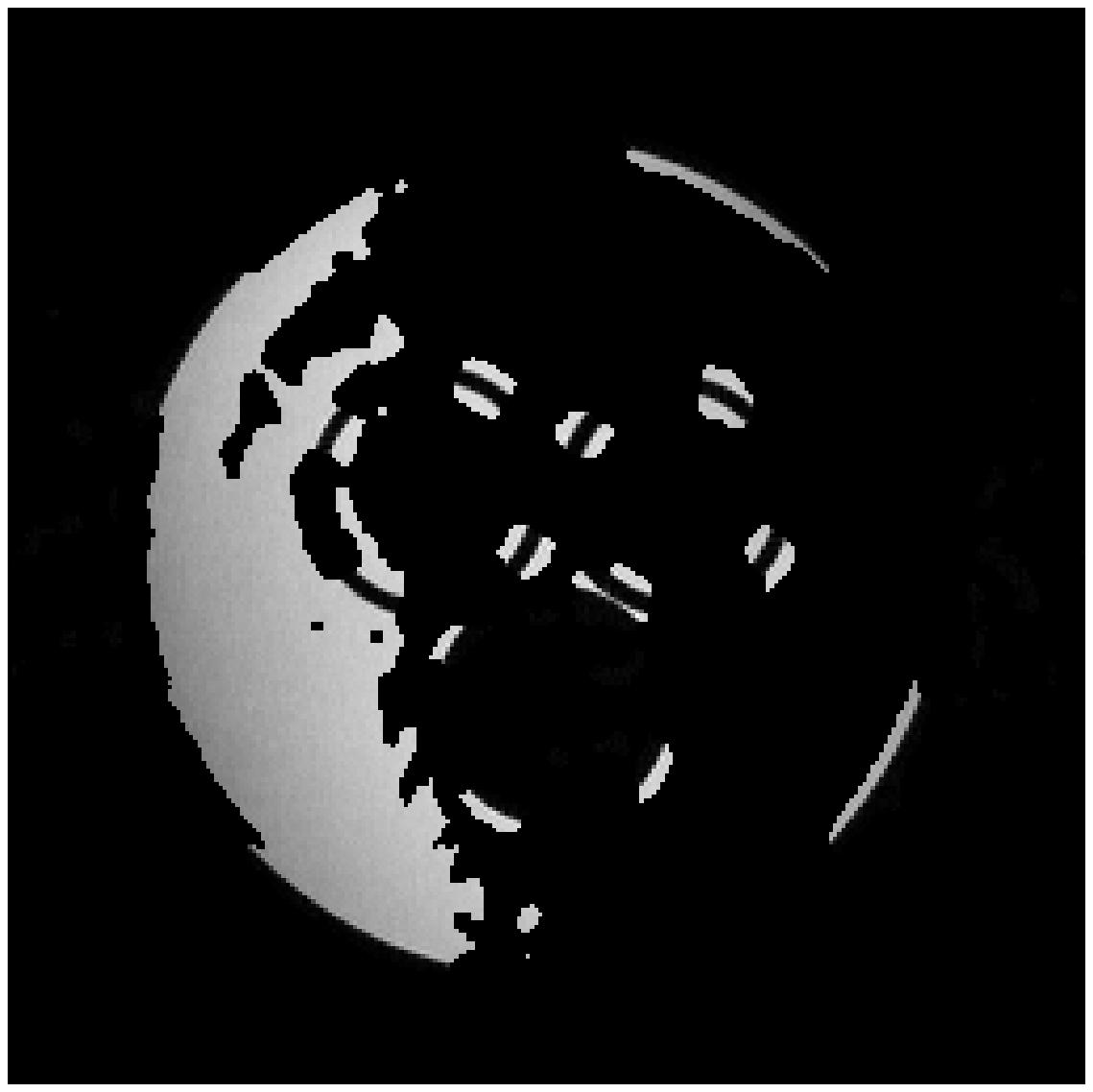}  \\
(c) &  (d)  \\
\includegraphics[height=1.2in]{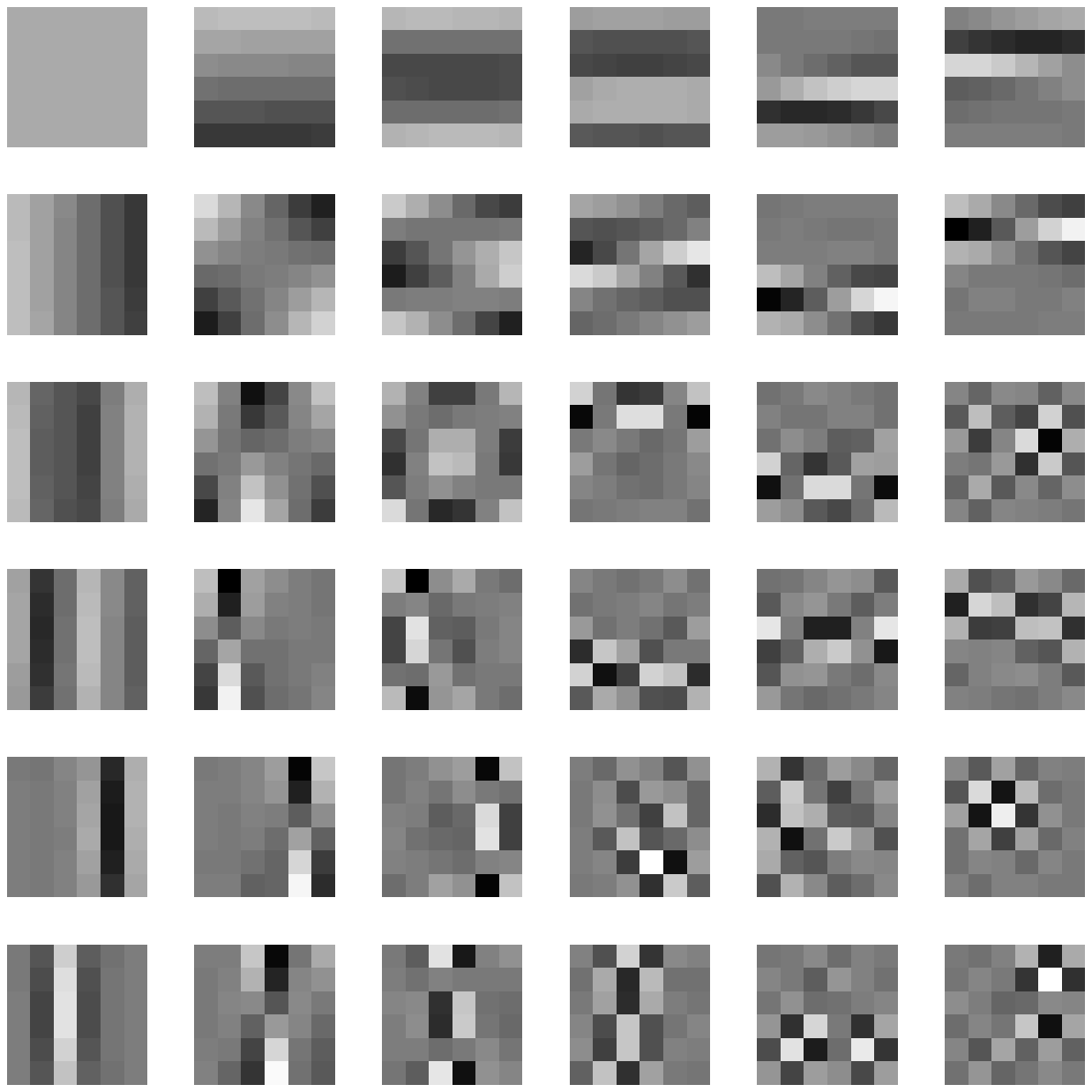} &
\includegraphics[height=1.2in]{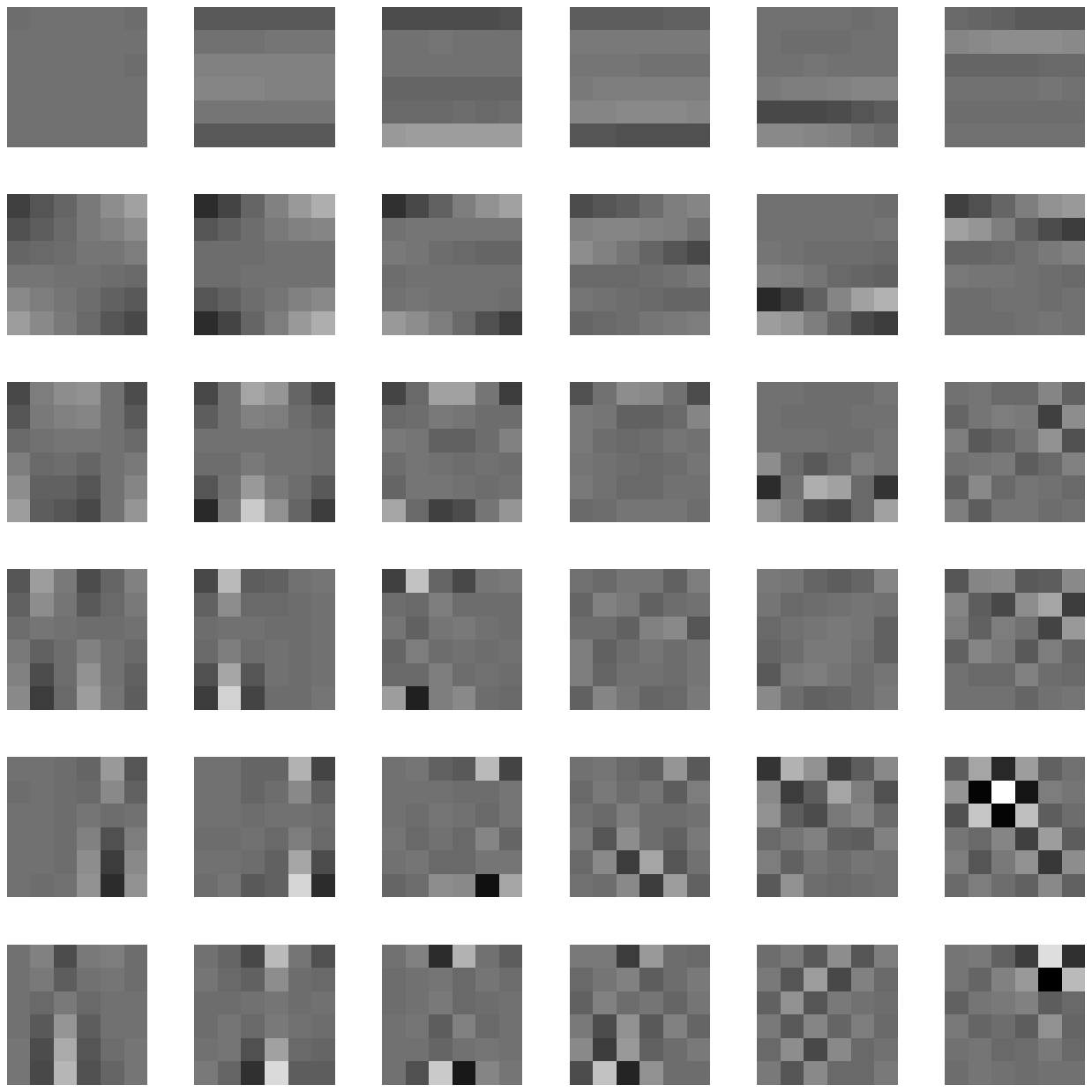} \\
 (e) &  (f)  \\
\end{tabular}
\caption{UNITE-MRI ($K=3$) clustering and learning: (a) UNITE-MRI reconstruction (magnitude shown);
(b)-(d) image pixels (with reconstructed intensities) in (a) grouped into the first, second, and third clusters, respectively, overlaid on black backgrounds;
(e) real and (f) imaginary parts of the learnt transform in the second cluster, with the matrix rows shown as patches. Each of the images (b)-(d) shows edges with specific orientations.}
\label{imcvbcs2}
\end{center}
\vspace{-0.25in}
\end{figure}


In this experiment, we consider the complex-valued reference image in Fig. \ref{im1bcs}(c), and perform 2.5 fold undersampling of the k-space of the reference. The variable density sampling mask is shown in Fig. \ref{imcvbcs}(a).
We study the behavior of the UTMRI and UNITE-MRI algorithms when used to reconstruct the image from the undersampled k-space measurements. UNITE-MRI is employed with $3$ clusters for the image patches. The objective function (Fig. \ref{imcvbcs}(e)) converged monotonically and quickly over the iterations for both UTMRI and UNITE-MRI. In particular, the objective for UNITE-MRI converged to a much lower value than for UTMRI. This is because UNITE-MRI learned a richer model and achieved a lower value than UTMRI for both the sparsification error and sparsity penalty terms in its cost (i.e., in (P2)). The sparsity fractions (i.e., the fraction of non-zeros in the sparse code matrix $B$) achieved by the UTMRI and UNITE-MRI algorithms over their iterations are shown in Fig. \ref{imcvbcs}(f). UNITE-MRI has clearly achieved lower (i.e., better) sparsities for image patches.
The changes between successive iterates $\left \| x^{t}-x^{t-1} \right \|_{2}$ (Fig. \ref{imcvbcs}(g)) decrease to small values for both UTMRI and UNITE-MRI. 
Such behavior was established for these algorithms by Theorems \ref{theorem1bc} and \ref{theorem2bc}, and is indicative (a necessary but not suffficient condition) of convergence of the entire sequence $\left \{ x^{t} \right \}$. 

The initial zero-filling reconstruction (Fig. \ref{imcvbcs}(b)) has large aliasing artifacts along the phase encoding (vertical) direction, as expected for the undersampled measurements scenario. The initial PSNR is only $24.9$ dB. In contrast, the UNITE-MRI reconstruction (Fig. \ref{imcvbcs}(c)) is much improved and has a PSNR of $37.3$ dB. Both the PSNR and HFEN metrics (Fig. \ref{imcvbcs}(d)) improve significantly and converge quickly for UNITE-MRI. UTMRI exhibited similar behavior.
However, the UTMRI reconstruction has a lower PSNR of $37.1$ dB.


Why does UNITE-MRI provide an improvement over UTMRI in reconstructing the rather simple (mostly smooth) image in this experiment?
To answer this question, we study the clustering results produced by the union-of-transforms based UNITE-MRI.
Since we work with overlapping image patches, each pixel in the image belongs to several different overlapping patches. We cluster an image pixel into a particular class $C_{k}$ if the majority of the patches to which it belongs are clustered into that class by the UNITE-MRI algorithm.
Figs. \ref{imcvbcs2}(b)-(d) show image pixels from the reconstructed image that are clustered into each of the 3 classes by UNITE-MRI. The pixels from each class (shown with the reconstructed intensities) are overlaid on a black background in these images. The results show that UNITE-MRI groups regions that share common orientations of edges. (Edges exist at a variety of orientations for the phantom image.) For example, the second cluster (Fig. \ref{imcvbcs2}(c)) captures near horizontal and vertical edges\footnote{The smooth regions of the image appear in all the clusters. This is because they are fairly (equally) well sparsified by any of the learnt directional transforms.}. The learnt transform for this cluster is also shown. This is a complex-valued transform. The real (Fig. \ref{imcvbcs2}(e)) and imaginary (Fig. \ref{imcvbcs2}(f)) parts of the transform display frequency like structures, and in particular contain horizontal and vertical features that were learnt to sparsify the corresponding edges better.

In this example, the UNITE-MRI algorithm is able to group patches together according to the directionality of their edges, and it learns transforms in each cluster that are better suited to sparsify specific types of edges. 
Since UTMRI learns only a square transform, the learned transform is unable to sparsify the diverse features (edges) of the phantom image as well as the more adaptive and overcomplete UNITE-MRI transform. The reconstruction error maps shown later in Fig. \ref{im4bcsbb} show UNITE-MRI recovering the image edges better than UTMRI.

\subsection{Results and Comparisons} \label{resu}

We now consider the images a-f in Fig. \ref{im1bcs} and simulate the performance of the proposed UTMRI and UNITE-MRI algorithms at various undersampling factors, and with Cartesian sampling or 2D random sampling of k-space.
Table \ref{tab2bcs} lists the reconstruction PSNRs corresponding to the zero-filling, Sparse MRI, DLMRI, PBDWS, PANO, TLMRI, UTMRI, and UNITE-MRI reconstructions for various cases.


\begin{table*}[t]
\centering
\fontsize{9}{10pt}\selectfont
\begin{tabular}{|c|c|c|c|c|c|c|c|c|c|c|}
\hline
Image & Sampling & UF       & Zero-filling  & Sparse MRI & DLMRI        & PBDWS           & PANO  & TLMRI              & UTMRI          & UNITE-MRI \\ 
\hline
   c        &      Cartesian        &   2.5x       &  24.9          &   29.9         &   36.6          &    35.8            &  34.8   &  37.2                &  37.2             &\textbf{37.4} \\
\hline
   d        &     2D random       &    10x        &  23.2          &  24.9          &  41.4           &   41.1             &  42.4  &  44.3                &  44.0             &\textbf{44.6} \\
\hline
   d       &      2D random        &   20x        &   21.6         &  22.9          &  34.1           &   36.7             &  37.8  &   38.8                &  38.4             &\textbf{39.4} \\
\hline
   a       &      Cartesian           &    6.9x     &   27.9         &  28.6          &  30.9           &   31.1             &   31.1 &  31.4                &  31.3             &\textbf{31.5} \\
\hline
   e        &      Cartesian          &   2.5x      &   28.1         &  31.7          &  37.5           &  42.5              &   40.0  & 40.7                &  40.8            &\textbf{43.4} \\
\hline
   b        &      Cartesian           &   2.5x     &   27.7         &  31.6          &   39.2          &   43.3             &   41.3  &  42.6               &  42.5             &\textbf{44.3} \\
\hline
   f          &   2D random          &   5x         &  26.3          &  27.4          &  30.5           &  30.4              &  30.4   &  30.6               &  30.6             &  \textbf{30.7} \\
\hline
   c         &    2D random          &  6x          &  13.9         &  14.5          &  15.4            &  15.2              &  33.0  &  33.2               &  32.4             &  \textbf{33.6} \\
\hline
   e        &   Cartesian              &  4x           & 28.5         &  30.6          &  32.4            &  \textbf{34.5}              &  33.7  &  33.6               &   33.5            &   \textbf{34.5}\\
\hline 
\end{tabular}
\caption{PSNRs corresponding to the Zero-filling, Sparse MRI \cite{lustig}, DLMRI \cite{bresai}, PBDWS \cite{Qu12}, PANO \cite{Qu2014843}, TLMRI \cite{sabressiims1}, UTMRI, and UNITE-MRI reconstructions for various images and undersampling factors (UF), with Cartesian or 2D random sampling. The best PSNRs are marked in bold. The image labels are as per Fig. \ref{im1bcs}.}
\label{tab2bcs}
\vspace{-0.2in}
\end{table*}


The transform-based blind compressed sensing algorithms typically provide the best reconstruction PSNRs in Table \ref{tab2bcs} (analogous results were observed to usually hold with respect to the HFEN metric not shown in the table).
Specifically, the UNITE-MRI method provides an improvement of 3.2 dB in reconstruction PSNR on average  in Table \ref{tab2bcs} over the recent partially adaptive PBDWS method, and an average improvement of 1.7 dB over the recent non-local patch similarity-based PANO method.
It also provides significant improvements in reconstruction quality over the non-adaptive Sparse MRI method, and an average improvement of 4.6 dB over the synthesis dictionary-based DLMRI method. (Unlike the proposed transform-based schemes, the overcomplete dictionary-based DLMRI algorithm lacks convergence guarantees, and the NP-hard synthesis sparse coding in DLMRI lacks closed-form solutions.)
The single transform-based TLMRI or UTMRI methods perform better than prior methods in most cases in Table \ref{tab2bcs}. TLMRI provides slightly better (about 0.2 dB better) PSNRs than UTMRI on the average.
Importantly, the union of transforms based UNITE-MRI provides 1 dB better reconstruction PSNR on the average compared to the single transform based UTMRI.
This indicates that the union of transforms (or OCTOBOS \cite{saiwen}) model is a better match for the characteristics of the MR images than a single transform model for all image patches.

\begin{figure}[!t]
\begin{center}
\begin{tabular}{cc}
\includegraphics[height=1.25in]{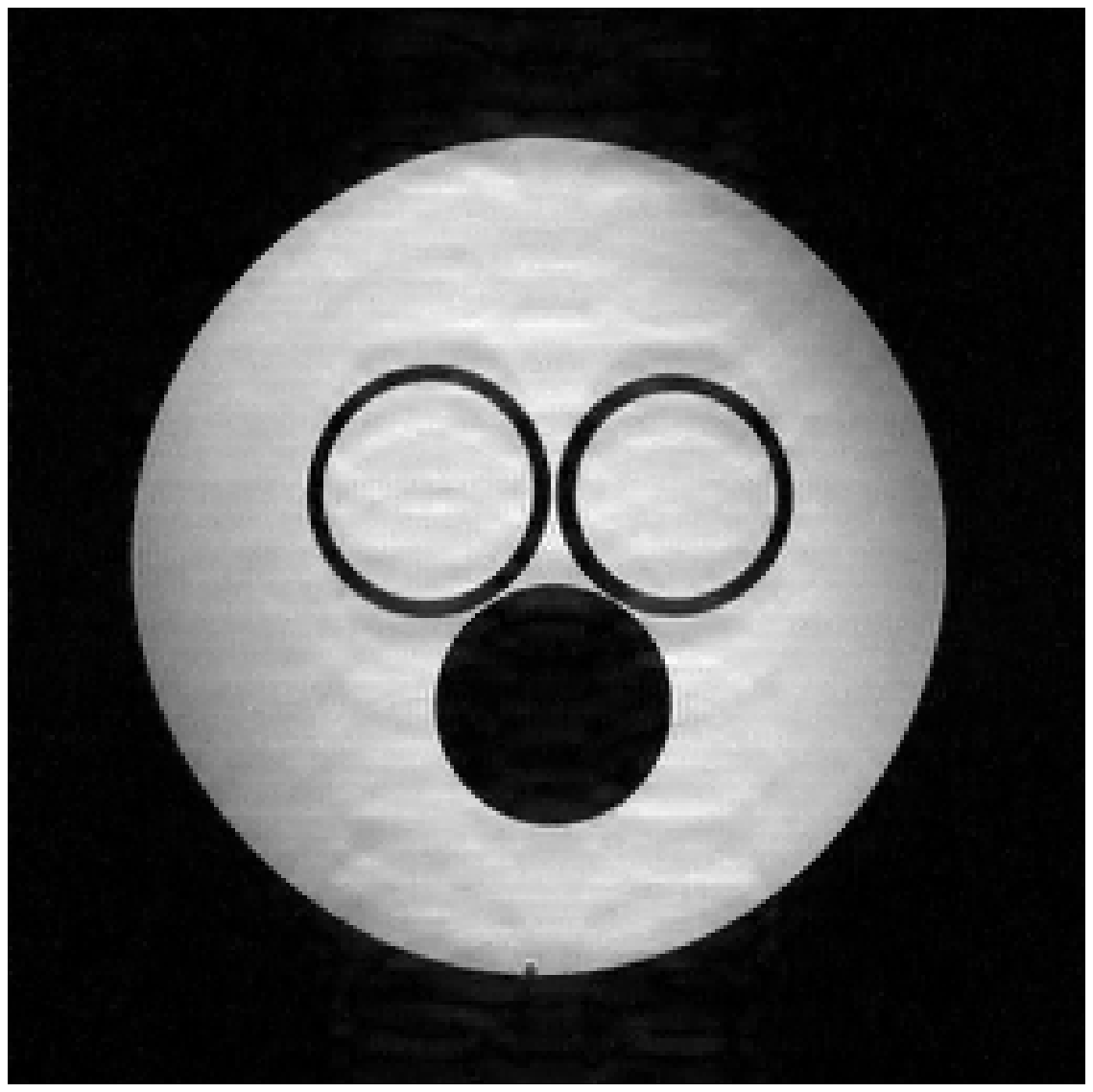}&
\includegraphics[height=1.28in]{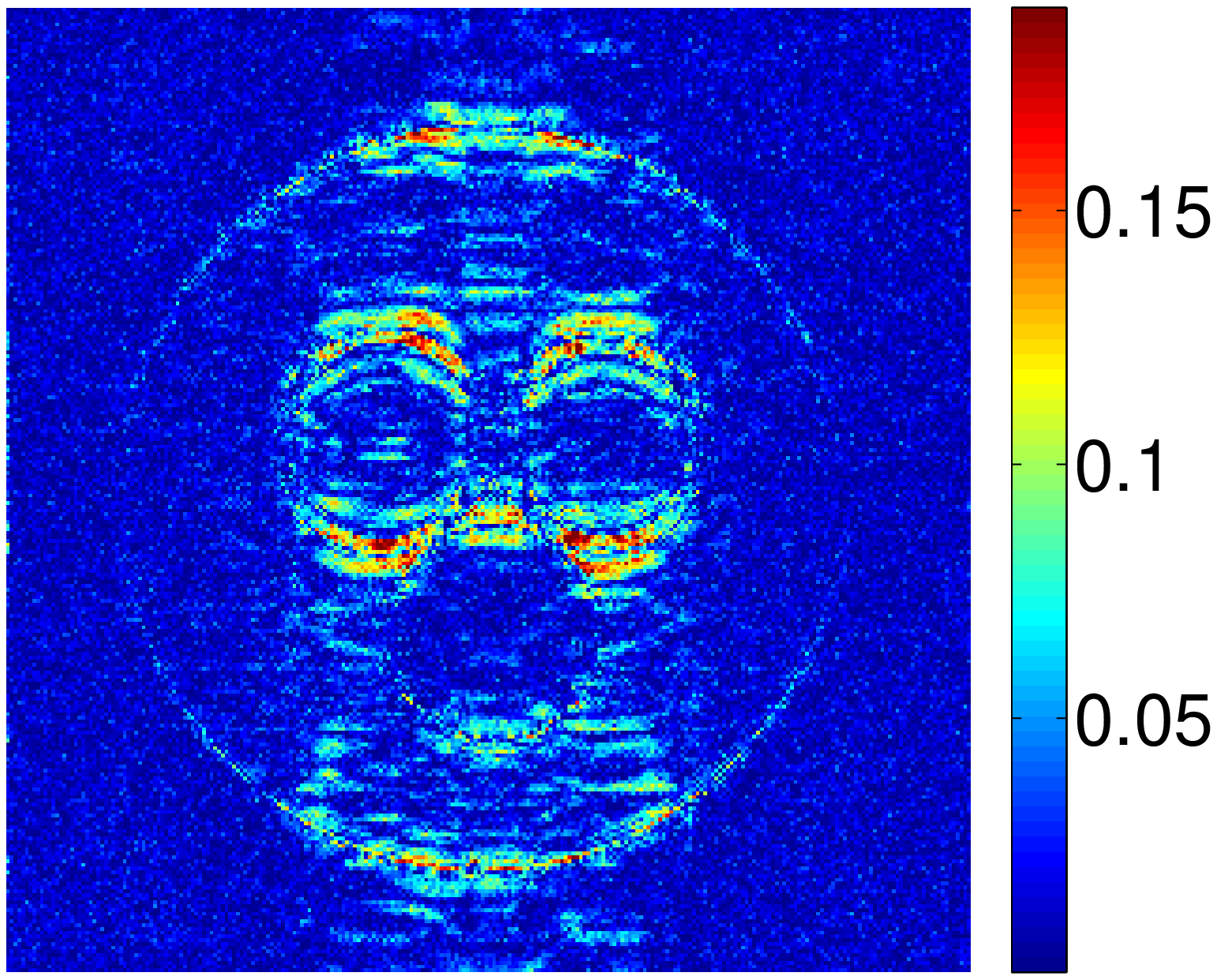}\\
(a) & (b) \\
\includegraphics[height=1.25in]{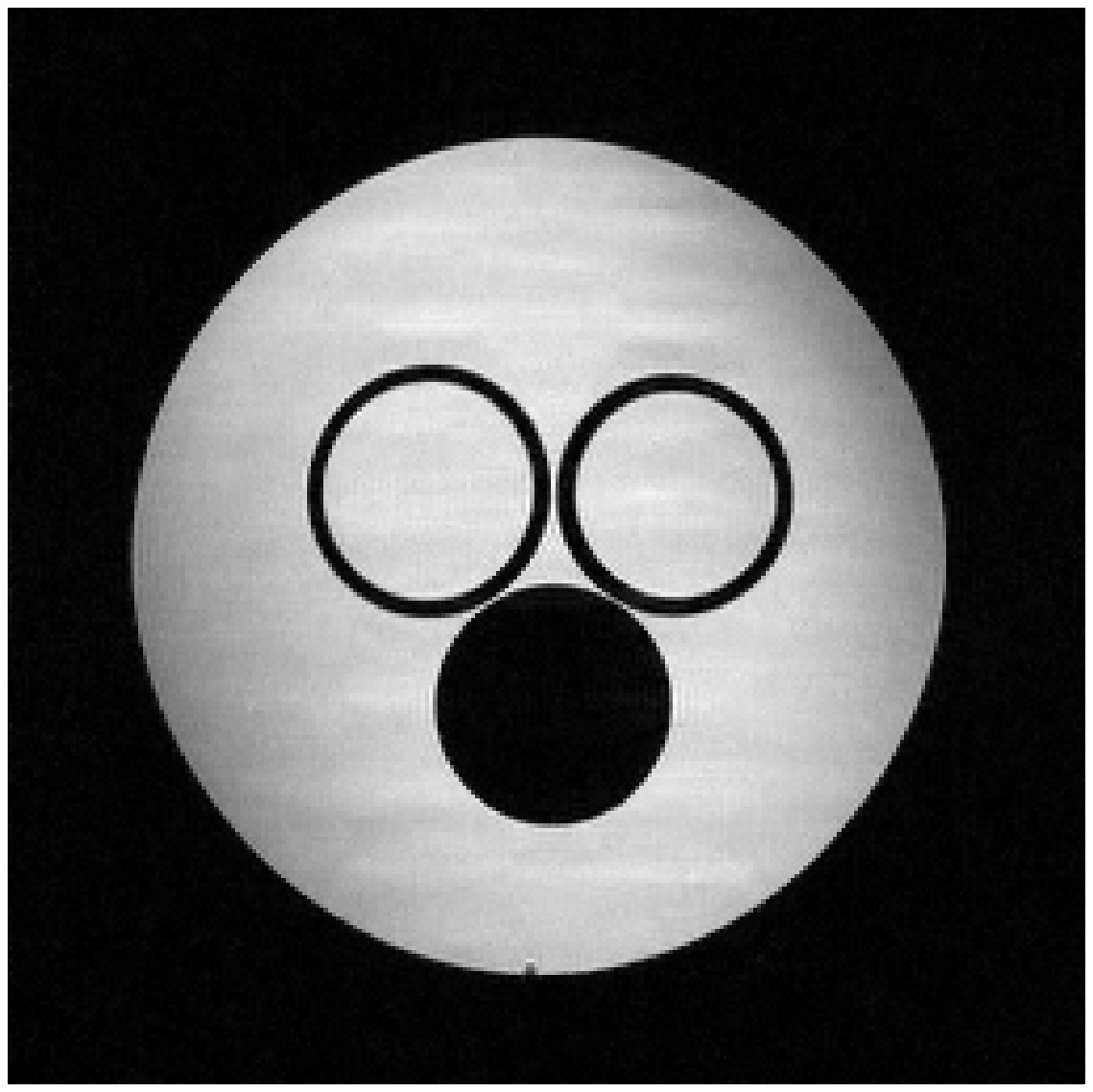}&
\includegraphics[height=1.28in]{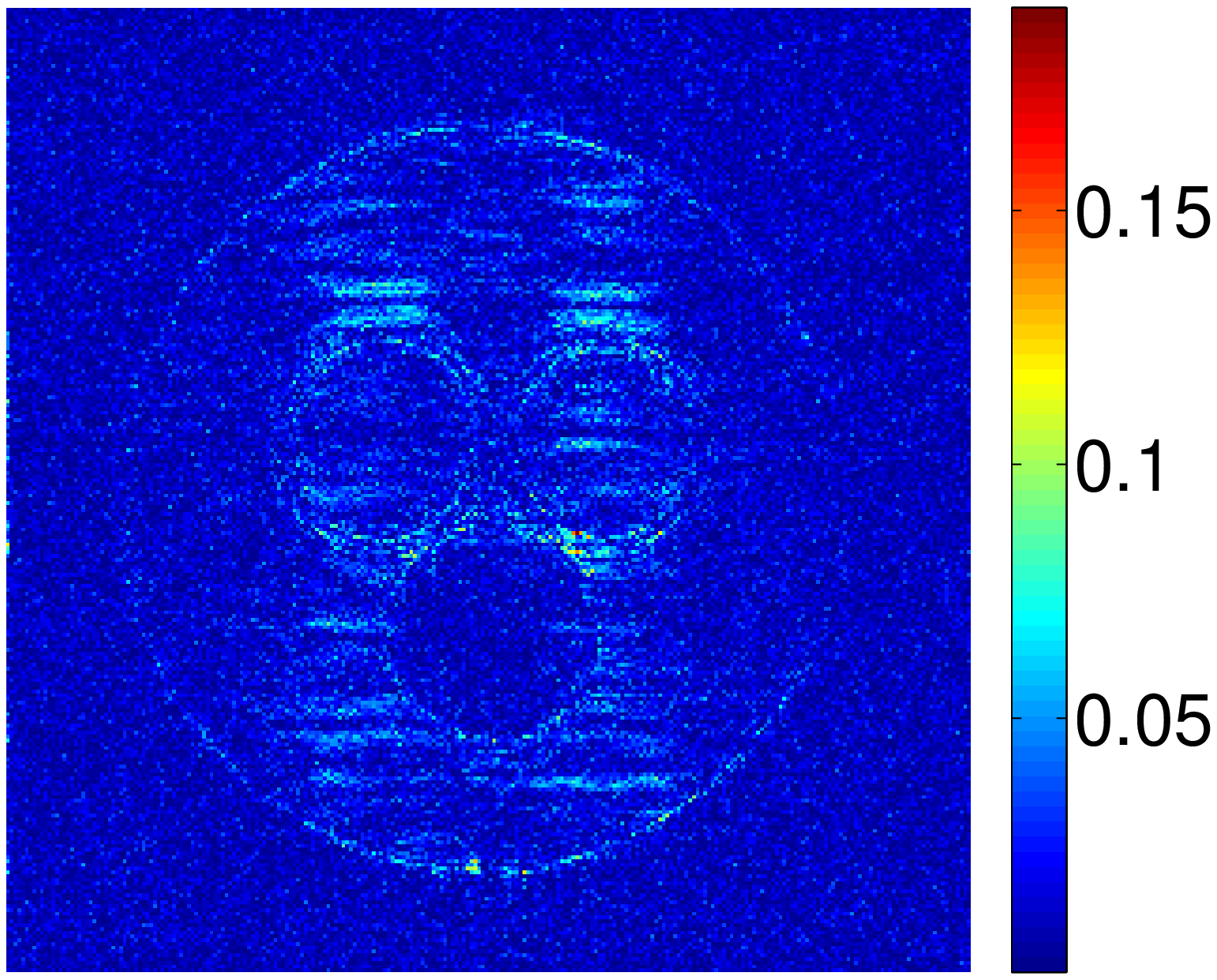}\\
 (c) & (d)\\
\includegraphics[height=1.25in]{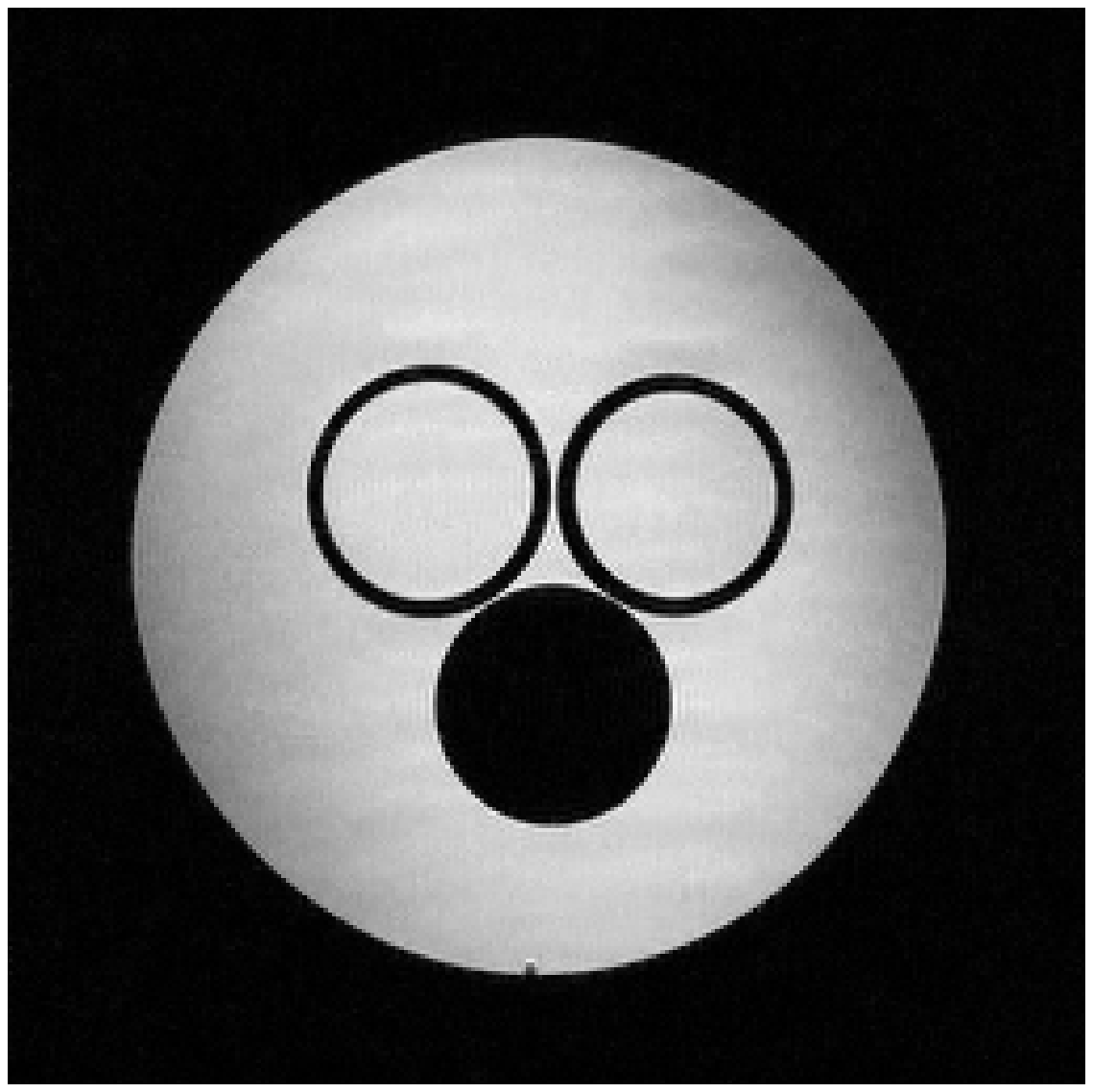}&
\includegraphics[height=1.28in]{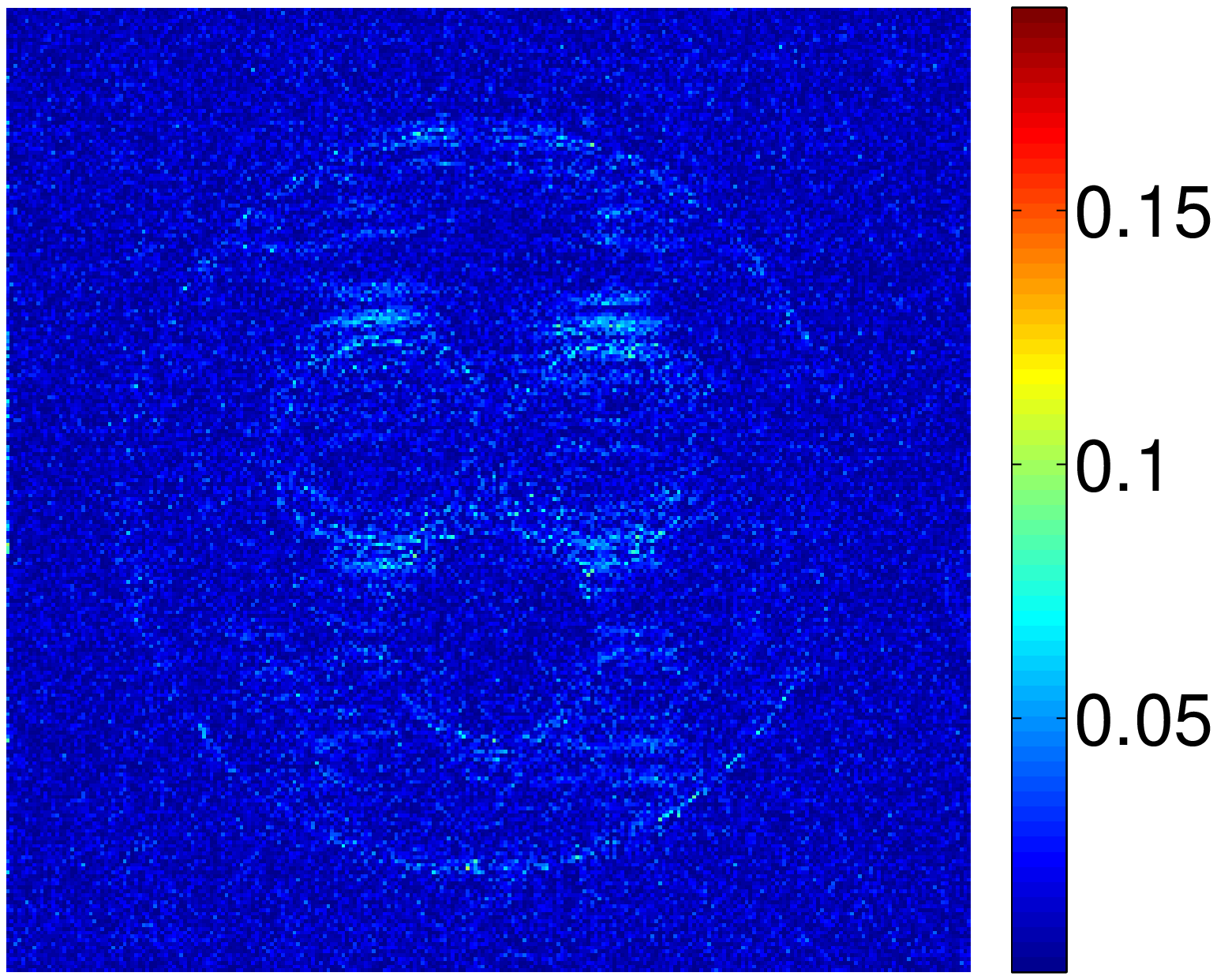}\\
(e) & (f) \\
\end{tabular}
\caption{Cartesian sampling with 2.5 fold undersampling. The sampling mask is shown in Fig. \ref{imcvbcs}(a). Reconstructions (magnitudes): (a) Sparse MRI \cite{lustig}; (c) PANO \cite{Qu2014843}; and (e) PBDWS \cite{Qu12}. Reconstruction error maps: (b)  Sparse MRI; (d) PANO; and (f) PBDWS.}
\label{im4bcs}
\end{center}
\vspace{-0.3in}
\end{figure}

\begin{figure}[!t]
\begin{center}
\begin{tabular}{cc}
\includegraphics[height=1.25in]{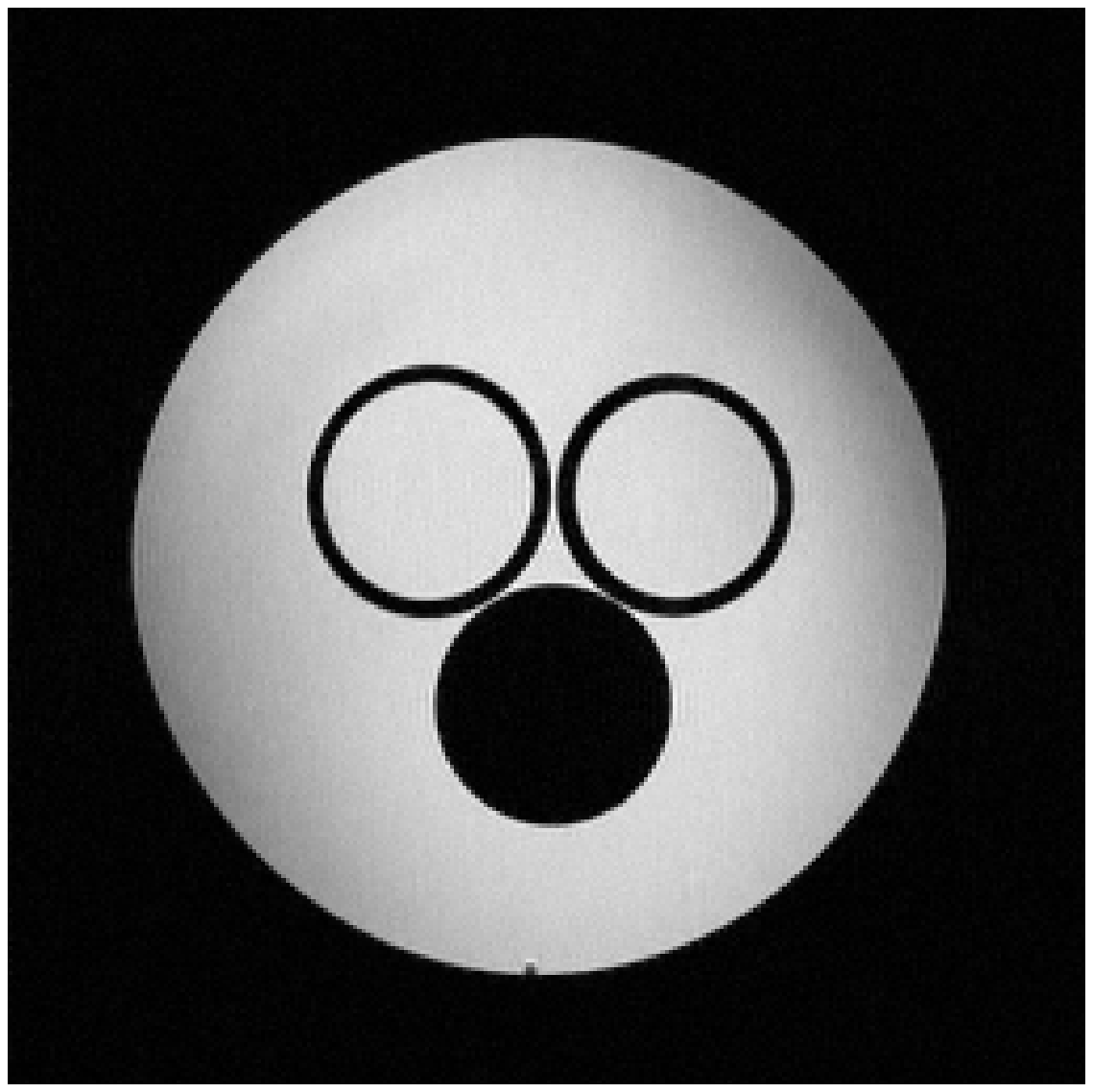}&
\includegraphics[height=1.28in]{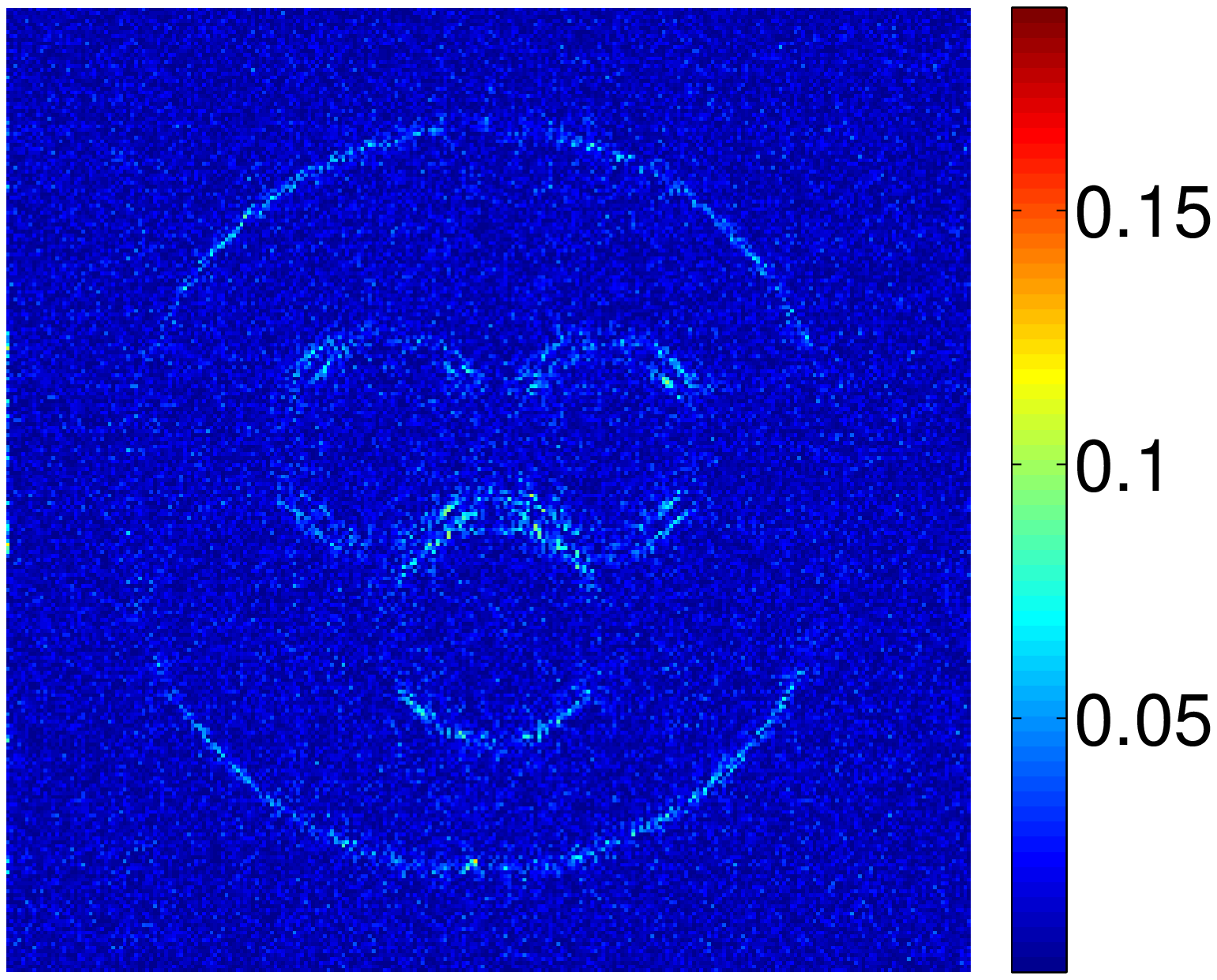}\\
(a) & (b) \\
\includegraphics[height=1.25in]{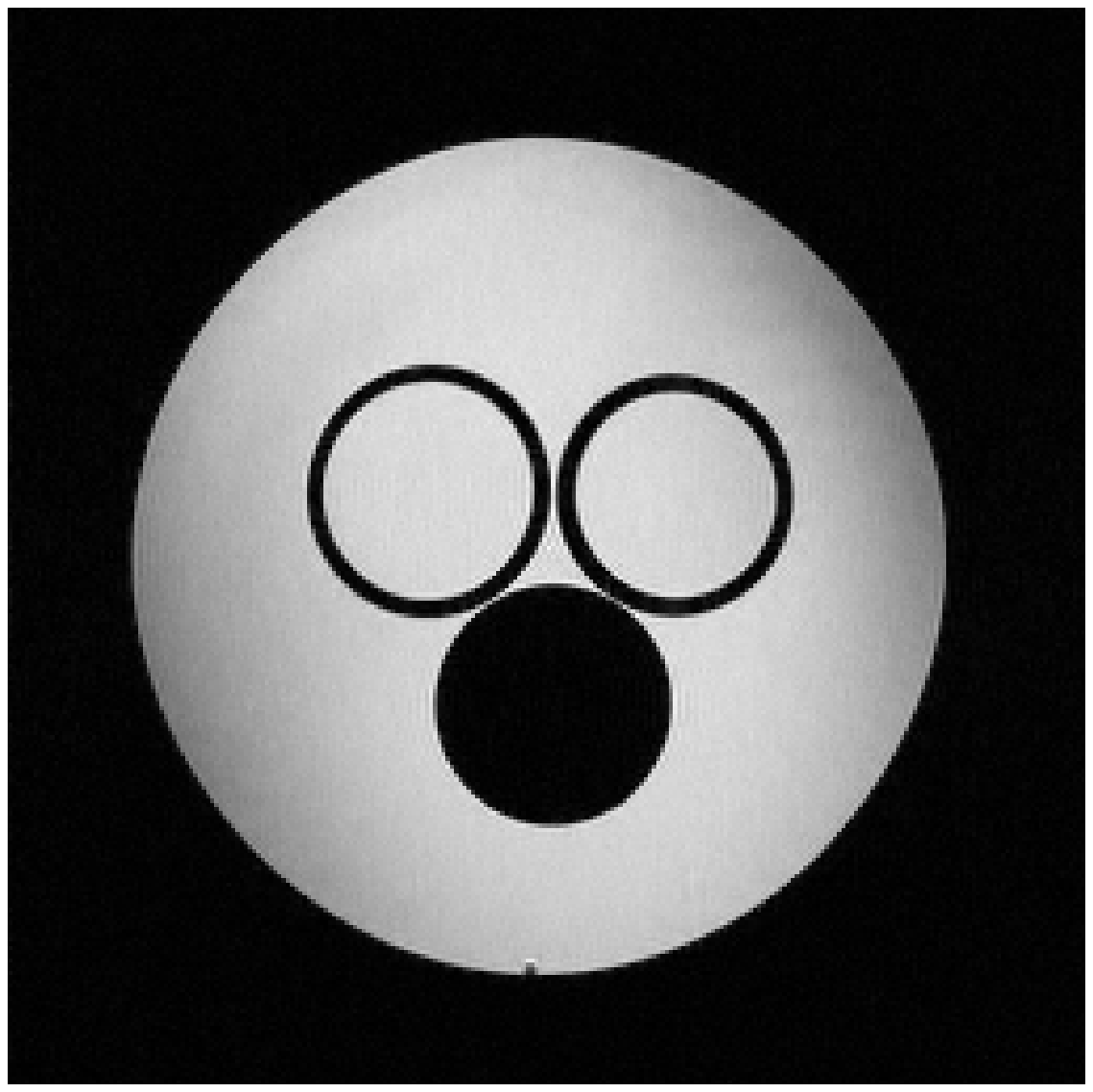}&
\includegraphics[height=1.28in]{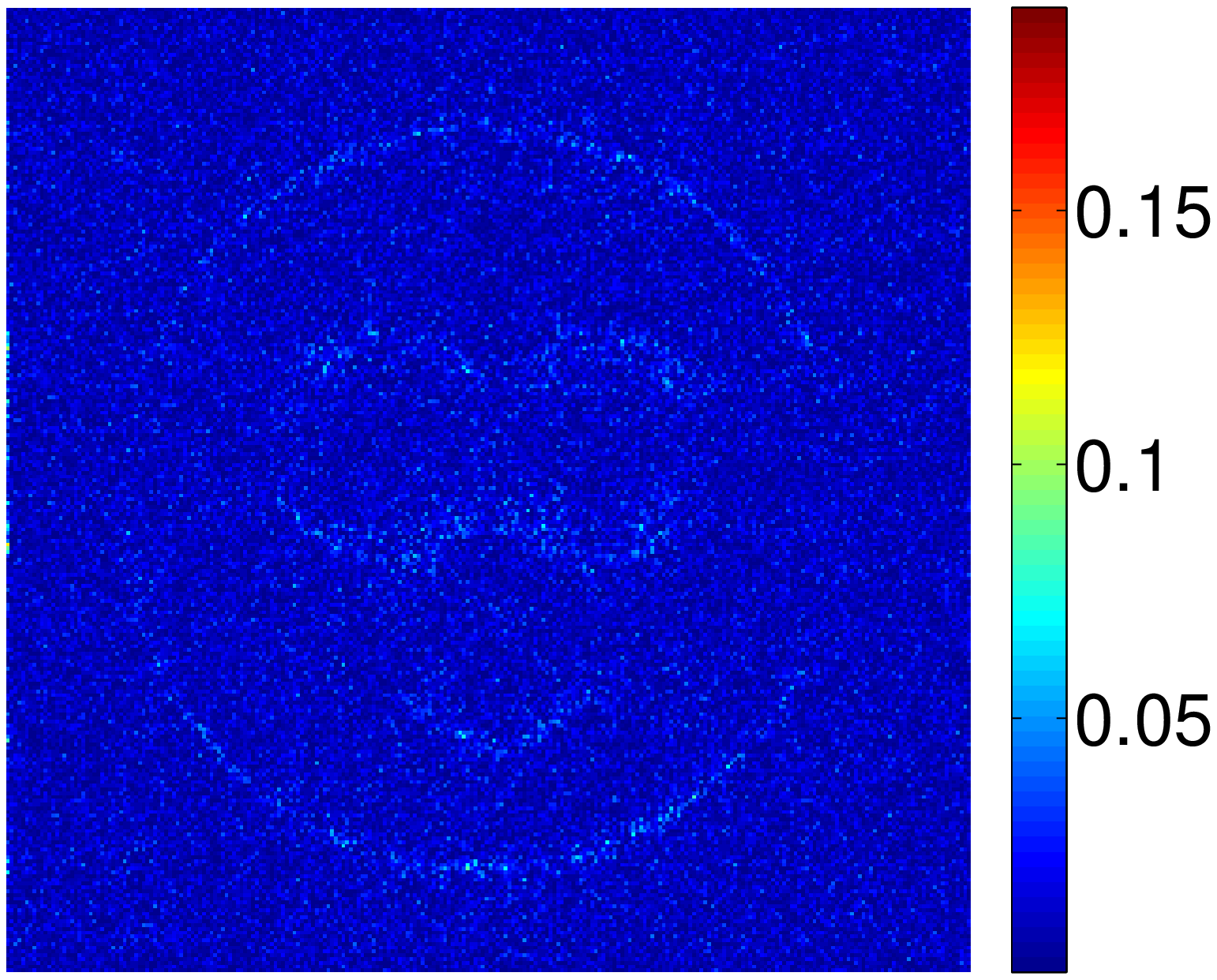}\\
 (c) & (d)\\
\includegraphics[height=1.25in]{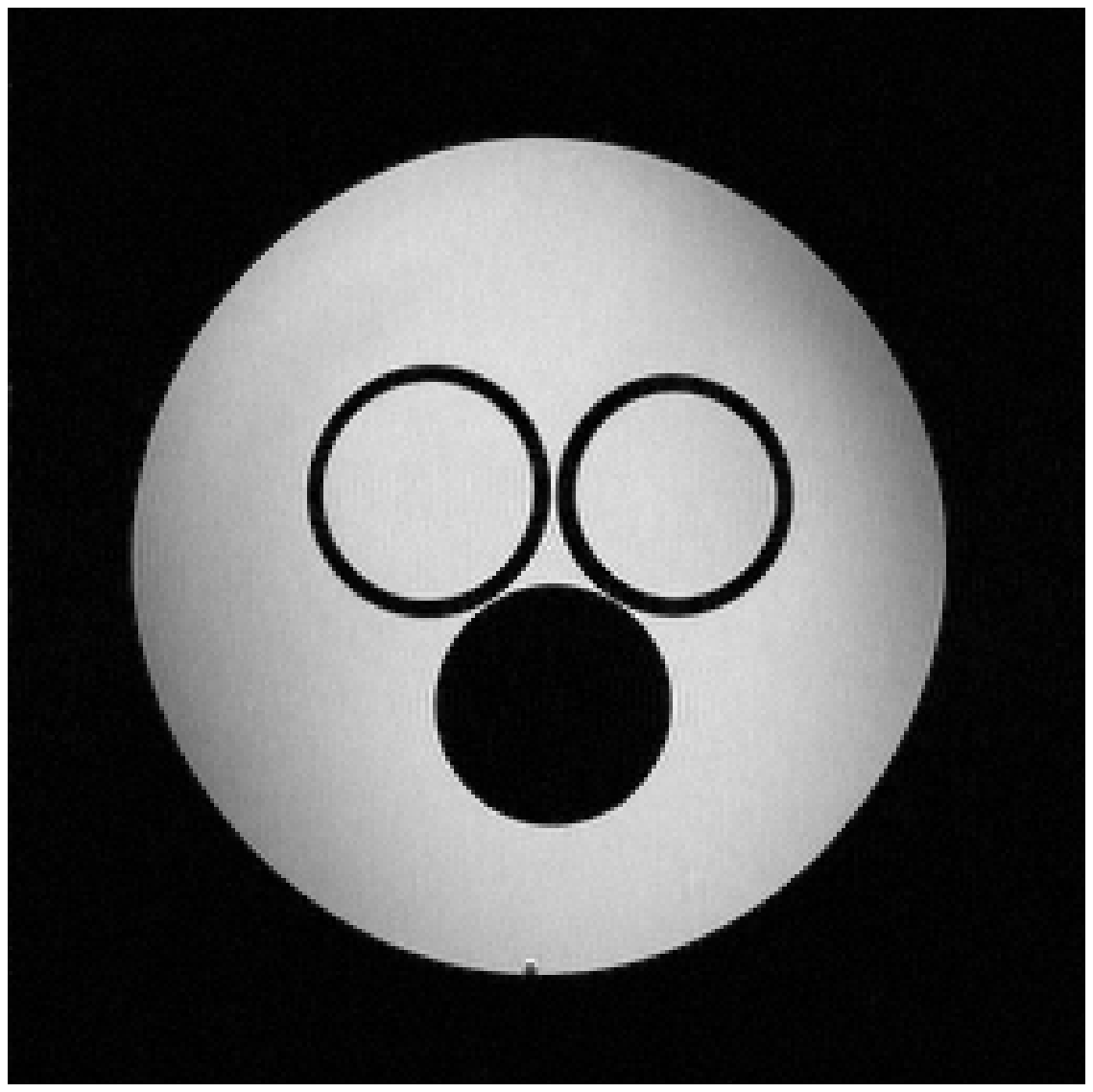}&
\includegraphics[height=1.28in]{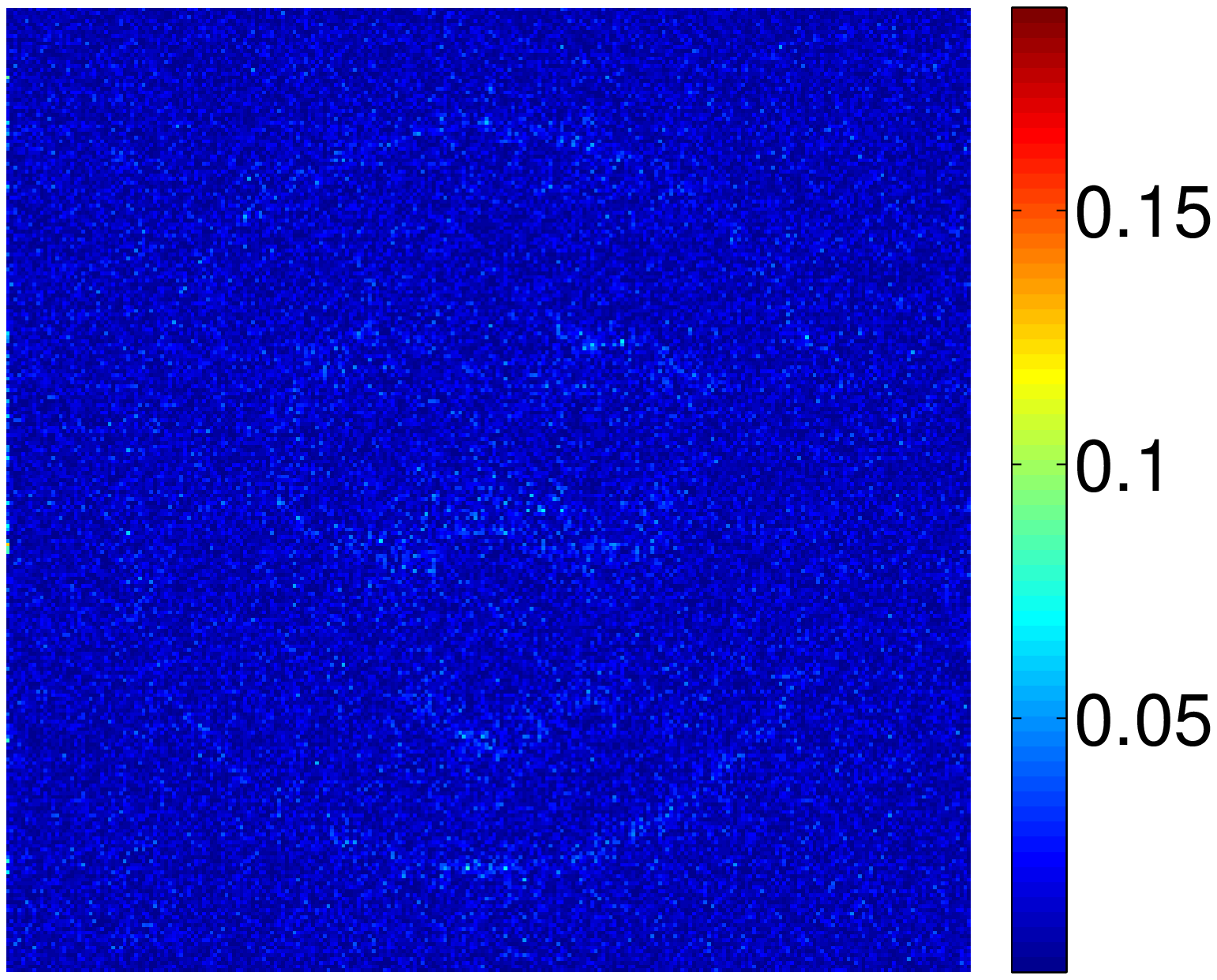}\\
(e) & (f) \\
\end{tabular}
\caption{Cartesian sampling with 2.5 fold undersampling. The sampling mask is shown in Fig. \ref{imcvbcs}(a). Reconstructions (magnitudes): (a) DLMRI \cite{bresai}; (c) UTMRI; and (e) UNITE-MRI. Reconstruction error maps: (b) DLMRI; (d) UTMRI; and (f) UNITE-MRI.}
\label{im4bcsbb}
\end{center}
\vspace{-0.2in}
\end{figure}


\begin{figure}[!t]
\begin{center}
\begin{tabular}{cc}
\includegraphics[height=1.25in]{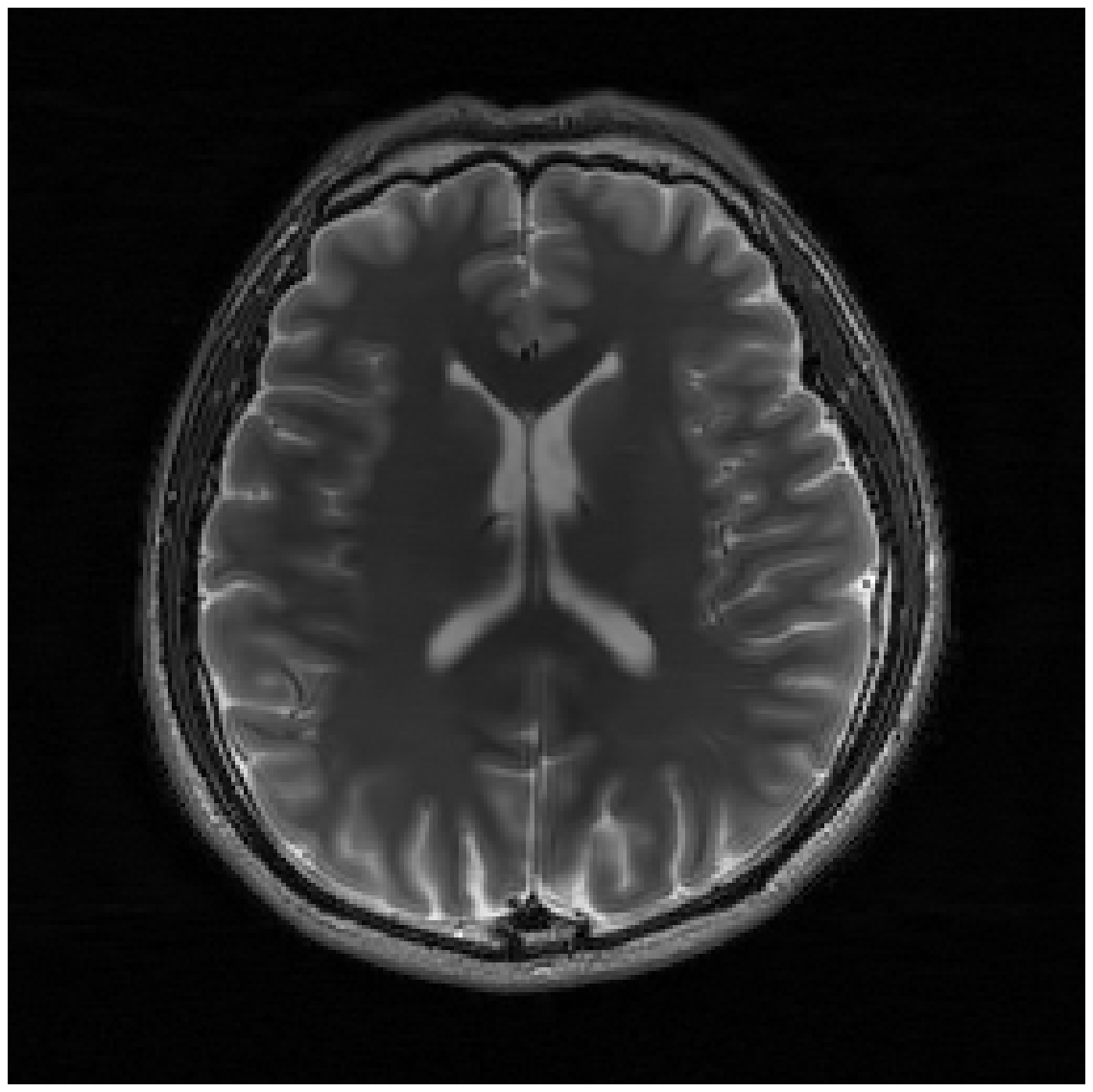} &
\includegraphics[height=1.276in]{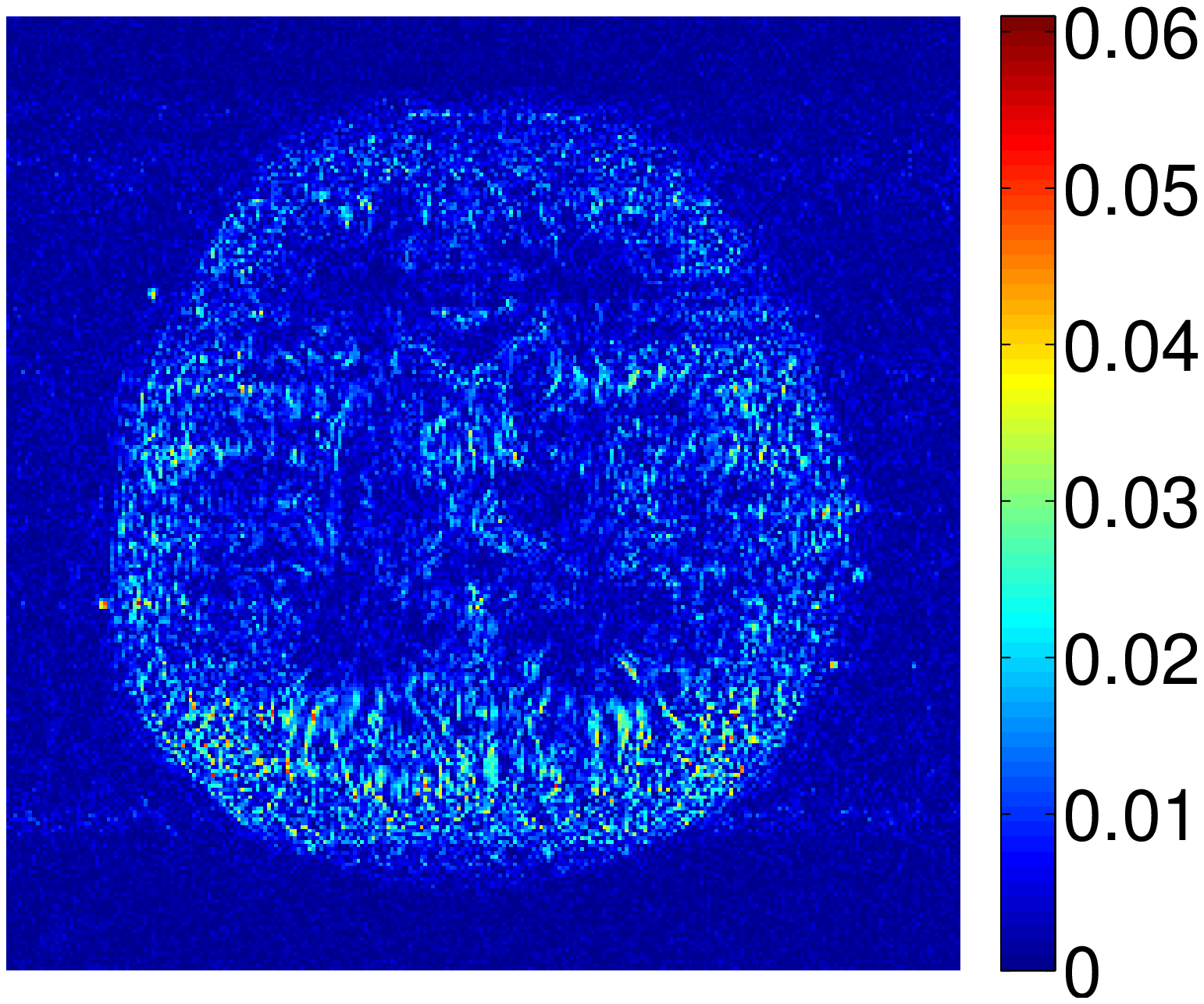} \\
(a) & (b)\\
\end{tabular}
\begin{tabular}{cc}
\includegraphics[height=1.25in]{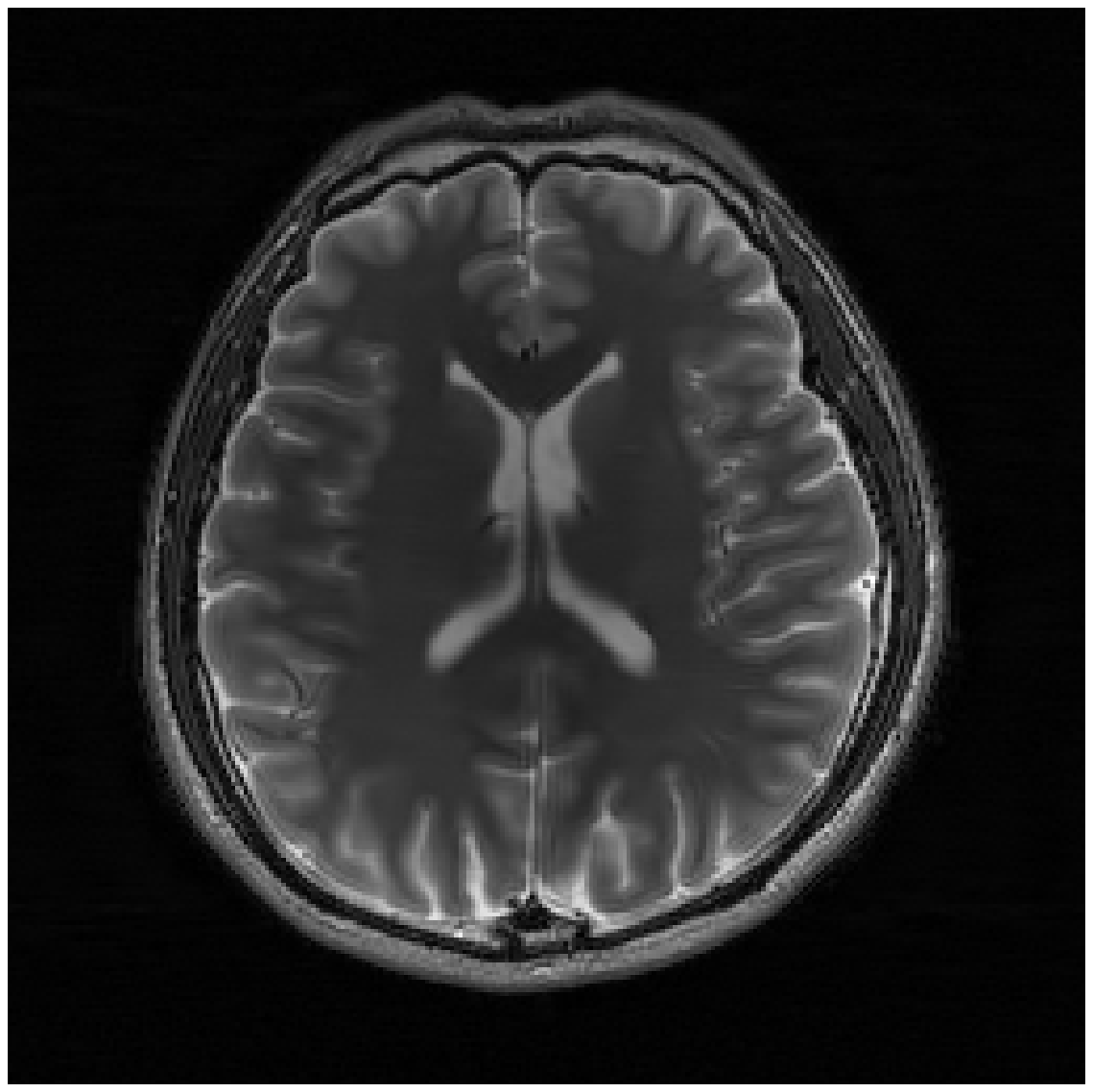}&
\includegraphics[height=1.276in]{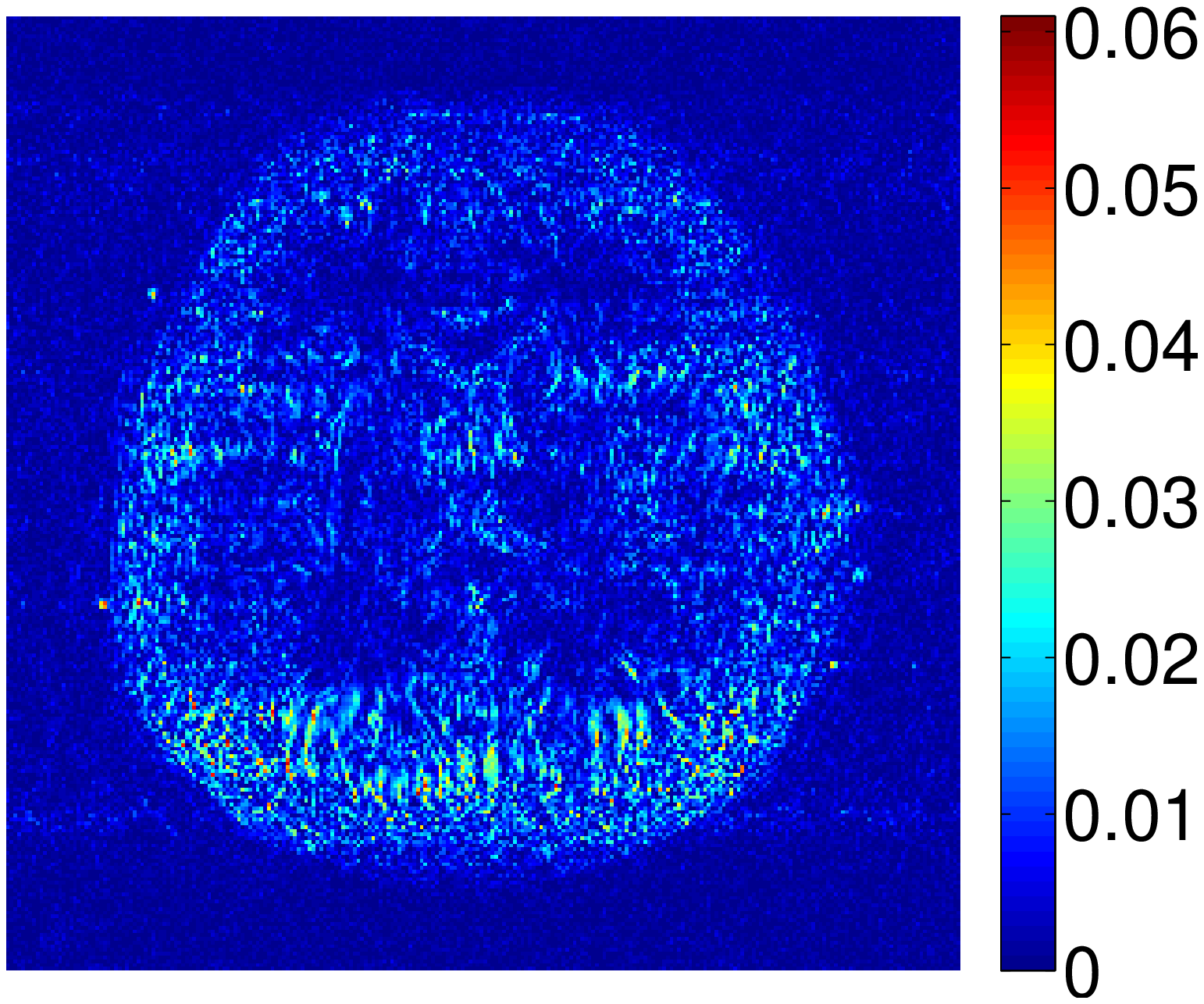} \\
(c) & (d)\\
\includegraphics[height=1.25in]{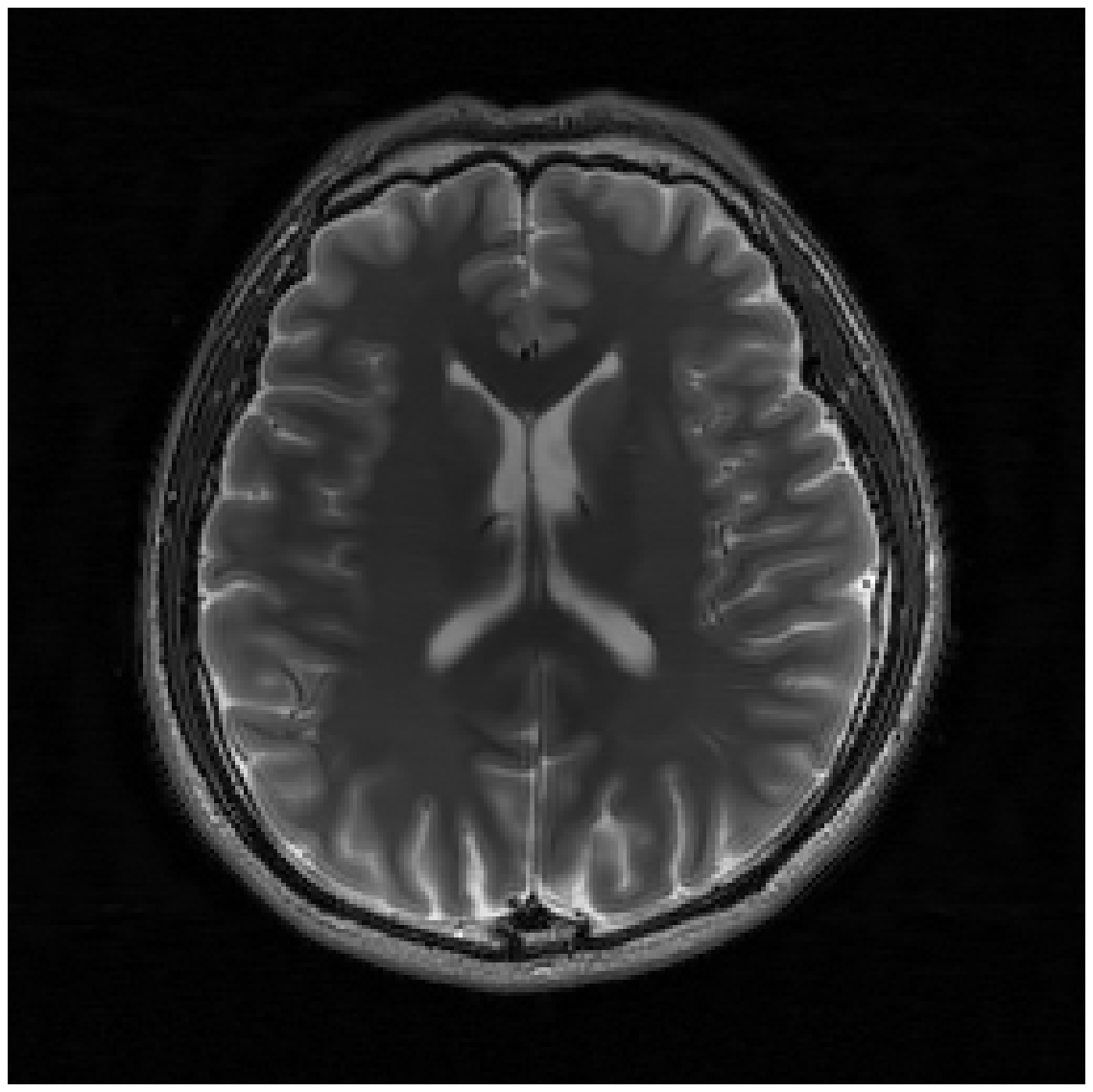}&
\includegraphics[height=1.276in]{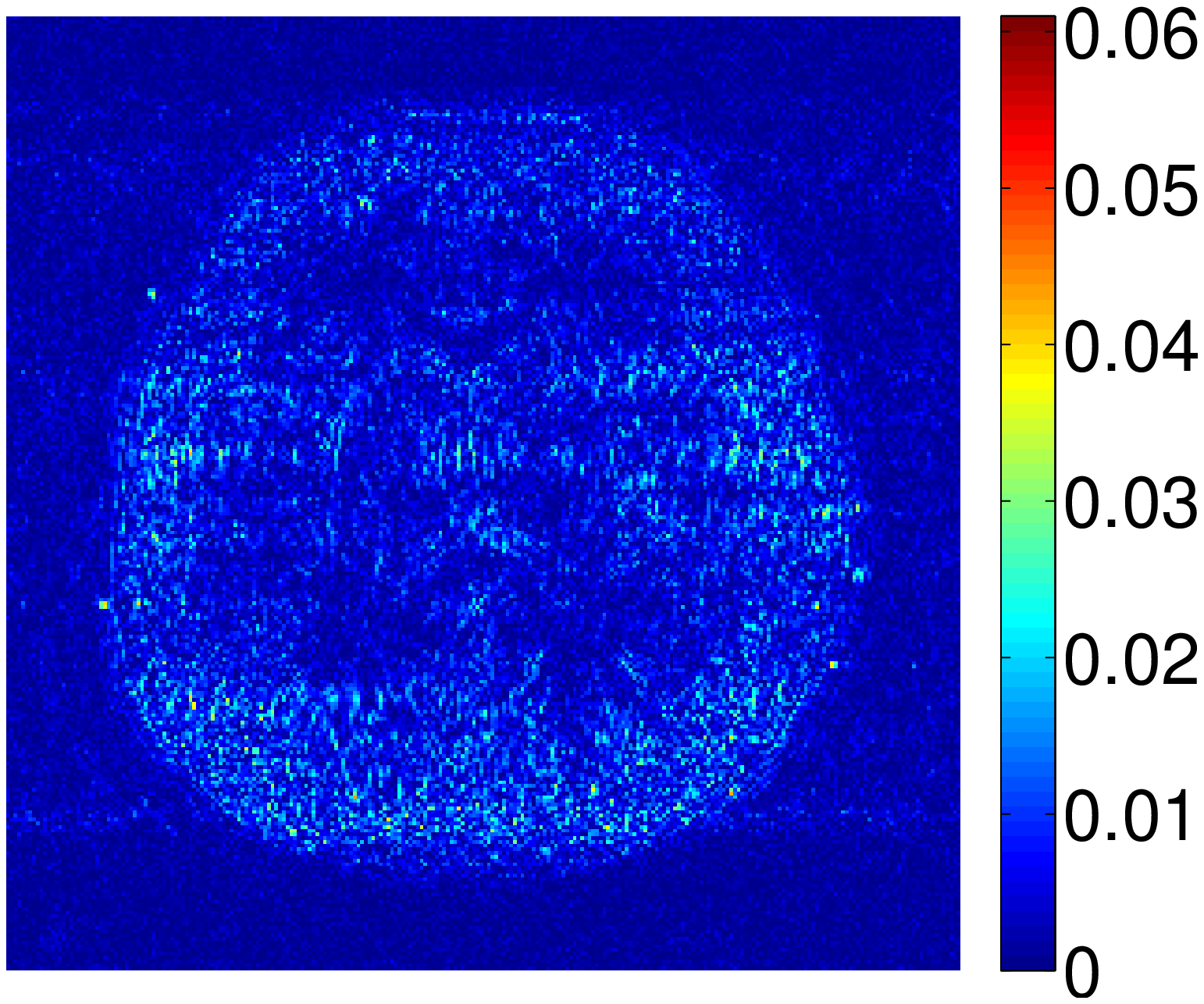} \\
(e) & (f) \\
\end{tabular}
\caption{Cartesian sampling with 2.5 fold undersampling. Sampling mask shown in Fig. \ref{im5bcsgg}(a). Reconstructions (magnitudes): (a) TLMRI (42.6 dB) \cite{sabressiims1}; (c) UTMRI (42.5 dB); and (e) UNITE-MRI with $K=16$ (44.3 dB). Reconstruction error maps: (b) TLMRI; (d) UTMRI;  and (f) UNITE-MRI.
All images here have been rotated clockwise by 90$^{\circ}$ for display.
}
\label{im5bcs}
\end{center}
\vspace{-0.2in}
\end{figure}

Fig. \ref{im4bcs} and Fig. \ref{im4bcsbb} show the reconstructions (only magnitudes are displayed here and elsewhere) obtained with several methods for the image in Fig. \ref{im1bcs}(c), with Cartesian sampling and 2.5 fold undersampling of k-space. The Sparse MRI  (Fig. \ref{im4bcs}(a)), PANO (Fig. \ref{im4bcs}(c)), PBDWS (Fig. \ref{im4bcs}(e)), and DLMRI (Fig. \ref{im4bcsbb}(a)) reconstructions show some residual artifacts that are mostly removed in the
UTMRI (Fig. \ref{im4bcsbb}(c)) 
and UNITE-MRI (Fig. \ref{im4bcsbb}(e)) reconstructions.
Figs. \ref{im4bcs} and  \ref{im4bcsbb} also show the reconstruction error maps (i.e., the magnitude of the difference between the magnitudes of the reconstructed and reference images)  for various methods.
The error maps for the transform-based BCS methods
(UTMRI, UNITE-MRI) clearly show much smaller image distortions than those for other methods. In particular, the error map for the UNITE-MRI method shows fewer artifacts along the image edges than that for the UTMRI scheme.


Fig. \ref{im5bcs} shows another example of reconstructions obtained with the UTMRI (Fig. \ref{im5bcs}(c)) and UNITE-MRI (Fig. \ref{im5bcs}(e)) methods, along with the TLMRI reconstruction  (Fig. \ref{im5bcs}(a)), for the image in Fig. \ref{im1bcs}(b) with Cartesian sampling and 2.5 fold undersampling of k-space.
The reconstruction error maps for TLMRI (Fig. \ref{im5bcs}(b)), UTMRI (Fig. \ref{im5bcs}(d)), and UNITE-MRI (Fig. \ref{im5bcs}(f)) are also shown. The UNITE-MRI method with 16 clusters ($K=16$) clearly provides a much better reconstruction of image features (i.e., fewer artifacts) than the single transform-based UTMRI and TLMRI schemes in this case.

The average runtimes for the Sparse MRI, DLMRI, PBDWS, PANO\footnote{Another faster version of the PANO method (that uses multi-core CPU parallel computing) is also publicly available \cite{PANOwebFast}. However, we found that although this version (employed with the built-in parameter settings) has an average runtime of only 25 seconds in Table \ref{tab2bcs}, it also provides 0.4 dB worse reconstruction PSNR on an average than the version \cite{PANOweb} used in Table \ref{tab2bcs}.}, TLMRI,
UTMRI, and UNITE-MRI methods in Table \ref{tab2bcs} are 166 seconds,
2581 seconds, 423 seconds, 223 seconds, 430 seconds, 291 seconds, and 1480 seconds, respectively.
The PBDWS runtime includes the time taken for computing the initial SIDWT-based reconstruction or guide image \cite{Qu12} in the PBDWS software package \cite{Quweb}.
Note that the runtimes for the proposed UTMRI and UNITE-MRI algorithms were obtained by employing our unoptimized Matlab implementations of these methods, whereas the implementations of PBDWS and PANO are based on MEX (or C) code.
While UTMRI has small runtimes, the larger runtimes for UNITE-MRI can be substantially reduced (at the price of a small degradation in the reconstruction PSNRs) by performing the (computationally expensive) clustering step less often (compared to the transform update, sparse coding, and image update steps) in the block coordinate descent algorithm.
For the same number of (120) algorithm iterations, UTMRI is faster than TLMRI because of the more efficient updates in UTMRI.
We expect speed-ups for our algorithms with conversion of the code to C/C++, code optimization, and parallel computing.

\vspace{-0.1in}
\subsection{The Number of Clusters in UNITE-MRI} \label{resu3}

\begin{figure}[!t]
\begin{center}
\begin{tabular}{cc}
\includegraphics[height=1.25in]{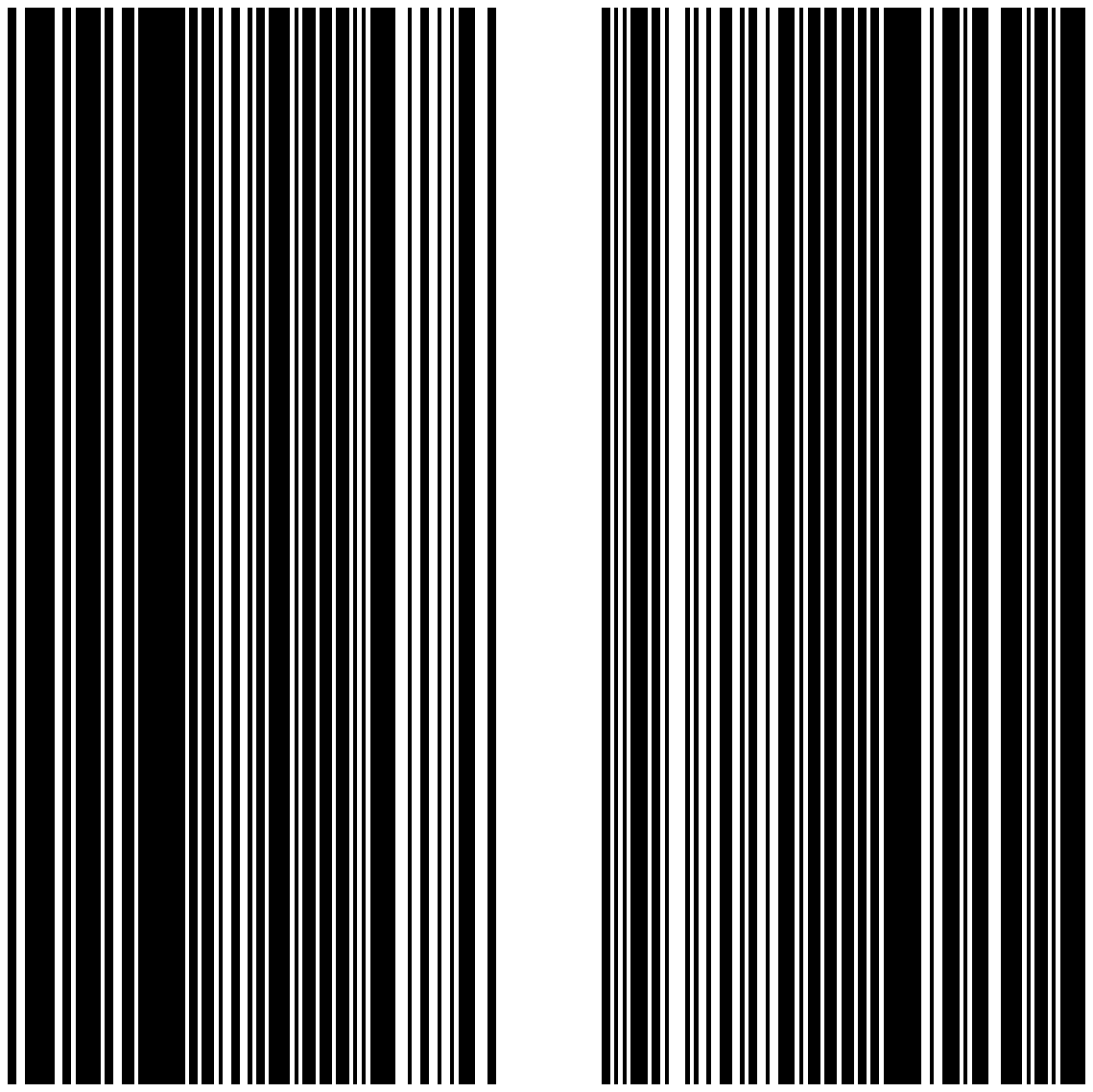} &
\includegraphics[height=1.25in]{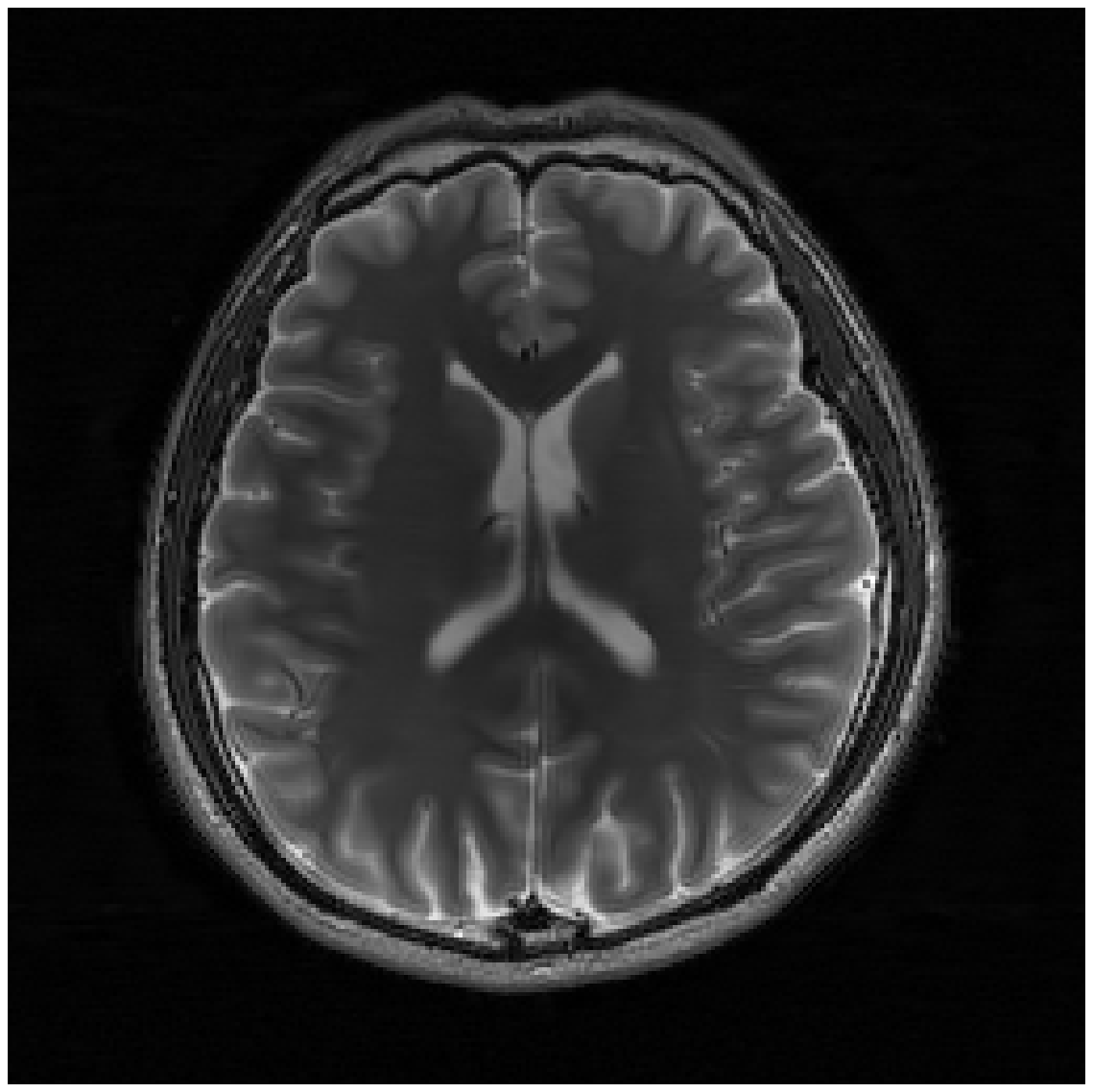}\\
(a) & (b)\\
\includegraphics[height=1.25in]{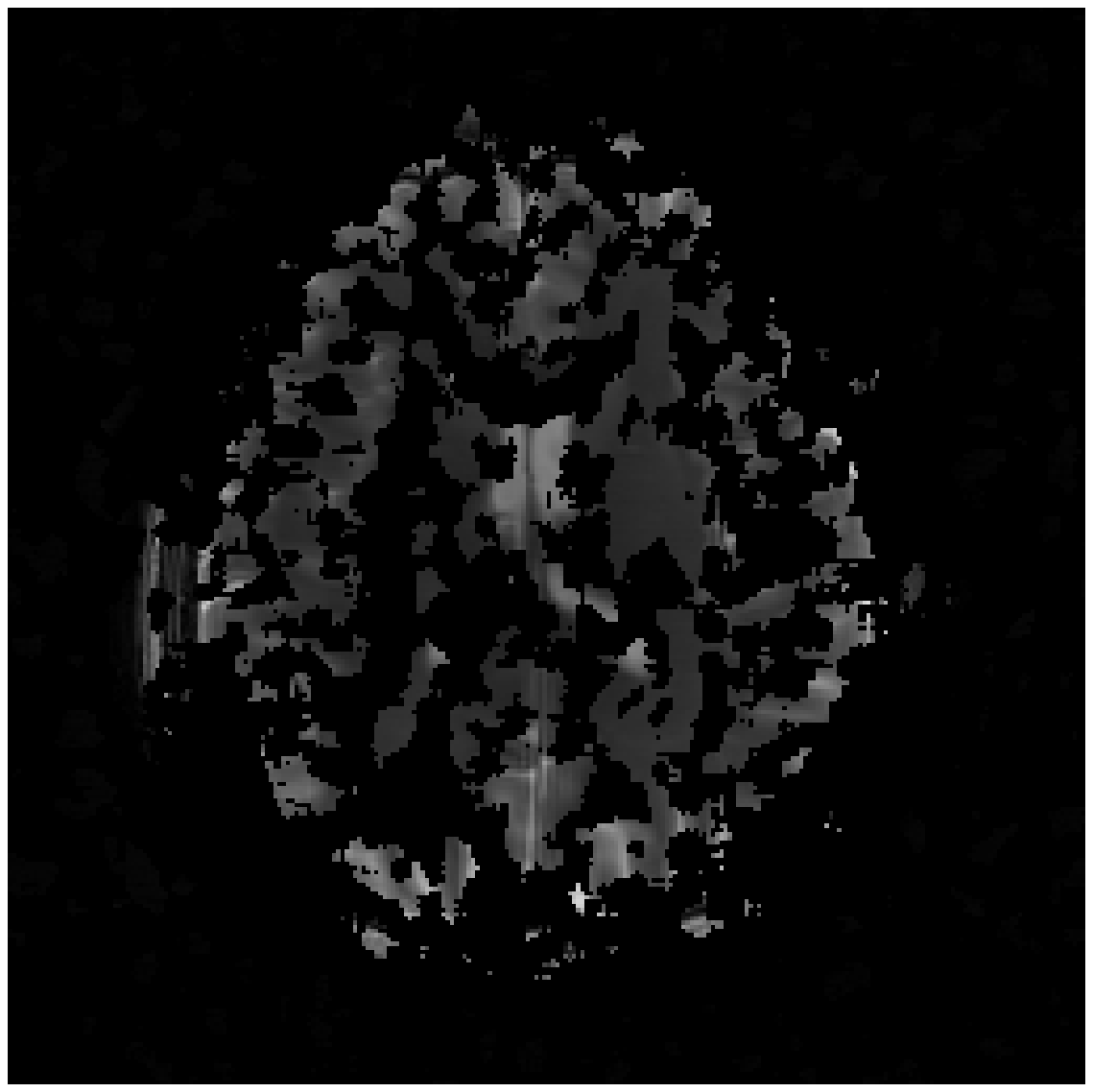} &
\includegraphics[height=1.25in]{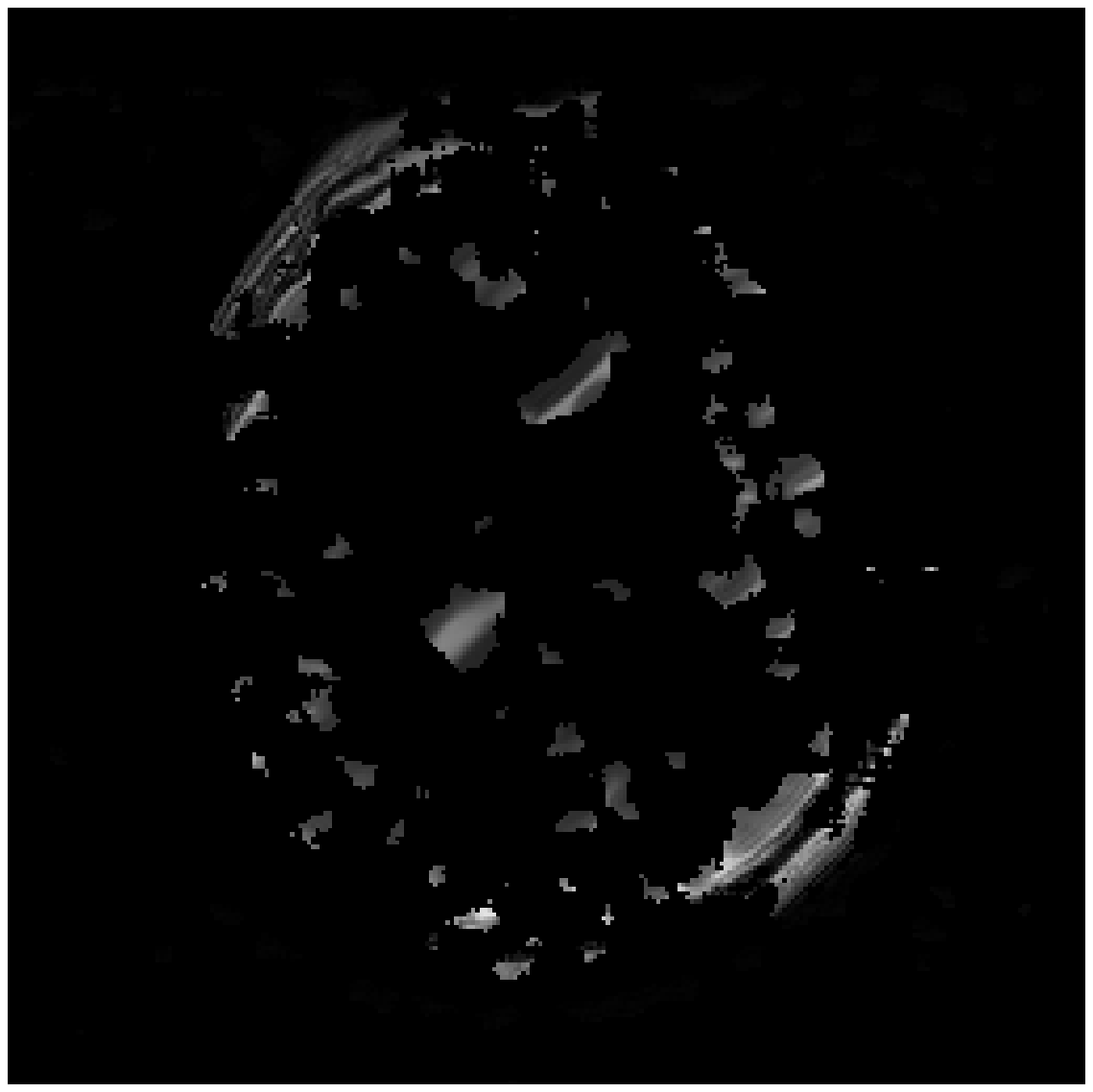} \\
(c) & (d)\\
\includegraphics[height=1.25in]{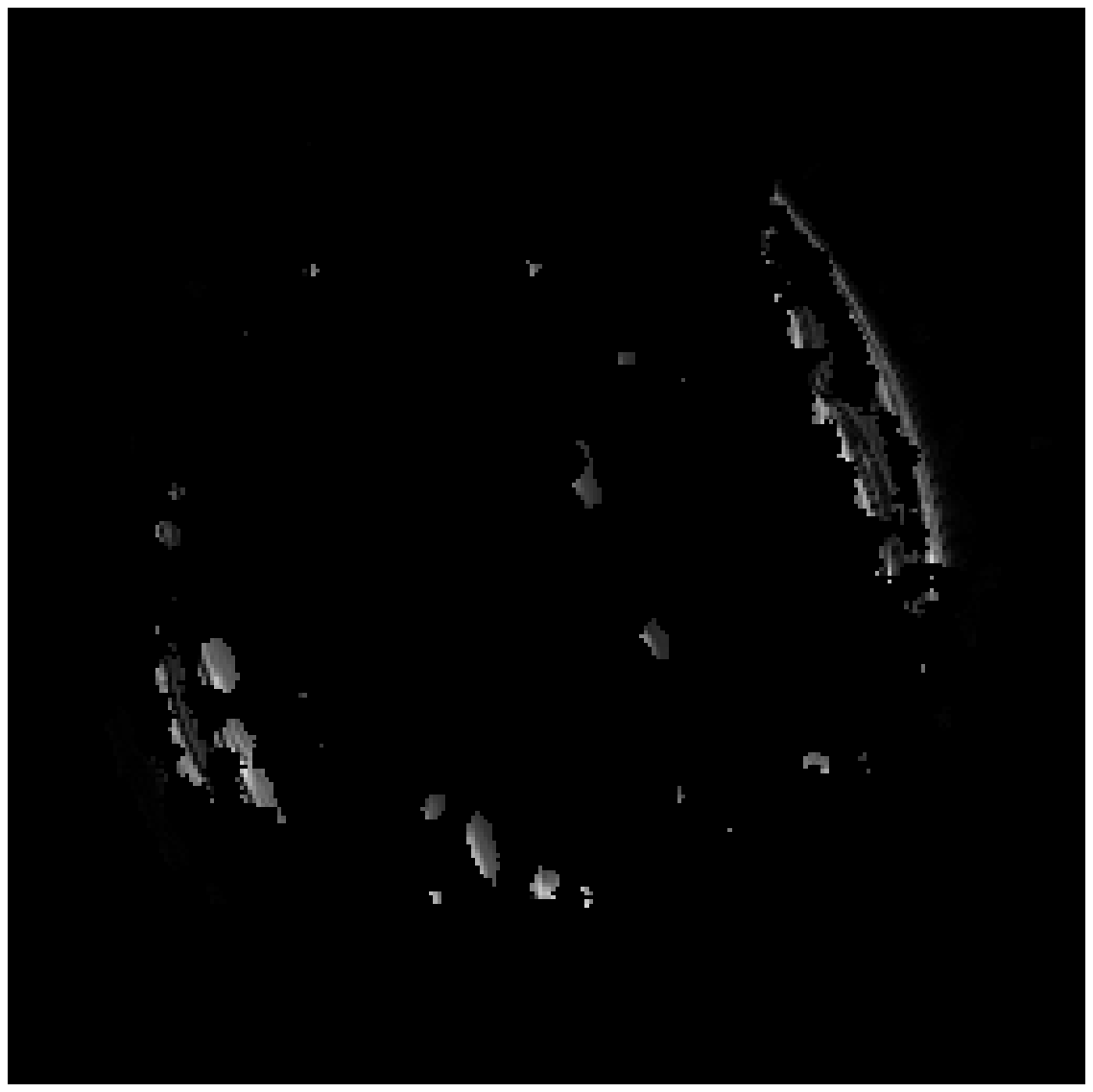}&
\includegraphics[height=1.25in]{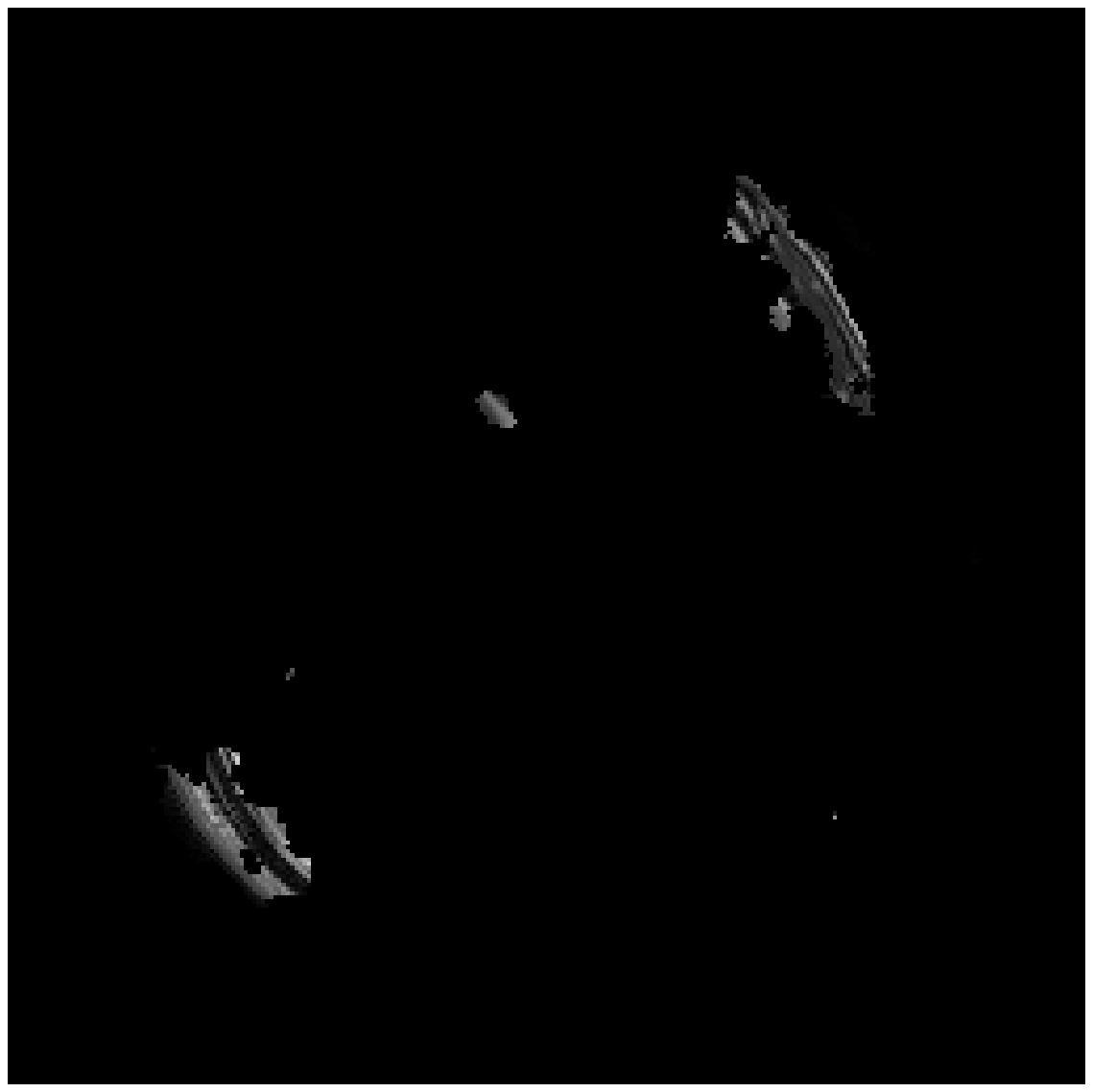} \\
(e) & (f) \\
\end{tabular}
\begin{tabular}{cc}
\includegraphics[height=1.16in]{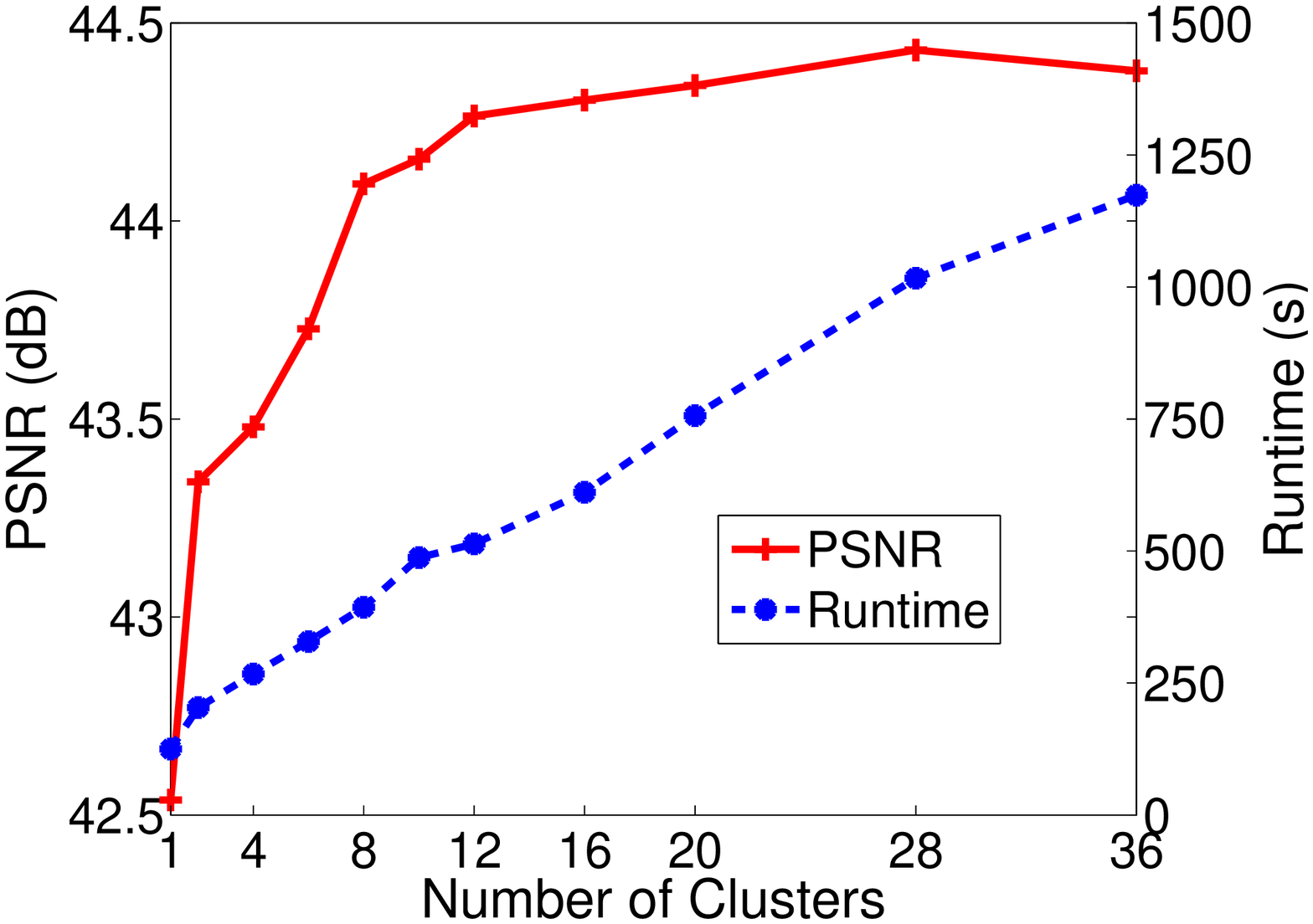} &
\includegraphics[height=1.17in]{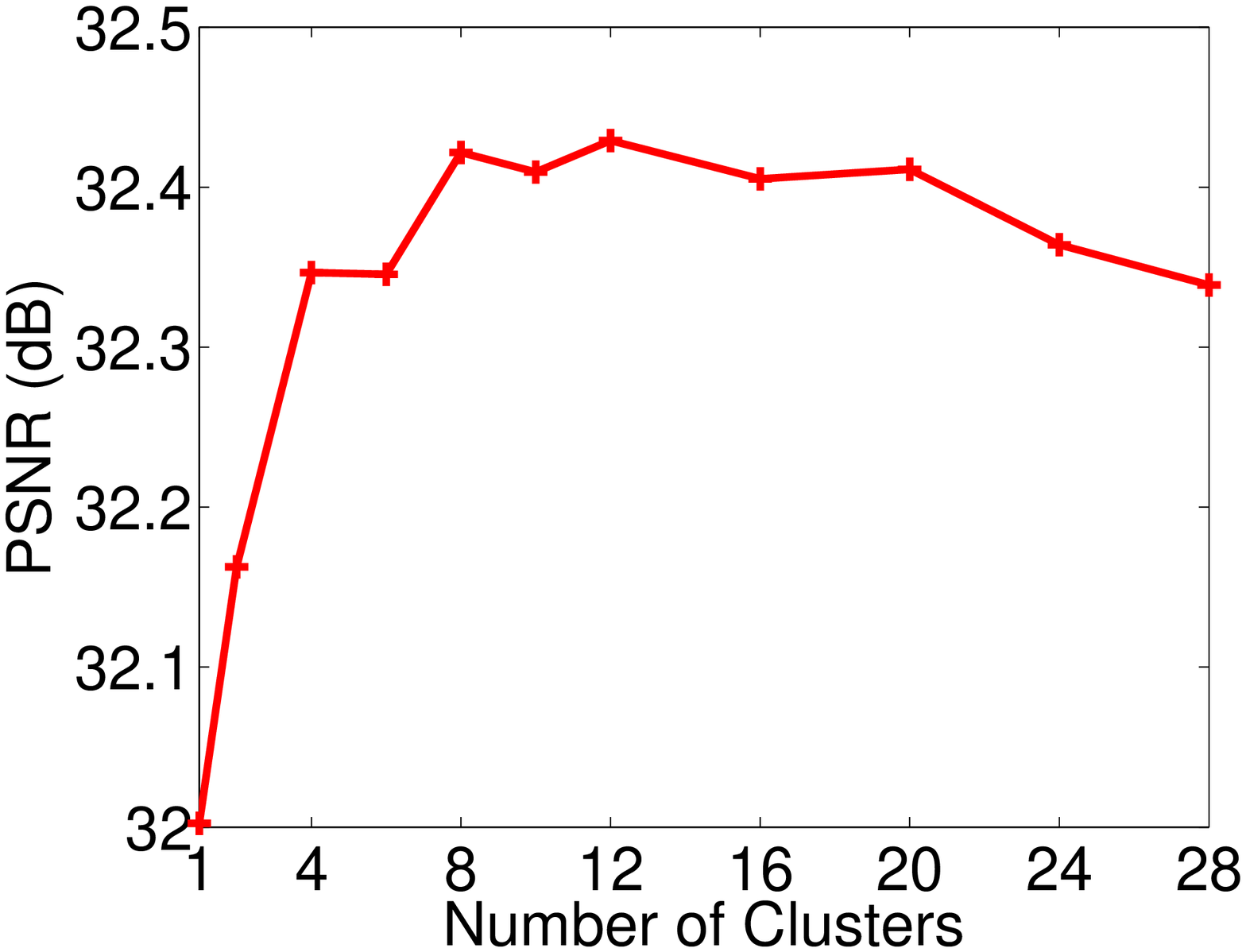} \\
(g) & (h) \\
\end{tabular}
\caption{Cartesian sampling with 2.5 fold undersampling: (a) k-space sampling mask;  (b) UNITE-MRI reconstruction (using k-space samples of image in Fig. \ref{im1bcs}(b)) magnitude for $K=10$ (44.2 dB); (c)-(f) image pixels (with reconstructed intensities) in (b) grouped into four specific clusters overlaid on black backgrounds; (g) reconstruction PSNR and runtime vs. number of clusters $K$, when k-space samples are obtained using data in Fig. \ref{im1bcs}(b); and (h) reconstruction PSNR (computed with respect to reference in Fig. \ref{im1bcs}(b)) vs. number of clusters $K$, when the measurements are obtained from a noisier version (PSNR = 26.7 dB) of the image in Fig. \ref{im1bcs}(b).
The images (a)-(f) have all been rotated clockwise by 90$^{\circ}$ for display.}
\label{im5bcsgg}
\end{center}
\vspace{-0.2in}
\end{figure}

Here, we investigate the peformance of the UNITE-MRI method as a function of the number of clusters $K$. We work with the same data and k-space sampling as in Fig. \ref{im5bcs}, and perform image reconstruction using the UNITE-MRI method at various values of $K$ (all other algorithm parameters are the same as in  Fig. \ref{im5bcs}). Fig. \ref{im5bcsgg}(g) shows the image reconstruction PSNRs for UNITE-MRI for various $K$ values. The PSNR improves monotonically and significantly as $K$ is increased above $1$ (i.e., UTMRI). This is because UNITE-MRI learns richer and more specific or adaptive models that provide sparser representations for patches, and hence better iterative reconstructions. However, for very large $K$ values, the PSNR saturates and begins to decrease.
This is because it becomes impossible to reliably learn very rich or complex (non-trivial) models from limited compressive measurements and from limited number of patches.
Fig. \ref{im5bcsgg}(g) shows the UNITE-MRI runtimes varying quite linearly with the number of clusters, although at a more gradual rate than $O(Kn^{2} N)$.\footnote{The actual runtimes would also be quite dependant on the specific implementation.}

Fig. \ref{im5bcsgg}(b) shows the UNITE-MRI reconstruction with $K=10$ clusters. Figs. \ref{im5bcsgg}(c)-(f) show image pixels from the reconstructed image that are clustered into four specific classes (similar to Fig. \ref{imcvbcs2}). The pixels from each of these classes (shown with the reconstructed intensities) are overlaid on a black background in these images. The results show that UNITE-MRI groups together regions of the brain image with specific types of features or edges.

The data obtained from MR scanners (in Fig. \ref{im1bcs}) and used in our experiments typically contain physical noise. To explicitly evaluate the effect of measurement noise in UNITE-MRI, we add simulated i.i.d. complex Gaussian noise to the image in Fig. \ref{im1bcs}(b).
This is equivalent to adding noise to the simulated k-space data thus modeling a realistic higher noise acquisition. The corrupted  image has a PSNR (computed with respect to  Fig. \ref{im1bcs}(b)) of 26.7 dB. We now repeat the experiment of Fig. \ref{im5bcsgg}(g), but sample the k-space of the corrupted image. All  algorithm parameters are the same as before, except that $\nu = 30/p$, and $\eta$ is set as for Fig. \ref{im1bcs}(a).
Fig. \ref{im5bcsgg}(h) shows the image reconstruction PSNRs computed with respect to Fig. \ref{im1bcs}(b), for UNITE-MRI for various $K$ values. In this case, the PSNR saturates earlier than in Fig. \ref{im5bcsgg}(g), but UNITE-MRI still provides up to 0.43 dB better PSNRs than UTMRI ($K=1$), and both methods achieve better PSNRs than the original (fully sampled) corrupted image (26.7 dB).

\vspace{-0.1in}
\subsection{Extensions and Improvements} \label{res6545}

Our results show the promise of the proposed blind compressed sensing methods for MRI. The PSNRs for our schemes in our experiments can be further improved with better parameter selection strategies.
There may be several directions to potentially improve the proposed methods. For example, combining the UNITE-MRI scheme (or, Algorithm A2) with the patch-based directional wavelets model \cite{Qu11, Qu12}, or with non-local patch similarity ideas \cite{Qu2014843, Y7027820} could potentially boost the BCS performance further. Incorporating additional information from related reference images (e.g., from a database) in the proposed framework could make our schemes more robust to noise and other artifacts. While we focused on learning a union of unitary transforms, because of the efficiency in computations that they provide, we plan to investigate unions of more general well-conditioned transforms \cite{sbclsTS2} or unions of overcomplete transforms \cite{ovicasp2} in applications in future work. 

In our experiments, we simulated single coil-based undersampled MRI acquisitions. To the extent that our results show reasonable signal to noise (and distortion) ratio at high acceleration, our approach would be equally applicable in practice, whether the acquisition is using a single coil or more. In usual parallel coil MRI (p-MRI) setups, there are also noise amplification issues at high acceleration factors (such as 10x or 20x), because of the increased condition number of the inverse problem. However, our data-driven and sparsity-driven approach may be able to provide accelerations on top of that provided by p-MRI. We have not explored this extension here, and leave its investigation to future work.

Finally, although image reconstruction is the primary goal in this work, the proposed BCS method involving a union of transforms model achieves joint image reconstruction and unsupervised patch clustering (resulting in image segmentation). There have been other methods  for joint reconstruction and segmentation involving
information theoretic criteria \cite{kerfoot1}, multiscale approaches \cite{kerfoot2}, or
mixture models \cite{hsiao2, sompel2, cabale2}. Incorporating additional prior information on the specific classes (e.g., tissue types) in our method (a semi-supervised strategy) could potentially lead to clinically meaningful segmentations. The exploration of such extensions is left for future work.
The union of transforms methodology could also potentially capture textures and other features in patient or disease datasets.


\section{Conclusions}
\label{sec5}  
In this work, we presented a novel sparsifying transform-based framework for blind compressed sensing. The patches of the underlying image(s) were modeled as approximately sparse in an unknown (unitary) sparsifying transform, and this transform was learnt jointly with the image from only the compressive measurements. We also considered a union of transforms model that better captures the diverse features of natural images. The proposed blind compressed sensing algorithms involve highly efficient updates. We demonstrated the superior performance of the proposed schemes over several recent methods for MR image reconstruction from highly undersampled measurements. In particular, the union of transforms model outperformed the single transform model in terms of the achieved quality of image reconstructions. The usefulness of the proposed BCS methods in other inverse problems and imaging applications merits further study.

\section*{Acknowledgment}
The authors thank Prof. Jeffrey A. Fessler at the University of Michigan for his feedback and comments on this work.

\ifCLASSOPTIONcaptionsoff
  \newpage
\fi



%



\bibliographystyle{./IEEEtran}
\bibliography{./IEEEabrv,./UNBCS_v9}

\end{document}